%% file: main.tex
\documentclass{article}
\usepackage[12pt]{extsizes}
\usepackage[utf8]{inputenc}
\usepackage{amsmath}
\usepackage{amssymb}
\usepackage{amsthm}
\usepackage{natbib}
\usepackage{graphicx}
\usepackage{lipsum}
\usepackage{enumitem}
\usepackage[frozencache,cachedir=.]{minted} 
\usepackage{caption}
\usepackage{mdframed}

\usepackage[dvipsnames,table]{xcolor}
\usepackage[hidelinks]{hyperref}

\definecolor{myurlcolor}{HTML}{123463}
\definecolor{tdc_color}{RGB}{10,128,122}
\definecolor{dc_color}{RGB}{230, 245, 244}
\definecolor{ds_color}{RGB}{195, 230, 227}
\definecolor{ms_color}{RGB}{150, 214, 209}
\hypersetup{
    colorlinks=true,
    linkcolor=tdc_color,
    filecolor=tdc_color,      
    urlcolor=tdc_color,
    citecolor=tdc_color
}

\usepackage{authblk}

\usepackage{geometry}
\usepackage{tocloft}

\usepackage{mathtools}
\usepackage{amsfonts}
\usepackage{relsize}
\usepackage{multirow}
\usepackage{booktabs}
\usepackage{tabularx}
\usepackage{bbding}
\usepackage{listings}

\usepackage{xspace}
\usepackage[ruled,vlined]{algorithm2e}

\usepackage{multirow}

\newcommand{\mname}{\text{TDC}\xspace}
\newcommand{\eg}{\emph{e.g.}\xspace}
\newcommand{\ie}{\emph{i.e.}\xspace}

\newtheorem*{problem*}{Problem}

\definecolor{light-gray}{gray}{0.96}

\newcommand{\code}[1]{\texttt{#1}}
\newcommand{\codebox}[1]{\colorbox{light-gray}{\textbf{\texttt{#1}}}}

\newcommand{\dataset}[1]{{\color{tdc_color}\noindent{{\bf TDC.#1}}}}
\newcommand{\xhdr}[1]{\vspace{2mm}\noindent{{\bf #1.}}}
\newcommand{\hide}[1]{}
\newcommand{\xhdrn}[1]{\noindent{{\bf #1.}}}

\usepackage[p,osf]{cochineal}
\usepackage[scale=.95,type1]{cabin}
\usepackage[cochineal,bigdelims,cmintegrals,vvarbb]{newtxmath}
\usepackage[zerostyle=c,scaled=.94]{newtxtt}
\usepackage[cal=boondoxo]{mathalfa}
\usepackage{microtype}
\usepackage{multirow}
\usepackage{adjustbox}

\usepackage{etoolbox}
\apptocmd{\thebibliography}{\raggedright}{}{}

\usepackage{lmodern}

\newcommand{\std}[1]{\scriptsize{$\pm$#1}}

\usepackage{dirtree}

\makeatletter
\patchcmd{\@maketitle}{\LARGE \@title}{\fontsize{30}{19.2}\selectfont\@title}{}{}
\makeatother

\input{math_commands.tex}


\title{\textbf{Therapeutics Data Commons} \\[1mm] {\fontsize{17}{20}\selectfont\textbf{ Machine Learning Datasets and Tasks for \\[1mm] Drug Discovery and Development}}}

\author[1,*]{\normalsize Kexin Huang}
\author[2,*]{\normalsize Tianfan Fu}
\author[3,*]{\normalsize Wenhao Gao}
\author[4]{\normalsize Yue Zhao}
\author[5]{\normalsize Yusuf Roohani}
\author[5]{\\\normalsize Jure Leskovec}
\author[3]{\normalsize Connor W. Coley}
\author[6]{\normalsize Cao Xiao}
\author[7]{\normalsize Jimeng Sun}
\author[1]{\normalsize Marinka Zitnik}

\affil[1]{\normalsize Harvard University, Boston, MA}
\affil[2]{\normalsize Georgia Institute of Technology, Atlanta, GA}
\affil[3]{\normalsize Massachusetts Institute of Technology, Cambridge, MA}
\affil[4]{\normalsize Carnegie Mellon University, Pittsburgh, PA}
\affil[5]{\normalsize Stanford University, Stanford, CA}
\affil[6]{\normalsize IQVIA, Cambridge, MA}
\affil[7]{\normalsize University of Illinois at Urbana-Champaign, Urbana, IL}
\affil[*]{\normalsize Equal Contribution}

\affil[ ]{}
\affil[ ]{\normalsize \texttt{kexinhuang@hsph.harvard.edu; tfu42@gatech.edu; whgao@mit.edu;}}
\affil[ ]{\normalsize \texttt{zhaoy@cmu.edu; yroohani@stanford.edu; jure@cs.stanford.edu;}}
\affil[ ]{\normalsize \texttt{ccoley@mit.edu; cao.xiao@iqvia.com; jimeng@illinois.edu;}}
\affil[ ]{\normalsize \texttt{marinka@hms.harvard.edu}}
\affil[ ]{}
\affil[ ]{}
\affil[ ]{Correspondence: \href{mailto:contact@tdcommons.ai}{\color{tdc_color}\texttt{contact@tdcommons.ai}\vspace{-8mm}}}
\date{}

\begin{document}
\maketitle

\begin{abstract}
    \input{000abstract}
\end{abstract}

\clearpage

{\footnotesize 
\tableofcontents
\listoftables
}

\input{010introduction}
\input{030relatedworks}
\input{040overview}
\input{050datasets}

\input{060datafunctions}

\input{080library}

\input{070leaderboard}

\input{090conclusion}

\bibliographystyle{agsm}
\bibliography{ref}

\clearpage
\appendix

\end{document}

%% file: math_commands.tex
\usepackage{amsmath,amsfonts,bm}

\def\eqref#1{equation~\ref{#1}}

\DeclareMathAlphabet{\mathsfit}{\encodingdefault}{\sfdefault}{m}{sl}
\SetMathAlphabet{\mathsfit}{bold}{\encodingdefault}{\sfdefault}{bx}{n}

%% file: 000abstract.tex
\normalsize
\noindent 
Therapeutics machine learning is an emerging field with incredible opportunities for innovatiaon and impact. However, advancement in this field requires formulation of meaningful learning tasks and careful curation of datasets. 
Here, we introduce Therapeutics Data Commons (TDC), the first unifying platform to systematically access and evaluate machine learning across the entire range of therapeutics.
To date, TDC includes 66 AI-ready datasets spread across 22 learning tasks and spanning the discovery and development of safe and effective medicines. 
TDC also provides an ecosystem of tools and community resources, including 33 data functions and types of meaningful data splits, 23 strategies for systematic model evaluation, 17 molecule generation oracles, and 29 public leaderboards.
All resources are integrated and accessible via an open Python library.
We carry out extensive experiments on selected datasets, demonstrating that even the strongest algorithms fall short of solving key therapeutics challenges, including real dataset distributional shifts, multi-scale modeling of heterogeneous data, and robust generalization to novel data points.
We envision that TDC can facilitate algorithmic and scientific advances and considerably accelerate machine-learning model development, validation and transition into biomedical and clinical implementation.
TDC is an open-science initiative available at \url{https://tdcommons.ai}. 

%% file: 010introduction.tex
\section{Introduction}

The overarching goal of scientific research is to find ways to cure, prevent, and manage all diseases. With the proliferation of high-throughput biotechnological techniques~\citep{karczewski2018integrative} and advances in the digitization of health information~\citep{abul2019personalized}, machine learning provides a promising approach to expedite the discovery and development of safe and effective treatments. 
Getting a drug to market currently takes 13-15 years and between US\$2 billion and \$3 billion on average, and the costs are going up~\citep{pushpakom2019drug}. Further, the number of drugs approved every year per dollar spent on development has remained flat or decreased for most of the past decade~\citep{pushpakom2019drug,nosengo2016new}. Faced with skyrocketing costs for developing new drugs and long, expensive processes with a high risk of failure, researchers are looking at ways to accelerate all aspects of drug development.
Machine learning has already proved useful in the search of antibiotics~\citep{stokes2020deep}, polypharmacy~\citep{zitnik2018modeling}, drug repurposing for emerging diseases~\citep{gysi2020network}, protein folding and design~\citep{jumper2020high, gao2020deep}, and biomolecular interactions~\citep{zitnik2015gene,agrawal2018large,huang2020skipgnn,gainza2020deciphering}. 

\begin{figure}[t]
    \centering
    \includegraphics[width=\textwidth]{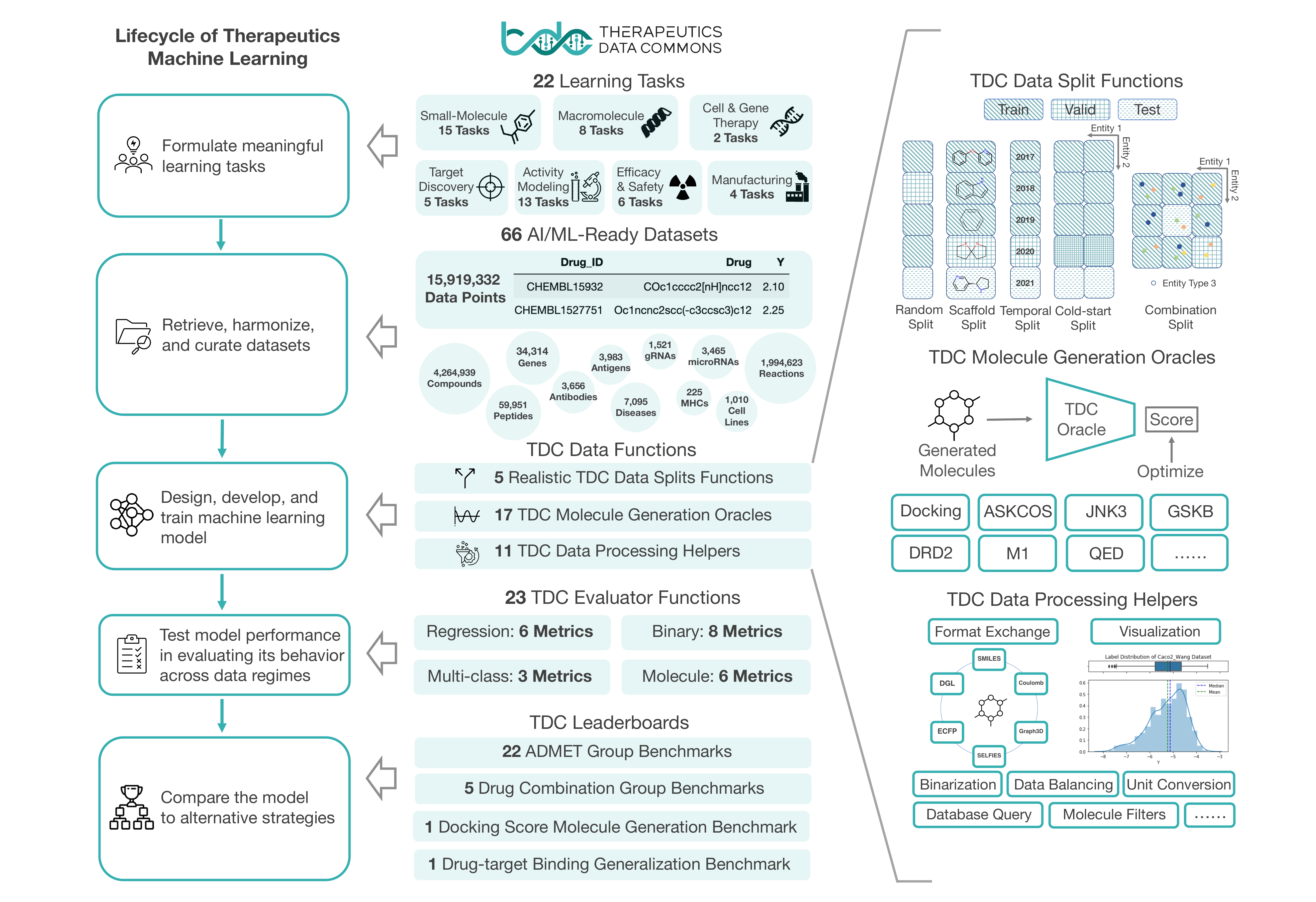}
    \caption{\textbf{Overview of Therapeutics Data Commons (TDC).} TDC is a platform with AI-ready datasets and learning tasks for therapeutics, spanning the discovery and development of safe and effective medicines. TDC provides an ecosystem of tools and data functions, including strategies for systematic model evaluation, meaningful data splits, data processors, and molecule generation oracles. All resources are integrated and accessible via a Python package. TDC also provides community resources with extensive documentation and tutorials, and leaderboards for systematic model comparison and evaluation. }
    \label{fig:overview} 
\end{figure}
Despite the initial success, the attention of the machine learning scientists to therapeutics remains relatively limited, compared to areas such as natural language processing and computer vision, even though therapeutics offer many hard algorithmic problems and applications of immense impact. We posit that is due to the following key challenges: (1) The lack of AI-ready datasets and standardized knowledge representations prevent scientists from formulating relevant therapeutic questions as solvable machine-learning tasks---the challenge is how to computationally operationalize these data to make them amenable to learning; (2) Datasets are of many different types, including experimental readouts, curated annotations and metadata, and are scattered around the biorepositories---the challenge for non-domain experts is how to identify, process, and curate datasets relevant to a task of interest; and (3) Despite promising performance of models, their use in practice, such as for rare diseases and novel drugs in development, is hindered---the challenge is how to assess algorithmic advances in a manner that allows for robust and fair model comparison and represents what one would expect in a real-world deployment or clinical implementation.

\xhdr{Present work}
To address the above challenges, we introduce Therapeutics Data Commons (TDC), a first of its kind platform to systematically access and evaluate machine learning across the entire range of therapeutics (Figure~\ref{fig:overview}). TDC provides AI-ready datasets and learning tasks, together with an ecosystem of tools, libraries, leaderboards, and community resources. To date, TDC contains 66 datasets (Table~\ref{tab:datasets}) spread across 22 learning tasks, 23 strategies for systematic model evaluation and comparison, 17 molecule generation oracles, and 33 data processors, including 5 types of data splits. Datasets in TDC are diverse and cover a range of therapeutic products (\eg, small molecules, biologics, and gene editing) across the entire range of drug development (\ie, target identification, hit discovery, lead optimization, and manufacturing). We develop a Python package that implements all functionality and can efficiently retrieve any TDC dataset. Finally, TDC has 29 leaderboards, each with carefully designed train, validation, and test split to support systematic model comparison and evaluation and test the extent to which model performance indicate utility in the real-world. 

Datasets and tasks in TDC are challenging for prevailing machine learning methods. To this end, we rigorously evaluate 21 domain-specific and state-of-the-art methods across 24 TDC benchmarks (Section~\ref{sec:leaderboard}): (1) a group of 22 ADMET benchmarks are designed to predict properties of small molecules---it is a graph representation learning problem; (2) the DTI-DG benchmark is designed to predict drug-target binding affinity using a patent temporal split---it is a domain generalization problem; (3) the docking benchmark is designed to generate novel molecules with high docking scores in limited resources---it is a low-resource generative modeling problem. We find that theoretic domain-specific methods often have better or comparable performance with state-of-the-art models, indicating urgent need for rigorous model evaluation and an ample opportunity for algorithmic innovation.

Finally, datasets and benchmarks in TDC lend themselves to the study of the following open questions in machine learning and can serve as a testbed for a variety of algorithmic approaches:

\begin{mdframed}[hidealllines=true,backgroundcolor=tdc_color!5]
\begin{itemize}[leftmargin=*,topsep=0pt]
    \setlength\itemsep{0em}
    \item \textbf{Low-resource learning}: Prevailing methods require abundant label information. However, labeled examples are typically scarce in drug development and discovery, considerably limiting the methods' use for problems that require reasoning about new phenomena, such as novel drugs in development, emerging pathogens, and therapies for rare-disease patients.
    \item \textbf{Multi-modal and knowledge graph learning}: Objects in TDC have diverse representations and assume various data modalities, including graphs, tensors/grids, sequences, and spatiotemporal objects.
    \item \textbf{Distribution shifts}: Objects (\eg, compounds, proteins) can change their behavior quickly across biological context (\eg, patients, tissues, cells), meaning that models need accommodate underlying distribution shifts and have robust generalizable performance on previously unseen data points.
    \item \textbf{Causal inference}: TDC contains datasets that quantify response of patients, molecules and cells to different kinds of perturbations, such as treatment, CRISPR gene over-expression, and knockdown perturbations. Observing how and when a cellular, molecular or patient phenotype is altered can provide clues about the underlying mechanisms involved in perturbation and, ultimately, disease. Such datasets represent a natural testbed for causal inference methods.
\end{itemize}
\end{mdframed}

\begin{figure}
    \centering
    \includegraphics[width = 0.8\textwidth]{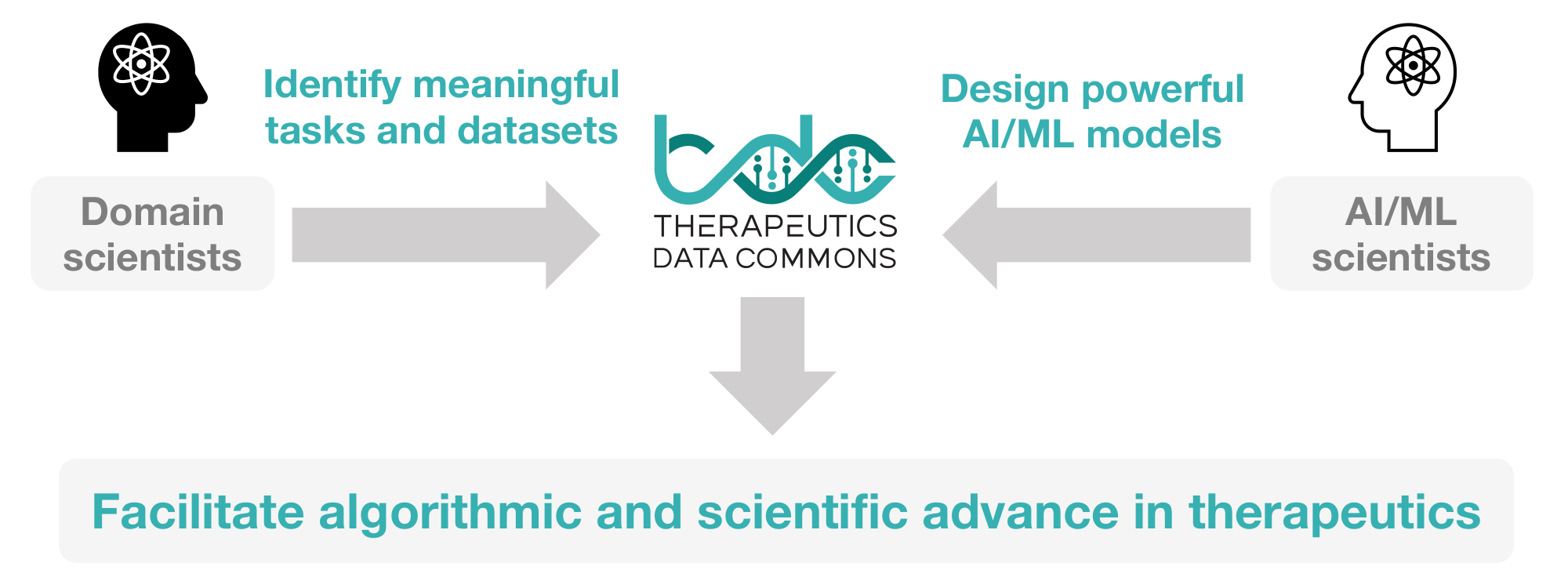}
    \caption{\textbf{Therapeutics Machine Learning.} Therapeutics machine learning offers incredible opportunities for expansion, innovation, and impact. Datasets and benchmarks in \mname provide a systematic model development and evaluation framework. We envision that \mname can considerably accelerate development, validation, and transition of machine learning into production and clinical implementation. }
    \label{fig:vision}
\end{figure}

\xhdr{Facilitating algorithmic and scientific advance in the broad area of therapeutics}
We envision \mname to be the meeting point between domain scientists and ML scientists (Figure~\ref{fig:vision}). Domain scientists can pose learning tasks and identify relevant datasets that are carefully processed and integrated into the TDC and formulated as a scientifically valid learning tasks. ML scientists can then rapidly obtain these tasks and ML-ready datasets through the TDC programming framework and use them to design powerful ML methods. Predictions and other outputs produced by ML models can then facilitate algorithmic and scientific advances in therapeutics. To this end, we strive to make datasets and tasks in TDC representative of real-world therapeutics discovery and development. We further provide realistic data splits, evaluation metrics, and performance leaderboards.

\xhdr{Organization of this manuscript}
This manuscript is organized as follows. We proceed with a brief review of biomedical and chemical data repositories, machine learning benchmarks and infrastructure (Section~\ref{sec:related}). We then give an overview of TDC (Section~\ref{sec:overview}) and describe its tiered structure and modular design (Section~\ref{sec:design}). In Sections~\ref{sec:data-single-pred}-\ref{sec:data-generation}, we provide details for each task in TDC, including the formulation, the level of generalization required for transition into production and clinical implementation, description of therapeutic products and pipeline, and the broader impact of each task. For each task, we also describe a collection of datasets included in TDC. Next, in Sections~\ref{sec:functions}-\ref{sec:library}, we overview TDC's ecosystem of tools, libraries, leaderboards, and community resources. Finally, we conclude with a discussion and directions for future work (Section~\ref{sec:discussion}).

%% file: 030relatedworks.tex
\section{Related Work}\label{sec:related}

TDC is the first unifying platform of datasets and learning tasks for drug discovery and development. We briefly review how TDC relates to data collections, benchmarks, and toolboxes in other areas. 

\xhdr{Relation to biomedical and chemical data repositories} 
There is a myriad of databases with therapeutically relevant information. For example, BindingDB~\citep{bindingdb} curates binding affinity data, ChEMBL~\citep{chembl} curates bioassay data, THPdb~\citep{usmani2017thpdb} and TTD~\citep{wang2020therapeutic} record information on therapeutic targets, and BioSNAP Datasets~\citep{zitnik2018biosnap} contains biological networks. While these biorepositories are important for data deposition and re-use, they do not contain AI-ready datasets (\eg, well-annotated metadata, requisite sample size, and granularity, provenance, multimodal data dynamics, and curation needs), meaning that extensive domain expertise is needed to process the them and construct datasets that can be used for machine learning. 



\xhdr{Relation to ML benchmarks}
Benchmarks have a critical role in facilitating progress in machine learning (\eg, ImageNet~\citep{deng2009imagenet}, Open Graph Benchmark~\citep{hu2020open},  SuperGLUE~\citep{wang2019superglue}). More related to us, MoleculeNet~\citep{molnet} provides datasets for molecular modeling and TAPE~\citep{rao2019evaluating} provides five tasks for protein transfer learning. In contrast, TDC broadly covers modalities relevant to therapeutics, including compounds, proteins, biomolecular interactions, genomic sequences, disease taxonomies, regulatory and clinical datasets. Further, while MoleculeNet and TAPE aim to advance representation learning for compounds and proteins, TDC focuses on drug discovery and development. 

\xhdr{Relation to therapeutics ML tools} 
Many open-science tools exist for biomedical machine learning. Notably, DeepChem~\citep{deepchem} implements models for molecular machine learning; DeepPurpose~\citep{deeppurpose} is a framework for compound and protein modeling; OpenChem~\citep{korshunovaopenchem} and ChemML~\citep{haghighatlari2020chemml} also provide models for drug discovery tasks. In contrast, TDC is not a model-driven framework; instead, it provides datasets and formulates learning tasks. Further, TDC provides tools and resources (Section~\ref{sec:functions}) for model development, evaluation, and comparison.

%% file: 040overview.tex
\section{Overview of \mname}\label{sec:overview}

TDC has three major components: a collection of datasets each with a formulations of a meaningful learning task; a comprehensive set of tools and community resources to support data processing, model development, validation, and evaluation; and a collection of leaderboards to support fair model comparison and benchmarking. The programmatic access is provided through the TDC Python package (Section~\ref{sec:library}). We proceed with a brief overview of each TDC's component. 

\xhdr{1) AI-ready datasets and learning tasks} 
At its core, TDC collects ML tasks and associated datasets spread across therapeutic domains. These tasks and datasets have the following properties:
\begin{itemize}[leftmargin=*,topsep=0pt]
    \setlength\itemsep{0em}
    \item \textit{Instrumenting disease treatment from bench to bedside with ML}: TDC covers a variety of learning tasks going from wet-lab target identification to biomedical product manufacturing.
    \item \textit{Building off the latest biotechnological platforms}: TDC is regularly updated with novel datasets and tasks, such as antibody therapeutics and gene editing.
    \item \textit{Providing ML-ready datasets}: TDC datasets provide rich representations of biomedical entities. The feature information is carefully curated and processed.
\end{itemize}

\xhdr{2) Tools and community resources} 
TDC includes numerous data functions that can be readily used with any TDC dataset. To date, TDC's programmatic functionality can be organized into the following categories:
\begin{itemize}[leftmargin=*,topsep=0pt]
    \setlength\itemsep{0em}
    \item \textbf{23 strategies for model evaluation}: TDC implements a series of metrics and performance functions to debug models, evaluate model performance for any task in TDC, and assess whether model predictions generalize to out-of-distribution datasets.
    \item \textbf{5 types of dataset splits}: TDC implements data splits that reflect real-world learning settings, including random split, scaffold split, cold-start split, temporal split, and combination split.
    \item \textbf{17 molecule generation oracles}: Molecular design tasks require oracle functions to measure the quality of generated entities. TDC implements 17 molecule generation oracles, representing the most comprehensive collection of molecule oracles, each tailored to measure the quality of generated molecules in a specific dimension.
    \item \textbf{11 data processing functions}: Datasets cover a range of modalities, each requiring distinct data processing. TDC provides functions for data format conversion, visualization, binarization, data balancing, unit conversion, database querying, molecule filtering, and more. 
\end{itemize}

\xhdr{3) Leaderboards} 
TDC provides leaderboards for systematic model evaluation and comparison. For a model to be useful for a particular therapeutic question, it needs to perform well across multiple related datasets and tasks. For this reason, we group individual benchmarks in TDC into meaningful groups, which we refer to as \textit{benchmark groups}. Datasets and tasks in a benchmark group are carefully selected and centered around a particular therapeutic question. Dataset splits and evaluation metrics are also carefully selected to indicate challenges of real-world implementation. The current release of TDC has 29 leaderboards (29 = 22 + 5 + 1 + 1; see Figure~\ref{fig:overview}). Section~\ref{sec:leaderboard} describes a subset of 24 selected leaderboards and presents extensive empirical results.

\section{Organization of \mname} \label{sec:design}

Next, we describe the modular design and organization of datasets and learning tasks in TDC.

\subsection{Tiered and Modular Design}\label{sec:tiered-design}

\begin{figure}[t]
    \centering
    \includegraphics[width = 0.7\textwidth]{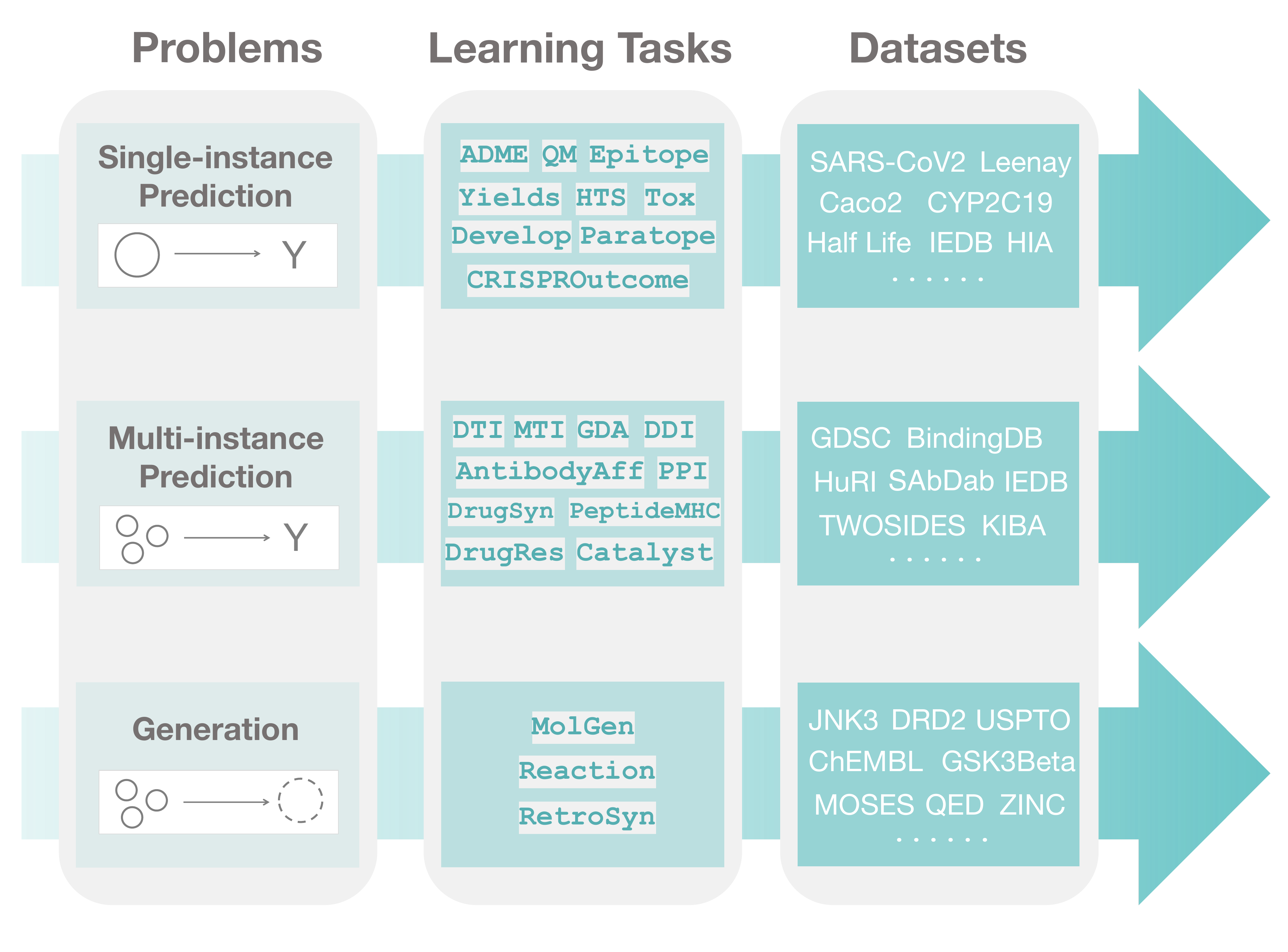}
    \caption{\textbf{Tiered design of Therapeutics Data Commons.} We organize TDC into three distinct \textit{problems}. For each problem, we give a collection of \textit{learning tasks}. Finally, for each task, we provide a collection of \textit{datasets}. In the first tier, we have three broad machine learning problems: (a) single-instance prediction is concerned with predicting properties of individual entities; (b) multi-instance prediction is concerned predicting properties of groups of entities; and (c) generation is concerned with the automatic generation of new entities. For each problem, we have a set of learning tasks. For example, the ADME learning task aims to predict experimental properties of individual compounds; it falls under single-instance prediction. At last, for each task, we have a collection of datasets.  For example, {\color{tdc_color}TDC.Caco2\_Wang} is a dataset under the ADME learning task, which, in turn, is under the single-instance prediction problem. This unique three-tier structure is, to the best of our knowledge, the first attempt at systematically organizing therapeutics ML.}
    \label{fig:design}
\end{figure}

TDC has a unique three-tier hierarchical structure, which to our knowledge, is the first attempt at systematically organizing machine learning for therapeutics (Figure~\ref{fig:design}). We organize TDC into three distinct \textit{problems}. For each problem, we provide a collection of \textit{learning tasks}. Finally, for each task, we provide a series of \textit{datasets}.

In the first tier, we identify three broad machine learning problems:
\begin{mdframed}[hidealllines=true,backgroundcolor=tdc_color!5]
\begin{itemize}[leftmargin=*,topsep=0pt]
    \setlength\itemsep{0em}
    \renewcommand{\thempfootnote}{$\dagger$}
    \item \textbf{Single-instance prediction} \texttt{single\_pred}: Predictions about individual biomedical entities.
    \item \textbf{Multi-instance prediction} \texttt{multi\_pred}: Predictions about multiple biomedical entities.
    \item \textbf{Generation} \texttt{generation}: Generation of biomedical entities with desirable properties.
\end{itemize}
\end{mdframed}

In the second tier, TDC is organized into learning tasks. TDC currently includes 22 learning tasks, covering a range of therapeutic products. The tasks spans small molecules and biologics, including antibodies, peptides, microRNAs, and gene editing. Further, TDC tasks can be mapped to the following drug discovery pipelines:

\begin{mdframed}[hidealllines=true,backgroundcolor=tdc_color!5]
\begin{itemize}[leftmargin=*,topsep=0pt]
    \setlength\itemsep{0em}
    \item \textbf{Target discovery}: Tasks to identify candidate drug targets.
    \item \textbf{Activity modeling}: Tasks to screen and generate individual or combinatorial candidates with high binding activity towards targets.
    \item \textbf{Efficacy and safety}: Tasks to optimize therapeutic signatures indicative of drug safety and efficacy.
    \item \textbf{Manufacturing}: Tasks to synthesize therapeutics.
\end{itemize}
\end{mdframed}

Finally, in the third tier of TDC, each task is instantiated via multiple datasets. For each dataset, we provide several splits of the dataset into training, validation, and test sets to simulate the type of understanding and generalization needed for transition into production and clinical implementation ({\em e.g.,} the model's ability to generalize to entirely unseen compounds or to granularly resolve patient response to a polytherapy).

\begin{table}[t]
\centering
\caption[\textbf{List of 22 learning tasks in Therapeutics Data Commons.}]{\textbf{List of 22 learning tasks in Therapeutics Data Commons.} SM, small molecules; MM, macromolecules; CGT, cell and gene therapy; TD, target discovery; A, bioactivity modeling; ES, efficacy and safety; M, manufacturing. See also Section~\ref{sec:learning tasks}.}
    {\small \begin{tabular}{lc|c@{\hspace{1.5em}}c@{\hspace{1.5em}}c|c@{\hspace{1.5em}}c@{\hspace{1.5em}}c@{\hspace{1.5em}}c}
    \toprule
    \multirow{2}{*}{Learning Task} & \multirow{2}{*}{Section} & \multicolumn{3}{c|}{Therapeutic Products} & \multicolumn{4}{c}{Development Pipelines}\\
     & & SM & MM & CGT & TD & A & ES & M \\\midrule
    \code{single\_pred.ADME} & Sec.~\ref{subsec:adme} & \CheckmarkBold &  &  &  &  & \CheckmarkBold &  \\
    \code{single\_pred.Tox} & Sec.~\ref{subsec:tox} & \CheckmarkBold &  &  &  &  & \CheckmarkBold & \\
    \code{single\_pred.HTS} & Sec.~\ref{subsec:hts} & \CheckmarkBold &  &  &  & \CheckmarkBold & \CheckmarkBold & \\
    \code{single\_pred.QM} & Sec.~\ref{subsec:qm} & \CheckmarkBold &  &  &  & \CheckmarkBold &  & \\
    \code{single\_pred.Yields} & Sec.~\ref{subsec:yields} & \CheckmarkBold &  &  &  &  &  & \CheckmarkBold \\
    \code{single\_pred.Paratope} & Sec.~\ref{subsec:paratope} &  & \CheckmarkBold  &  &  & \CheckmarkBold &  &  \\
    \code{single\_pred.Epitope} & Sec.~\ref{subsec:epitope} &  & \CheckmarkBold &  & \CheckmarkBold & \CheckmarkBold &  &  \\
    \code{single\_pred.Develop} & Sec.~\ref{subsec:develop} & &  \CheckmarkBold &  &  &  & \CheckmarkBold & \\
    \code{single\_pred.CRISPROutcome} & Sec.~\ref{subsec:crispr} &  &  & \CheckmarkBold &  & \CheckmarkBold &  &  \\ \midrule
    \code{multi\_pred.DTI} & Sec.~\ref{subsec:dti} & \CheckmarkBold &  &  & \CheckmarkBold & \CheckmarkBold &  & \\
    \code{multi\_pred.DDI} & Sec.~\ref{subsec:ddi} & \CheckmarkBold &  &  &  &  & \CheckmarkBold & \\
    \code{multi\_pred.PPI} & Sec.~\ref{subsec:ppi} & \CheckmarkBold & \CheckmarkBold &  & \CheckmarkBold & \CheckmarkBold &  & \\
    \code{multi\_pred.GDA} & Sec.~\ref{subsec:gda} & \CheckmarkBold & \CheckmarkBold & \CheckmarkBold & \CheckmarkBold &  &  &  \\
    \code{multi\_pred.DrugRes} & Sec.~\ref{subsec:drugres} & \CheckmarkBold &  &  &  & \CheckmarkBold &  & \\
    \code{multi\_pred.DrugSyn} & Sec.~\ref{subsec:drugsyn} & \CheckmarkBold &  &  &  & \CheckmarkBold &  &  \\
    \code{multi\_pred.PeptideMHC} & Sec.~\ref{subsec:peptidemhc} &  & \CheckmarkBold &  &  & \CheckmarkBold &  & \\
    \code{multi\_pred.AntibodyAff} & Sec.~\ref{subsec:antibodyaff} &  & \CheckmarkBold &  &  & \CheckmarkBold &  & \\
    \code{multi\_pred.MTI} & Sec.~\ref{subsec:mti} &  & \CheckmarkBold &  & \CheckmarkBold & \CheckmarkBold &  & \\
    \code{multi\_pred.Catalyst} & Sec.~\ref{subsec:catalyst} & \CheckmarkBold &  &  &  &  &  & \CheckmarkBold \\ \midrule
    \code{generation.MolGen} & Sec.~\ref{subsec:molgen} & \CheckmarkBold &  &  &  & \CheckmarkBold & \CheckmarkBold & \\
    \code{generation.RetroSyn} & Sec.~\ref{subsec:retrosyn} & \CheckmarkBold &  &  &  &  &  & \CheckmarkBold \\
    \code{generation.Reaction} & Sec.~\ref{subsec:reaction} & \CheckmarkBold &  &  &  &  &  & \CheckmarkBold \\
    \bottomrule
    \end{tabular}}
    \label{tab:learning-tasks}
\end{table}

\subsection{Diverse Learning Tasks}\label{sec:learning tasks}


Table~\ref{tab:learning-tasks} lists 22 learning tasks included in TDC to date. For each task, TDC provides multiple datasets that vary in size between 200 and 2 million data points. We provide the following information for each learning task in TDC:
\begin{mdframed}[hidealllines=true,backgroundcolor=tdc_color!20]
\begin{itemize}[itemsep=0pt,topsep=0pt,label={},leftmargin=*]
\item \xhdrn{Definition} Background and a formal definition of the learning task.
\item \xhdrn{Impact} The broader impact of advancing research on the task.
\item \xhdrn{Generalization} Understanding needed for transition into production and clinical implementation.
\item \xhdrn{Product} The type of therapeutic product examined in the task.
\item \xhdrn{Pipeline} The therapeutics discovery and development pipeline the task belongs to.
\end{itemize}
\end{mdframed}

%
%

\subsection{Machine-Learning Ready Datasets}

Table~\ref{tab:datasets} gives an overview of 66 datasets included in TDC to date. 

\begin{table}[t!]
\centering
\caption[\textbf{List of 66 datasets in Therapeutics Data Commons.}]{\textbf{List of 66 datasets in Therapeutics Data Commons.} Size is the number of data points; Feature is the type of data features; Task is the type of prediction task; Metric is the suggested performance metric; Split is the recommended dataset split. For units, '---' is used when the dataset is either a classification task that does not have units or is a regression task where the numeric label units are not meaningful. For \code{generation.MolGen}, the metrics do not apply as it is defined by the task of interest. }

\adjustbox{max width=\textwidth}{
    
    \begin{tabular}{ll|cccccc}
    \toprule
    Dataset & Learning Task & Size & Unit & Feature & Task & Rec. Metric & Rec. Split\\\midrule
    \dataset{Caco2\_Wang} & \code{single\_pred.ADME} & 906 & cm/s & Seq/Graph & Regression & MAE & Scaffold\\
    \dataset{HIA\_Hou} & \code{single\_pred.ADME} & 578 & --- & Seq/Graph & Binary & AUROC & Scaffold\\
    \dataset{Pgp\_Broccatelli} & \code{single\_pred.ADME} & 1,212 & --- & Seq/Graph & Binary & AUROC & Scaffold\\
    \dataset{Bioavailability\_Ma} & \code{single\_pred.ADME} & 640 & --- & Seq/Graph & Binary & AUROC & Scaffold\\
    \dataset{Lipophilicity\_AstraZeneca} & \code{single\_pred.ADME} & 4,200 & log-ratio & Seq/Graph & Regression & MAE & Scaffold\\
    \dataset{Solubility\_AqSolDB} & \code{single\_pred.ADME} & 9,982 & log-mol/L & Seq/Graph & Regression & MAE & Scaffold\\
    \dataset{BBB\_Martins} & \code{single\_pred.ADME} & 1,975 & --- & Seq/Graph & Binary & AUROC & Scaffold\\
    \dataset{PPBR\_AZ} & \code{single\_pred.ADME} & 1,797 & \% & Seq/Graph & Regression & MAE & Scaffold\\
    \dataset{VDss\_Lombardo} & \code{single\_pred.ADME} & 1,130 & L/kg & Seq/Graph & Regression & Spearman & Scaffold\\
    \dataset{CYP2C19\_Veith} & \code{single\_pred.ADME} & 12,092 & --- & Seq/Graph & Binary & AUPRC & Scaffold\\
    \dataset{CYP2D6\_Veith} & \code{single\_pred.ADME} & 13,130 & ---& Seq/Graph & Binary & AUPRC & Scaffold\\
    \dataset{CYP3A4\_Veith} & \code{single\_pred.ADME} & 12,328 & ---& Seq/Graph & Binary & AUPRC & Scaffold\\
    \dataset{CYP1A2\_Veith} & \code{single\_pred.ADME} & 12,579 & ---& Seq/Graph & Binary & AUPRC & Scaffold\\
    \dataset{CYP2C9\_Veith} & \code{single\_pred.ADME} & 12,092 & ---& Seq/Graph & Binary & AUPRC & Scaffold\\
    \dataset{CYP2C9\_Substrate} & \code{single\_pred.ADME} & 666 & ---& Seq/Graph & Binary & AUPRC & Scaffold\\
    \dataset{CYP2D6\_Substrate} & \code{single\_pred.ADME} & 664 & ---& Seq/Graph & Binary & AUPRC & Scaffold\\
    \dataset{CYP3A4\_Substrate} & \code{single\_pred.ADME} & 667 & ---& Seq/Graph & Binary & AUROC & Scaffold\\
    \dataset{Half\_Life\_Obach} & \code{single\_pred.ADME} & 667 & hr & Seq/Graph & Regression & Spearman & Scaffold\\
    \dataset{Clearance\_Hepatocyte\_AZ} & \code{single\_pred.ADME} & 1,020 & uL.min$^{-1}$.($10^6$cells)$^{-1}$ & Seq/Graph & Regression & Spearman & Scaffold\\
    \dataset{Clearance\_Microsome\_AZ} & \code{single\_pred.ADME} & 1,102 & mL.min$^{-1}$.g$^{-1}$ & Seq/Graph & Regression & Spearman & Scaffold\\
    \dataset{LD50\_Zhu} & \code{single\_pred.Tox} & 7,385 & log(1/(mol/kg)) & Seq/Graph & Regression & MAE & Scaffold\\
    \dataset{hERG} & \code{single\_pred.Tox} & 648 & ---  & Seq/Graph & Binary & AUROC & Scaffold\\
    \dataset{AMES} & \code{single\_pred.Tox} & 7,255 & --- & Seq/Graph & Binary & AUROC & Scaffold\\
    \dataset{DILI} & \code{single\_pred.Tox} & 475 & --- & Seq/Graph & Binary & AUROC & Scaffold\\
    \dataset{Skin\_Reaction} & \code{single\_pred.Tox} & 404 & --- & Seq/Graph & Binary & AUROC & Scaffold\\
    \dataset{Carcinogens\_Lagunin} & \code{single\_pred.Tox} & 278 & --- & Seq/Graph & Binary & AUROC & Scaffold\\
    \dataset{Tox21} & \code{single\_pred.Tox} & 7,831 & --- & Seq/Graph & Binary & AUROC & Scaffold\\
    \dataset{ClinTox} & \code{single\_pred.Tox} & 1,484 & --- & Seq/Graph & Binary & AUROC & Scaffold\\
    \dataset{SARSCoV2\_Vitro\_Touret} & \code{single\_pred.HTS} & 1,480 & --- &  Seq/Graph & Binary & AUPRC & Scaffold\\
    \dataset{SARSCoV2\_3CLPro\_Diamond} & \code{single\_pred.HTS} & 879 & --- & Seq/Graph & Binary & AUPRC & Scaffold\\
    \dataset{HIV} & \code{single\_pred.HTS} & 41,127 & --- & Seq/Graph & Binary & AUPRC & Scaffold\\
    \dataset{QM7b} & \code{single\_pred.QM} & 7,211 & eV/$\AA^3$ & Coulomb & Regression & MAE & Random\\
    \dataset{QM8} & \code{single\_pred.QM} & 21,786 & eV & Coulomb & Regression & MAE & Random\\
    \dataset{QM9} & \code{single\_pred.QM} & 133,885 & GHz/D/$\aa^2_0$/$\aa^3_0$ & Coulomb & Regression & MAE & Random\\
    \dataset{USPTO\_Yields} & \code{single\_pred.Yields} & 853,638 & \% & Seq/Graph & Regression & MAE & Random\\
    \dataset{Buchwald-Hartwig} & \code{single\_pred.Yields} & 55,370 & \% &  Seq/Graph & Regression & MAE & Random\\
    \dataset{SAbDab\_Liberis} & \code{single\_pred.Paratope} & 1,023 & --- & Seq & Token-Binary & Avg-AUROC & Random\\
    \dataset{IEDB\_Jespersen} & \code{single\_pred.Epitope} & 3,159 & --- & Seq & Token-Binary & Avg-AUROC & Random\\
    \dataset{PDB\_Jespersen} & \code{single\_pred.Epitope} & 447 & --- & Seq & Token-Binary & Avg-AUROC & Random\\
    \dataset{TAP} & \code{single\_pred.Develop} & 242 & --- & Seq & Regression & MAE & Random\\
    \dataset{SAbDab\_Chen} & \code{single\_pred.Develop} & 2,409 & --- & Seq & Regression & MAE & Random\\
    \dataset{Leenay} & \code{single\_pred.CRISPROutcome} & 1,521 & \#/\%/bits & Seq & Regression & MAE & Random\\
    \dataset{BindingDB\_Kd} & \code{multi\_pred.DTI} & 52,284 & nM & Seq/Graph & Regression & MAE & Cold-start \\
    \dataset{BindingDB\_IC50} & \code{multi\_pred.DTI} & 991,486 & nM & Seq/Graph & Regression & MAE & Cold-start \\
    \dataset{BindingDB\_Ki} & \code{multi\_pred.DTI} & 375,032 & nM & Seq/Graph & Regression & MAE & Cold-start \\
    \dataset{DAVIS} & \code{multi\_pred.DTI} & 27,621 & nM & Seq/Graph & Regression & MAE & Cold-start \\
    \dataset{KIBA} & \code{multi\_pred.DTI} & 118,036 & --- & Seq/Graph & Regression & MAE & Cold-start \\
    \dataset{DrugBank\_DDI} & \code{multi\_pred.DDI} & 191,808 & --- & Seq/Graph & Multi-class & Macro-F1 & Random \\
    \dataset{TWOSIDES} & \code{multi\_pred.DDI} & 4,649,441 & --- & Seq/Graph & Multi-label & Avg-AUROC & Random \\
    \dataset{HuRI} & \code{multi\_pred.PPI} & 51,813 & --- & Seq & Binary & AUROC & Random \\
    \dataset{DisGeNET} & \code{multi\_pred.GDA} & 52,476 & --- & Numeric/Text & Regression & MAE & Random \\
    \dataset{GDSC1} & \code{multi\_pred.DrugRes} & 177,310 & $\mu$M & Seq/Graph/Numeric & Regression & MAE & Random \\
    \dataset{GDSC2} & \code{multi\_pred.DrugRes} & 92,703 & $\mu$M & Seq/Graph/Numeric & Regression & MAE & Random \\
    \dataset{DrugComb} & \code{multi\_pred.DrugSyn} & 297,098 & --- & Seq/Graph/Numeric & Regression & MAE & Combination \\
    \dataset{OncoPolyPharmacology} & \code{multi\_pred.DrugSyn} & 23,052 & --- & Seq/Graph/Numeric & Regression & MAE & Combination \\
    \dataset{MHC1\_IEDB-IMGT\_Nielsen} & \code{multi\_pred.PeptideMHC} & 185,985 & log-ratio & Seq/Numeric & Regression & MAE & Random \\
    \dataset{MHC2\_IEDB\_Jensen} & \code{multi\_pred.PeptideMHC} & 134,281 & log-ratio & Seq/Numeric & Regression & MAE & Random \\
    \dataset{Protein\_SAbDab} & \code{multi\_pred.AntibodyAff} & 493 & $K_D$(M) & Seq/Numeric & Regression & MAE & Random \\
    \dataset{miRTarBase} & \code{multi\_pred.MTI} & 400,082 & --- & Seq/Numeric & Regression & MAE & Random \\
    \dataset{USPTO\_Catalyst} & \code{multi\_pred.Catalyst} & 721,799 & --- & Seq/Graph & Multi-class & Macro-F1 & Random \\
    \dataset{MOSES} & \code{generation.MolGen} & 1,936,962 & --- & Seq/Graph & Generation & --- & --- \\
    \dataset{ZINC} & \code{generation.MolGen} & 249,455 & --- & Seq/Graph & Generation & --- & --- \\
    \dataset{ChEMBL} & \code{generation.MolGen} & 1,961,462  & --- & Seq/Graph & Generation & --- & --- \\
    \dataset{USPTO-50K} & \code{generation.RetroSyn} & 50,036 & --- & Seq/Graph & Generation & Top-K Acc & Random \\
    \dataset{USPTO\_RetroSyn} & \code{generation.RetroSyn} & 1,939,253 & ---& Seq/Graph & Generation & Top-K Acc & Random \\
    \dataset{USPTO\_Reaction} & \code{generation.Reaction}& 1,939,253 & ---  & Seq/Graph & Generation & Top-K Acc & Random \\
    
    \bottomrule
    \end{tabular}
    }

    \label{tab:datasets}
\end{table}

Next, we give detailed information on learning tasks in Sections~\ref{sec:data-single-pred}-\ref{sec:data-generation}. Following the task description, we briefly describe each dataset for the task. For each dataset, we provide a data description and statistics, together with the recommended dataset splits and evaluation metrics and units in the case of numeric labels.

\captionsetup[table]{list=no}

%% file: 050datasets.tex
\section{Single-Instance Learning Tasks in TDC} \label{sec:data-single-pred}

In this section, we describe single-instance learning tasks and the associated datasets in TDC.

\subsection{\code{single\_pred.ADME}: ADME Property Prediction} \label{subsec:adme}

\begin{mdframed}[hidealllines=true,backgroundcolor=tdc_color!20]
\xhdrn{Definition} A small-molecule drug is a chemical and it needs to travel from the site of administration ({\em e.g.}, oral) to the site of action ({\em e.g.,} a tissue) and then decomposes, exits the body. To do that safely and efficaciously, the chemical is required to have numerous ideal absorption, distribution, metabolism, and excretion (ADME) properties. This task aims to predict various kinds of ADME properties accurately given a drug candidate's structural information.

\xhdrn{Impact} Poor ADME profile is the most prominent reason of failure in clinical trials~\citep{kennedy1997managing}. Thus, an early and accurate ADME profiling during the discovery stage is a necessary condition for successful development of small-molecule candidate.

\xhdrn{Generalization} In real-world discovery, the drug structures of interest evolve over time~\citep{sheridan2013time}. Thus, ADME prediction requires a model to generalize to a set of unseen drugs that are structurally distant to the known drug set. While time information is usually unavailable for many datasets, one way to approximate the similar effect is via scaffold split, where it forces training and test set have distant molecular structures~\citep{bemis1996properties}.

\xhdrn{Product} Small-molecule.

\xhdrn{Pipeline} Efficacy and safety - lead development and optimization.
\end{mdframed}

\subsubsection{Datasets for \code{single\_pred.ADME}}

\dataset{Caco2\_Wang}: The human colon epithelial cancer cell line, Caco-2, is used as an in vitro model to simulate the human intestinal tissue. The experimental result on the rate of drug passing through the Caco-2 cells can approximate the rate at which the drug permeates through the human intestinal tissue~\citep{sambuy2005caco}. This dataset contains experimental values of Caco-2 permeability of 906 drugs~\citep{caco2}. \\ \textit{Suggested data split: scaffold split; Evaluation: MAE; Unit: cm/s.}

\dataset{HIA\_Hou}: When a drug is orally administered, it needs to be absorbed from the human gastrointestinal system into the bloodstream of the human body. This ability of absorption is called human intestinal absorption (HIA) and it is crucial for a drug to be delivered to the target~\citep{wessel1998prediction}. This dataset contains 578 drugs with the HIA index~\citep{hia}. \\ \textit{Suggested data split: scaffold split; Evaluation: AUROC.}

\dataset{Pgp\_Broccatelli}: P-glycoprotein (Pgp) is an ABC transporter protein involved in intestinal absorption, drug metabolism, and brain penetration, and its inhibition can seriously alter a drug's bioavailability and safety~\citep{amin2013p}. In addition, inhibitors of Pgp can be used to overcome multidrug resistance~\citep{shen2013inhibition}. This dataset is from~\cite{pgp} and contains 1,212 drugs with their activities of the Pgp inhibition. \\ \textit{Suggested data split: scaffold split; Evaluation: AUROC.} 

\dataset{Bioavailability\_Ma}: Oral bioavailability is measured by the ability to which the active ingredient in the drug is absorbed to systemic circulation and becomes available at the site of action~\citep{toutain2004bioavailability}. This dataset contains 640 drugs with bioavailability activity from~\cite{bioavail}. \\ \textit{Suggested data split: scaffold split; Evaluation: AUROC.}

\dataset{Lipophilicity\_AstraZeneca}: Lipophilicity measures the ability of a drug to dissolve in a lipid (e.g. fats, oils) environment. High lipophilicity often leads to high rate of metabolism, poor solubility, high turn-over, and low absorption~\citep{waring2010lipophilicity}. This dataset contains 4,200 experimental values of lipophilicity from~\cite{astrazeneca}. We obtained it via MoleculeNet~\citep{molnet}. \\ \textit{Suggested data split: scaffold split; Evaluation: MAE; Unit: log-ratio.}

\dataset{Solubility\_AqSolDB}: Aqeuous solubility measures a drug's ability to dissolve in water. Poor water solubility could lead to slow drug absorptions, inadequate bioavailablity and even induce toxicity. More than 40\% of new chemical entities are not soluble~\citep{savjani2012drug}. This dataset is collected from AqSolDb~\citep{aqsoldb}, which contains 9,982 drugs curated from 9 different publicly available datasets. \\ \textit{Suggested data split: scaffold split; Evaluation: MAE; Unit: log mol/L.}

\dataset{BBB\_Martins}: As a membrane separating circulating blood and brain extracellular fluid, the blood-brain barrier (BBB) is the protection layer that blocks most foreign drugs. Thus the ability of a drug to penetrate the barrier to deliver to the site of action forms a crucial challenge in development of drugs for central nervous system~\citep{abbott2010structure}. This dataset from~\cite{bbb} contains 1,975 drugs with information on drugs' penetration ability. We obtained this dataset from MoleculeNet~\citep{molnet}. \\ \textit{Suggested data split: scaffold split; Evaluation: AUROC.}

\dataset{PPBR\_AZ}: The human plasma protein binding rate (PPBR) is expressed as the percentage of a drug bound to plasma proteins in the blood. This rate strongly affect a drug's efficiency of delivery. The less bound a drug is, the more efficiently it can traverse and diffuse to the site of actions~\citep{lindup1981clinical}. This dataset contains 1,797 drugs with experimental PPBRs~\citep{astrazeneca}. \\ \textit{Suggested data split: scaffold split; Evaluation: MAE; Unit: \% (binding rate).} 

\dataset{VDss\_Lombardo}: The volume of distribution at steady state (VDss) measures the degree of a drug's concentration in body tissue compared to concentration in blood. Higher VD indicates a higher distribution in the tissue and usually indicates the drug with high lipid solubility, low plasma protein binidng rate~\citep{sjostrand1953volume}. This dataset is curated by~\cite{VDss} and contains 1,130 drugs. \\ \textit{Suggested data split: scaffold split; Evaluation: Spearman Coefficient; Unit: L/kg.}

\dataset{CYP2C19\_Veith}: The CYP P450 genes are essential in the breakdown (metabolism) of various molecules and chemicals within cells~\citep{mcdonnell2013basic}. A drug that can inhibit these enzymes would mean poor metabolism to this drug and other drugs, which could lead to drug-drug interactions and adverse effects~\citep{mcdonnell2013basic}. Specifically, the CYP2C19 gene provides instructions for making an enzyme called the endoplasmic reticulum, which is involved in protein processing and transport. This dataset is from~\cite{cypp450}, consisting of 12,665 drugs with their ability to inhibit CYP2C19. \\ \textit{Suggested data split: scaffold split; Evaluation: AUPRC.}

\dataset{CYP2D6\_Veith}: The role and mechanism of general CYP 450 system to metabolism can be found in CYP2C19 Inhibitor. CYP2D6 is responsible for metabolism of around 25\% of clinically used drugs via addition or removal of certain functional groups in the drugs~\citep{teh2011pharmacogenomics}. This dataset is from~\cite{cypp450}, consisting of 13,130 drugs with their ability to inhibit CYP2D6. \\ \textit{Suggested data split: scaffold split; Evaluation: AUPRC.}

\dataset{CYP3A4\_Veith}: The role and mechanism of general CYP 450 system to metabolism can be found in CYP2C19 Inhibitor. CYP3A4 oxidizes the foreign organic molecules and is responsible for metabolism of half of all the prescribed drugs~\citep{zanger2013cytochrome}. This dataset is from~\cite{cypp450}, consisting of 12,328 drugs with their ability to inhibit CYP3A4. \\ \textit{Suggested data split: scaffold split; Evaluation: AUPRC.}

\dataset{CYP1A2\_Veith}: The role and mechanism of general CYP 450 system to metabolism can be found in CYP2C19 Inhibitor. CYP1A2 is induced by some polycyclic aromatic hydrocarbons (PAHs) and it is able to metabolize some PAHs to carcinogenic intermediates. It can also metabolize caffeine, aflatoxin B1, and acetaminophen. This dataset is from~\cite{cypp450}, consisting of 12,579 drugs with their ability to inhibit CYP1A2. \\ \textit{Suggested data split: scaffold split; Evaluation: AUPRC.}

\dataset{CYP2C9\_Veith}: The role and mechanism of general CYP 450 system to metabolism can be found in CYP2C19 Inhibitor. Around 100 drugs are metabolized by CYP2C9 enzymes. This dataset is from~\cite{cypp450}, consisting of 12,092 drugs with their ability to inhibit CYP2C9. \\ \textit{Suggested data split: scaffold split; Evaluation: AUPRC.}

\dataset{CYP2C9\_Substrate\_CarbonMangels}: In contrast to CYP inhibitors where we want to see if a drug can inhibit the CYP enzymes, substrates measure if a drug can be metabolized by CYP enzymes. See CYP2C9 Inhibitor about description of CYP2C9. This dataset is collected from~\cite{cyp_substrate} consisting of 666 drugs experimental values. \\ \textit{Suggested data split: scaffold split; Evaluation: AUPRC.}

\dataset{CYP2D6\_Substrate\_CarbonMangels}: See CYP2C9 Substrate for a description of substrate and see CYP2D6 Inhibitor for CYP2D6 information. This dataset is collected from~\cite{cyp_substrate} consisting of 664 drugs experimental values. \\ \textit{Suggested data split: scaffold split; Evaluation: AUPRC.}

\dataset{CYP3A4\_Substrate\_CarbonMangels}: See CYP2C9 Substrate for a description of substrate and see CYP3A4 Inhibitor for CYP3A4 information. This dataset is collected from~\cite{cyp_substrate} consisting of 667 drugs experimental values.  \\ \textit{Suggested data split: scaffold split; Evaluation: AUROC.}

\dataset{Half\_Life\_Obach}: Half life of a drug is the duration for the concentration of the drug in the body to be reduced by half. It measures the duration of actions of a drug~\citep{benet1995basic}. This dataset is from~\cite{halflife} and it consists of 667 drugs and their half life duration. \\ \textit{Suggested data split: scaffold split; Evaluation: Spearman Coefficient; Unit: hr.}

\dataset{Clearance\_AZ}: Drug clearance is defined as the volume of plasma cleared of a drug over a specified time period and it measures the rate at which the active drug is removed from the body~\citep{toutain2004plasma}. This dataset is from~\cite{astrazeneca} and it contains clearance measures from two experiments types, hepatocyte (\dataset{Clearance\_Hepatocyte\_AZ}) and microsomes (\dataset{Clearance\_Microsome\_AZ}). As studies~\citep{di2012mechanistic} have shown various clearance outcomes given these two different types, we separate them. It has 1,102 drugs for microsome clearance and 1,020 drugs for hepatocyte clearance. \\ \textit{Suggested data split: scaffold split; Evaluation: Spearman Coefficient; Unit: uL.min$^{-1}$.($10^6$cells)$^{-1}$ for Hepatocyte and mL.min$^{-1}$.g$^{-1}$ for Microsome.}

\subsection{\code{single\_pred.Tox}: Toxicity Prediction} \label{subsec:tox}

\begin{mdframed}[hidealllines=true,backgroundcolor=tdc_color!20]
\xhdrn{Definition} Majority of the drugs have some extents of toxicity to the human organisms. This learning task aims to predict accurately various types of toxicity of a drug molecule towards human organisms. 

\xhdrn{Impact} Toxicity is one of the primary causes of compound attrition. Study shows that approximately 70\% of all toxicity-related attrition occurs preclinically (i.e., in cells, animals) while they are strongly predictive of toxicities in humans~\citep{kramer2007application}. This suggests that an early but accurate prediction of toxicity can significantly reduce the compound attribution and boost the likelihood of being marketed. 

\xhdrn{Generalization} Similar to the ADME prediction, as the drug structures of interest evolve over time~\citep{sheridan2013time}, toxicity prediction requires a model to generalize to a set of novel drugs with small structural similarity to the existing drug set. 

\xhdrn{Product} Small-molecule.

\xhdrn{Pipeline} Efficacy and safety - lead development and optimization.
\end{mdframed}

\subsubsection{Datasets for \code{single\_pred.Tox}} 

\dataset{LD50\_Zhu}: Acute toxicity LD50 measures the most conservative dose that can lead to lethal adverse effects. The higher the dose, the more lethal of a drug. This dataset is from~\cite{ld50}, consisting of 7,385 drugs with experimental LD50 values. \\ \textit{Suggested data split: scaffold split; Evaluation: MAE; Unit: log(1/(mol/kg)).}

\dataset{hERG}: Human ether-à-go-go related gene (hERG) is crucial for the coordination of the heart's beating. Thus, if a drug blocks the hERG, it could lead to severe adverse effects. This dataset is from~\cite{herg}, which has 648 drugs and their blocking status. \\ \textit{Suggested data split: scaffold split; Evaluation: AUROC.}

\dataset{AMES}: Mutagenicity means the ability of a drug to induce genetic alterations. Drugs that can cause damage to the DNA can result in cell death or other severe adverse effects. This dataset is from~\cite{ames}, which contains experimental values in Ames mutation assay of 7,255 drugs.  \\ \textit{Suggested data split: scaffold split; Evaluation: AUROC.}

\dataset{DILI}: Drug-induced liver injury (DILI) is fatal liver disease caused by drugs and it has been the single most frequent cause of safety-related drug marketing withdrawals for the past 50 years (e.g. iproniazid, ticrynafen, benoxaprofen)~\citep{assis2009human}. This dataset is aggregated from U.S. FDA’s National Center for Toxicological Research and is collected from~\cite{dili}. It has 475 drugs with labels about their ability to cause liver injury. \\ \textit{Suggested data split: scaffold split; Evaluation: AUROC.}

\dataset{Skin\_Reaction}: Exposure to chemicals on skins can cause reactions, which should be circumvented for dermatology therapeutics products. This dataset from~\cite{skin_reaction} contains 404 drugs with their skin reaction outcome. \\ \textit{Suggested data split: scaffold split; Evaluation: AUROC.}

\dataset{Carcinogens\_Lagunin}: A drug is a carcinogen if it can cause cancer to tissues by damaging the genome or cellular metabolic process. This dataset from~\cite{carcinogen} contains 278 drugs with their abilities to cause cancer.\\ \textit{Suggested data split: scaffold split; Evaluation: AUROC.}

\dataset{Tox21} Tox21 is a data challenge which contains qualitative toxicity measurements for 7,831 compounds on 12 different targets, such as nuclear receptors and stree response pathways~\citep{tox21}. Depending on different assay, we have different number of drugs. They usually range around 6,000 drugs. \\ \textit{Suggested data split: scaffold split; Evaluation: AUROC.}

\dataset{ClinTox}: The clinical toxicity measures if a drug has fail the clinical trials for toxicity reason. It contains 1,484 drugs from clinical trials records~\citep{clintox}. \\ \textit{Suggested data split: scaffold split; Evaluation: AUROC.}

\subsection{\code{single\_pred.HTS}: High-Throughput Screening} \label{subsec:hts}

\begin{mdframed}[hidealllines=true,backgroundcolor=tdc_color!20]
\xhdrn{Definition} High-throughput screening (HTS) is the rapid automated testing of thousands to millions of samples for biological activity at the model organism, cellular, pathway, or molecular level. The assay readout can vary from target binding affinity to fluorescence microscopy of cells treated with drug. HTS can be applied to different kinds of therapeutics however most available data is from testing of small-molecule libraries. In this task, a machine learning model is asked to predict the experimental assay values given a small-molecule compound structure. 

\xhdrn{Impact} High throughput screening is a critical component of small-molecule drug discovery in both industrial and academic research settings. Increasingly more complex assays are now being automated to gain biological insights on compound activity at a large scale. However, there are still limitations on the time and cost for screening a large library that limit experimental throughput. Machine learning models that can predict experimental outcomes can alleviate these effects and save many times and costs by looking at a larger chemical space and narrowing down a small set of highly likely candidates for further smaller-scale HTS. 

\xhdrn{Generalization} The model should be able to generalize over structurally diverse drugs. It is also important for methods to generalize across cell lines. Drug dosage and measurement time points are also very important factors in determining the efficacy of the drug.

\xhdrn{Product} Small-molecule.

\xhdrn{Pipeline} Activity - hit identification.
\end{mdframed}

\subsubsection{Datasets for \code{single\_pred.HTS}}

\dataset{SARSCoV2\_Vitro\_Touret}: An in-vitro screen of the Prestwick chemical library composed of 1,480 approved drugs in an infected cell-based assay. Given the SMILES string for a drug, the task is to predict its activity against SARSCoV2~\citep{sarscov2vitro, mitaicures}. \\ \textit{Suggested data split: scaffold split; Evaluation: AUPRC.}

\dataset{SARSCoV2\_3CLPro\_Diamond}: A large XChem crystallographic fragment screen of 879 drugs against SARS-CoV-2 main protease at high resolution. Given the SMILES string for a drug, the task is to predict its activity against SARSCoV2 3CL Protease~\citep{diamondprotease, mitaicures}. \\ \textit{Suggested data split: scaffold split; Evaluation: AUPRC.}

\dataset{HIV}: The HIV dataset consists of 41,127 drugs and the task is to predict their ability to inhibit HIV replication. It was introduced by the Drug Therapeutics Program AIDS Antiviral Screen~\citep{aids, molnet}. \\ \textit{Suggested data split: scaffold split; Evaluation: AUPRC.}

\subsection{\code{single\_pred.QM}: Quantum Mechanics} \label{subsec:qm}

\begin{mdframed}[hidealllines=true,backgroundcolor=tdc_color!20]
\xhdrn{Definition} The motion of molecules and protein targets can be described accurately with quantum theory, \emph{i.e.}, Quantum Mechanics (QM). However, \emph{ab initio} quantum calculation of many-body system suffers from large computational overhead that is impractical for most applications. Various approximations have been applied to solve energy from electronic structure but all of them have a trade-off between accuracy and computational speed. Machine learning models raise a hope to break this bottleneck by leveraging the knowledge of existing chemical data. This task aims to predict the QM results given a drug's structural information. 

\xhdrn{Impact} A well-trained model can describe the potential energy surface accurately and quickly, so that more accurate and longer simulation of molecular systems are possible. The result of simulation can reveal the biological processes in molecular level and help study the function of protein targets and drug molecules.

\xhdrn{Generalization} A machine learning model trained on a set of QM calculations require to extrapolate to unseen or structurally diverse set of compounds. 

\xhdrn{Product} Small-molecule.

\xhdrn{Pipeline} Activity - lead development.
\end{mdframed}

\subsubsection{Datasets for \code{single\_pred.QM}}

\dataset{QM7b}: QM7 is a subset of GDB-13 (a database of nearly 1 billion stable and synthetically accessible organic molecules) composed of all molecules of up to 23 atoms, where 14 properties (e.g. polarizability, HOMO and LUMO eigenvalues, excitation energies) using different calculation (ZINDO, SCS, PBE0, GW) are provided. This dataset is from~\cite{qm7_1,qm7_2} and contains 7,211 drugs with their 3D coulomb matrix format. \\ \textit{Suggested data split: random split; Evaluation: MAE; Units: eV for energy, $\AA^3$ for polarizability, and intensity is dimensionless. } 

\dataset{QM8}: QM8 consists of electronic spectra and excited state energy of small molecules calculated by multiple quantum mechanic methods. Consisting of low-lying singlet-singlet vertical electronic spectra of over 20,000 synthetically feasible small organic molecules with up to eight CONF atom. This dataset is from~\cite{qm89_1,qm89_2} and contains 21,786 drugs with their 3D coulomb matrix format. \\ \textit{Suggested data split: random split; Evaluation: MAE; Units: eV.}

\dataset{QM9}: QM9 is a dataset of geometric, energetic, electronic, and thermodynamic properties for 134k stable small organic molecules made up of CHONF. The labels consist of geometries minimal in energy, corresponding harmonic frequencies, dipole moments, polarizabilities, along with energies, enthalpies, and free energies of atomization. This dataset is from~\cite{qm89_1,ramakrishnan2014quantum} and contains 133,885 drugs with their 3D coulomb matrix format. \\ \textit{Suggested data split: random split; Evaluation: MAE; Units: GHz for rotational constant, D for dipole moment, $\aa^3_0$ for polarizabily, Ha for energy, $\aa^2_0$ for spatial extent, cal/molK for heat capacity.} 

\subsection{\code{single\_pred.Yields}: Yields Outcome Prediction} \label{subsec:yields}

\begin{mdframed}[hidealllines=true,backgroundcolor=tdc_color!20]
\xhdrn{Definition} Vast majority of small-molecule drugs are synthesized through chemical reactions. Many factors during reactions could lead to suboptimal reactants-products conversion rate, i.e. yields. Formally, it is defined as the percentage of the reactants successfully converted to the target product. This learning task aims to predict the yield of a given single chemical reaction~\citep{schwaller2020prediction}.

\xhdrn{Impact} To maximize the synthesis efficiency of interested products, an accurate prediction of the reaction yield could help chemists to plan ahead and switch to alternate reaction routes, by which avoiding investing hours and materials in wet-lab experiments and reducing the number of attempts.

\xhdrn{Generalization} The models are expected to extrapolate to unseen reactions with diverse chemical structures and reaction types. 

\xhdrn{Product} Small-molecule. 

\xhdrn{Pipeline} Manufacturing - Synthesis planning.
\end{mdframed}

\subsubsection{Datasets for \code{single\_pred.Yields}}

\dataset{USPTO\_Yields}: USPTO dataset is derived from the United States Patent and Trademark Office patent database~\citep{lowe2017chemical} using a refined extraction pipeline from NextMove software. We selected a subset of USPTO that have ``TextMinedYield'' label. It contains 853,638 reactions with reactants and products. \\ \textit{Suggested data split: random split; Evaluation: MAE; Unit: \% (yield rate).}
 
\dataset{Buchwald-Hartwig}: \cite{ahneman2018predicting} performed high-throughput experiments on Pd-catalysed Buchwald–Hartwig C-N cross coupling reactions, measuring the yields for each reaction. This dataset is included as recent study~\citep{schwaller2020prediction} shows USPTO has limited applicability. It contains 55,370 reactions (reactants and products). \\ \textit{Suggested data split: random split; Evaluation: MAE; Unit: \% (yield rate).}

\subsection{\code{single\_pred.Paratope}: Paratope Prediction} \label{subsec:paratope}

\begin{mdframed}[hidealllines=true,backgroundcolor=tdc_color!20]
\xhdrn{Definition}
Antibodies, also known as immunoglobulins, are large, Y-shaped proteins that can identify and neutralize a pathogen's unique molecule, usually called an antigen. They play essential roles in the immune system and are powerful tools in research and diagnostics. A paratope, also called an antigen-binding site, is the region that selectively binds the epitope. Although we roughly know the hypervariable regions that are responsible for binding, it is still challenging to pinpoint the interacting amino acids. This task is to predict which amino acids are in the active position of antibody that can bind to the antigen.

\xhdrn{Impact} 
Identifying the amino acids at critical positions can accelerate the engineering processes of novel antibodies.

\xhdrn{Generalization}
The models are expected to be generalized to unseen antibodies with distinct structures and functions.

\xhdrn{Product}
Antibody.

\xhdrn{Pipeline}
Activity, efficacy and safety.
\end{mdframed}

\subsubsection{Datasets for \code{single\_pred.Paratope}}

\dataset{SAbDab\_Liberis}: \cite{liberis}'s data set is a subset of Structural Antibody Database (SAbDab)~\citep{sabdab} filtered by quality such as resolution and sequence identity. There are in total 1023 antibody chain sequence, covering both heavy and light chains. \\ \textit{Suggested data split: random split; Evaluation: Average-AUROC.}

\subsection{\code{single\_pred.Epitope}: Epitope Prediction} \label{subsec:epitope}

\begin{mdframed}[hidealllines=true,backgroundcolor=tdc_color!20]
\xhdrn{Definition}
An epitope, also known as antigenic determinant, is the region of a pathogen that can be recognized by antibody and cause adaptive immune response. This task is to classify the active and non-active sites from the antigen protein sequences.

\xhdrn{Impact}
Identifying the potential epitope is of primary importance in many clinical and biotechnologies, such as vaccine design and antibody development, and for our general understanding of the immune system.

\xhdrn{Generalization}
The models are expected to be generalized to unseen pathogens antigens amino acid sequences with diverse set of structures and functions.

\xhdrn{Product}
Immunotherapy.

\xhdrn{Pipeline}
Target discovery.
\end{mdframed}

\subsubsection{Datasets for \code{single\_pred.Epitope}}

\dataset{IEDB\_Jespersen}: This dataset collects B-cell epitopes and non-epitope amino acids determined from crystal structures. It is from \cite{jespersen}, curates a dataset from IEDB~\citep{iedb}, containing 3159 antigens. \\ \textit{Suggested data split: random split; Evaluation: Average-AUROC.}

\dataset{PDB\_Jespersen}: This dataset collects B-cell epitopes and non-epitope amino acids determined from crystal structures. It is from \cite{jespersen}, curates a dataset from PDB~\citep{pdb}, containing 447 antigens.\\ \textit{Suggested data split: random split; Evaluation: Average-AUROC.}

\subsection{\code{single\_pred.Develop}: Antibody Developability Prediction} \label{subsec:develop}

\begin{mdframed}[hidealllines=true,backgroundcolor=tdc_color!20]
\xhdrn{Definition}
Immunogenicity, instability, self-association, high viscosity, polyspecificity, or poor expression can all preclude an antibody from becoming a therapeutic. Early identification of these negative characteristics is essential. This task is to predict the developability from the amino acid sequences.

\xhdrn{Impact}
A fast and reliable developability predictor can accelerate the antibody development by reducing wet-lab experiments. They can also alert the chemists to foresee potential efficacy and safety concerns and provide signals for modifications. Previous works have devised accurate developability index based on 3D structures of antibody~\citep{lauer2012developability}. However, 3D information are expensive to acquire. A machine learning that can calculate developability based on sequence information is thus highly ideal. 

\xhdrn{Generalization}
The model is expected to be generalized to unseen classes of antibodies with various structural and functional characteristics.

\xhdrn{Product}
Antibody.

\xhdrn{Pipeline}
Efficacy and safety.
\end{mdframed}

\subsubsection{Datasets for \code{single\_pred.Develop}}

\dataset{TAP}: This data set is from \cite{tap}. Akin to the Lipinski guidelines, which measure druglikeness in small-molecules, Therapeutic Antibody Profiler (TAP) highlights antibodies that possess characteristics that are rare/unseen in clinical-stage mAb therapeutics. In this dataset, TDC includes five metrics measuring developability of an antibody: CDR length, patches of surface hydrophobicity (PSH), patches of positive charge (PPC), patches of negative charge (PNC), structural Fv charge symmetry parameter (SFvCSP). This data set contains 242 antibodies. \\ \textit{Suggested data split: random split; Evaluation: MAE.}

\dataset{SAbDab\_Chen}: This data set is from \cite{sabdab_chen}, containing 2,409 antibodies processed from SAbDab~\citep{sabdab}. The label is calculated through an accurate heuristics algorithm based on antibody's 3D structures, from BIOVIA’s proprietary Pipeline Pilot~\citep{biovia}. \\ \textit{Suggested data split: random split; Evaluation: AUPRC.}

\subsection{\code{single\_pred.CRISPROutcome}: CRISPR Repair Outcome Prediction} \label{subsec:crispr}

\begin{mdframed}[hidealllines=true,backgroundcolor=tdc_color!20]
\xhdrn{Definition} CRISPR-Cas9 is a gene editing technology that allows targeted deletion or modification of specific regions of the DNA within an organism. This is achieved through designing a guide RNA sequence that binds upstream of the target site which is then cleaved through a Cas9-mediated double stranded DNA break. The cell responds by employing DNA repair mechanisms (such as non-homologous end joining) that result in heterogeneous outcomes including gene insertion or deletion mutations (indels) of varying lengths and frequencies. This task aims to predict the repair outcome given a DNA sequence.

\xhdrn{Impact} Gene editing offers a powerful new avenue of research for tackling intractable illnesses that are infeasible to treat using conventional approaches. For example, the FDA recently approved engineering of T-cells using gene editing to treat patients with acute lymphoblastic leukemia~\citep{lim2017principles}. However, since many human genetic variants associated with disease arise from insertions and deletions~\citep{landrum2013rdkit}, it is critical to be able to better predict gene editing outcomes to ensure efficacy and avoid unwanted pathogenic mutations.

\xhdrn{Generalization} \cite{van2016dna} showed that the distribution of Cas9-mediated editing products at a given target site is reproducible and dependent on local sequence context. Thus, it is expected that repair outcomes predicted using well-trained models should be able to generalize across cell lines and reagent delivery methods.

\xhdrn{Product} Cell and gene therapy.

\xhdrn{Pipeline} Efficacy and safety.
\end{mdframed}

\subsubsection{Datasets for \code{single\_pred.CRISPROutcome}}

\dataset{Leenay}: Primary T cells are a promising cell type for therapeutic genome editing, as they can be engineered efficiently ex vivo and then transferred to patients. This dataset consists of the DNA repair outcomes of CRISPR-CAS9 knockout experiments on primary CD4+ T cells drawn from 15 donors~\citep{leenay}. For each of the 1,521 unique genomic locations from 553 genes, the 20-nucleotide guide sequence is provided along with the 3-nucletoide PAM sequence. 5 repair outcomes are included for prediction: fraction of indel reads with an insertion, average insertion length, average deletion length, indel diversity, fraction of repair outcomes with a frameshift. \\ \textit{Suggested data split: random split; Evaluation: MAE; Units: \# for lengths, \% for fractions, bits for diversity.}

\section{Multi-Instance Learning Tasks in TDC} \label{sec:multi-single-pred}

In this section, we describe multi-instance learning tasks and the associated datasets in TDC.

\subsection{\code{multi\_pred.DTI}: Drug-Target Interaction Prediction} \label{subsec:dti}

\begin{mdframed}[hidealllines=true,backgroundcolor=tdc_color!20]
\xhdrn{Definition} The activity of a small-molecule drug is measured by its binding affinity with the target protein. Given a new target protein, the very first step is to screen a set of potential compounds to find their activity. Traditional method to gauge the affinities are through high-throughput screening wet-lab experiments~\citep{hughes2011principles}. However, they are very expensive and are thus restricted by their abilities to search over a large set of candidates. Drug-target interaction prediction task aims to predict the interaction activity score in silico given only the accessible compound structural information and protein amino acid sequence. 

\xhdrn{Impact} Machine learning models that can accurately predict affinities can not only save pharmaceutical research costs on reducing the amount of high-throughput screening, but also to enlarge the search space and avoid missing potential candidates.

\xhdrn{Generalization} Models require extrapolation on unseen compounds, unseen proteins, and unseen compound-protein pairs. Models also are expected to have consistent performance across a diverse set of disease and target groups. 

\xhdrn{Product} Small-molecule.

\xhdrn{Pipeline} Activity - hit identification.
\end{mdframed}

\subsubsection{Datasets for \code{multi\_pred.DTI}}

\dataset{BindingDB}: BindingDB is a public, web-accessible database that aggregates drug-target binding affinities from various sources such as patents, journals, and assays~\citep{bindingdb}. We partitioned the BindingDB dataset into three sub-datasets, each with different units (Kd, IC50, Ki). There are 52,284 pairs for \dataset{BindingDB\_Kd}, 991,486 pairs for \dataset{BindingDB\_IC50}, and 375,032 pairs for \dataset{BindingDB\_Ki}. Alternatively, a negative log10 transformation to pIC50, pKi, pKd can be conducted for easier regression. The current version is 2020m2. \\ \textit{Suggested data split: cold drug split, cold target split; Evaluation: MAE, Pearson Correlation; Unit: nM.}

\dataset{DAVIS}: This dataset is a large-scale assay of DTI of 72 kinase inhibitors with 442 kinases covering >80\% of the human catalytic protein kinome. It is from~\cite{davis} and consists of 27,621 pairs. \\ \textit{Suggested data split: cold drug split, cold target split; Evaluation: MAE, Pearson Correlation; Unit: nM.}

\dataset{KIBA}: As various experimental assays have different units during experiments, \cite{kiba} propose KIBA score to aggregate the IC50, Kd, and Ki scores. This dataset contains KIBA score for 118,036 DTI pairs. \\ \textit{Suggested data split: cold drug split, cold target split; Evaluation: MAE, Pearson Correlation; Unit: dimensionless.}

\subsection{\code{multi\_pred.DDI}: Drug-Drug Interaction Prediction} \label{subsec:ddi}

\begin{mdframed}[hidealllines=true,backgroundcolor=tdc_color!20]
\xhdrn{Definition} Drug-drug interactions occur when two or more drugs interact with each other. These could result in a range of outcomes from reducing the efficacy of one or both drugs to dangerous side effects such as increased blood pressure or drowsiness. Polypharmacy side-effects are associated with drug pairs (or higher-order drug combinations) and cannot be attributed to either individual drug in the pair. This task is to predict the interaction between two drugs.

\xhdrn{Impact} Increasing co-morbidities with age often results in the prescription of multiple drugs simultaneously. Meta analyses of patient records showed that drug-drug interactions were the cause of admission for prolonged hospital stays in 7\% of the cases~\citep{thomsen2007systematic, lazarou1998incidence}. Predicting possible drug-drug interactions before they are prescribed is thus an important step in preventing these adverse outcomes. In addition, as the number of combinations or even higher-order drugs is astronomical, wet-lab experiments or real-world evidence are insufficient. Machine learning can provide an alternative way to inform drug interactions.

\xhdrn{Generalization} As there is a very large space of possible drug-drug interactions that have not been explored, the model needs to extrapolate from known interactions to new drug combinations that have not been prescribed together in the past. Models should also taken into account dosage as that can have a significant impact on the effect of the drugs.

\xhdrn{Product} Small-molecule.

\xhdrn{Pipeline} Efficacy and safety - adverse event detection.
\end{mdframed}

\subsubsection{Datasets for \code{multi\_pred.DDI}}

\dataset{DrugBank\_DDI}: This dataset is manually sourced from FDA and Health Canada drug labels as well as from the primary literature. Given the SMILES strings of two drugs, the goal is to predict their interaction type. It contains 191,808 drug-drug interaction pairs between 1,706 drugs and 86 interaction types~\citep{drugbank}. \\ \textit{Suggested data split: random split; Evaluation: Macro-F1, Micro-F1.}

\dataset{TWOSIDES}: This dataset contains 4,649,441 drug-drug interaction pairs between 645 drugs~\citep{twosides}. Given the SMILES strings of two drugs, the goal is to predict the side effect caused as a result of an interaction. \\ \textit{Suggested data split: random split; Evaluation: Average-AUROC.}

\subsection{\code{multi\_pred.PPI}: Protein-Protein Interaction Prediction} \label{subsec:ppi}

\begin{mdframed}[hidealllines=true,backgroundcolor=tdc_color!20]
\xhdrn{Definition} Proteins are the fundamental function units of human biology. However, they rarely act alone but usually interact with each other to carry out functions. Protein-protein interactions (PPI) are very important to discover new putative therapeutic targets to cure disease~\citep{szklarczyk2015string}. Expensive and time-consuming wet-lab results are usually required to obtain PPI activity. PPI prediction aims to predict the PPI activity given a pair of proteins' amino acid sequences.

\xhdrn{Impact} Vast amounts of human PPIs are unknown and untested. Filling in the missing parts of the PPI network can improve human's understanding of diseases and potential disease target. With the aid of an accurate machine learning model, we can greatly facilitate this process. As protein 3D structure is expensive to acquire, prediction based on sequence data is desirable. 

\xhdrn{Generalization} As the majority of PPIs are unknown, the model needs to extrapolate from a given gold-label training set to a diverse of unseen proteins from various tissues and organisms.

\xhdrn{Product} Small-molecule, macromolecule.

\xhdrn{Pipeline} Basic biomedical research, target discovery, macromolecule discovery.
\end{mdframed}

\subsubsection{Datasets for \code{multi\_pred.PPI}}

\dataset{HuRI}: The human reference map of the human binary protein interactome interrogates all pairwise combinations of human protein-coding genes. This is an ongoing effort and we retrieved the third phase release of the project (HuRI~\citep{huri}), which contains 51,813 positive PPI pairs from 8,248 proteins. \\ \textit{Suggested data split: random split; Evaluation: AUPRC with Negative Samples.}

\subsection{\code{multi\_pred.GDA}: Gene-Disease Association Prediction} \label{subsec:gda}

\begin{mdframed}[hidealllines=true,backgroundcolor=tdc_color!20]
\xhdrn{Definition} Many diseases are driven by genes aberrations. Gene-disease associations (GDA) quantify the relation among a pair of gene and disease. The GDA is usually constructed as a network where we can probe the gene-disease mechanisms by taking into account multiple genes and diseases factors. This task is to predict the association of any gene and disease from both a biochemical modeling and network edge classification perspectives.

\xhdrn{Impact} A high association between a gene and disease could hint at a potential therapeutics target for the disease. Thus, to fill in the vastly incomplete GDA using machine learning accurately could bring numerous therapeutic opportunities. 

\xhdrn{Generalization} Extrapolating to unseen gene and disease pairs with accurate association prediction.

\xhdrn{Product} Any therapeutics.

\xhdrn{Pipeline} Basic biomedical research, target discovery. 
\end{mdframed}

\subsubsection{Datasets for \code{multi\_pred.GDA}}

\dataset{DisGeNET}: DisGeNET integrates gene-disease association data from expert curated repositories, GWAS catalogues, animal models and the scientific literature~\citep{disgenet}. This dataset is the curated subset of DisGeNET. We map disease ID to disease definition and maps Gene ID to amino acid sequence. \\ \textit{Suggested data split: random split; Evaluation: MAE; Unit: dimensionless.}

\subsection{\code{multi\_pred.DrugRes}: Drug Response Prediction} \label{subsec:drugres}

\begin{mdframed}[hidealllines=true,backgroundcolor=tdc_color!20]
\xhdrn{Definition} The same drug compound could have various levels of responses in different patients. To design drug for individual or a group with certain characteristics is the central goal of precision medicine. For example, the same anti-cancer drug could have various responses to different cancer cell lines~\citep{baptista2020deep}. This task aims to predict the drug response rate given a pair of drug and the cell line genomics profile.  

\xhdrn{Impact} The combinations of available drugs and all types of cell line genomics profiles are very large while to test each combination in the wet lab is prohibitively expensive. A machine learning model that can accurately predict a drug's response given various cell lines in silico can thus make the combination search feasible and greatly reduce the burdens on experiments. The fast prediction speed also allows us to screen a large set of drugs to circumvent the potential missing potent drugs.  

\xhdrn{Generalization} A model trained on existing drug cell-line pair should be able to predict accurately on new set of drugs and cell-lines. This requires a model to learn the biochemical knowledge instead of memorizing the training pairs.

\xhdrn{Product} Small-molecule.

\xhdrn{Pipeline} Activity.
\end{mdframed}

\subsubsection{Datasets for \code{multi\_pred.DrugRes}}

\dataset{GDSC}: Genomics in Drug Sensitivity in Cancer (GDSC) is a public database that curates experimental values (IC50) of drug response in various cancer cell lines~\citep{gdsc}. We include two versions of GDSC, with the second one uses improved experimental procedures. The first dataset (\dataset{GDSC1}) contains 177,310 measurements across 958 cancer cells and 208 drugs. The second dataset (\dataset{GDSC2}) contains 92,703 pairs, 805 cell lines, and 137 drugs. \\ \textit{Suggested data split: random split; Evaluation: MAE; Unit: \textit{$\mu$M.}}

\subsection{\code{multi\_pred.DrugSyn}: Drug Synergy Prediction} \label{subsec:drugsyn}

\begin{mdframed}[hidealllines=true,backgroundcolor=tdc_color!20]
\xhdrn{Definition} Synergy is a dimensionless measure of deviation of an observed drug combination response from the expected effect of non-interaction. Synergy can be calculated using different models such as the Bliss model, Highest Single Agent (HSA), Loewe additivity model and Zero Interaction Potency (ZIP). Another relevant metric is CSS which measures the drug combination sensitivity and is derived using relative IC50 values of compounds and the area under their dose-response curves.

\xhdrn{Impact} Drug combination therapy offers enormous potential for expanding the use of existing drugs and in improving their efficacy. For instance, the simultaneous modulation of multiple targets can address the common mechanisms of drug resistance in the treatment of cancers. However, experimentally exploring the entire space of possible drug combinations is not a feasible task. Computational models that can predict the therapeutic potential of drug combinations can thus be immensely valuable in guiding this exploration.

\xhdrn{Generalization} It is important for model predictions to be able to adapt to varying underlying biology as captured through different cell lines drawn from multiple tissues of origin. Dosage is also an important factor that can impact model generalizability. 

\xhdrn{Product} Small-molecule.

\xhdrn{Pipeline} Activity.
\end{mdframed}

\subsubsection{Datasets for \code{multi\_pred.DrugSyn}}

\dataset{DrugComb}: This dataset contains the summarized results of drug combination screening studies for the NCI-60 cancer cell lines (excluding the MDA-N cell line). A total of 129 drugs are tested across 59 cell lines resulting in a total of 297,098 unique drug combination-cell line pairs. For each of the combination drugs, its canonical SMILES string is queried from PubChem~\citep{drugcomb}. For each cell line, the following features are downloaded from NCI’s CellMiner interface: 25,723 gene features capturing transcript expression levels averaged from five microarray platforms, 627 microRNA expression features and 3171 proteomic features that capture the abundance levels of a subset of proteins~\citep{cellminer}. The labels included are CSS and four different synergy scores. \\ \textit{Suggested data split: drug combination split; Evaluation: MAE; Unit: dimensionless.}

\dataset{OncoPolyPharmacology}: A large-scale oncology screen produced by Merck \& Co., where each sample consists of two compounds and a cell line. The dataset covers 583 distinct combinations, each tested against 39 human cancer cell lines derived from 7 different tissue types. Pairwise combinations were constructed from 38 diverse anticancer drugs (14 experimental and 24 approved). The synergy score is calculated by Loewe Additivity values using the batch processing mode of Combenefit. The genomic features are from ArrayExpress database (accession number: E-MTAB-3610), and are quantile-normalized and summarized by \cite{preuer2018deepsynergy} using a factor analysis algorithm for robust microarray summarization (FARMS~\citep{hochreiter2006new}). \\ \textit{Suggested data split: drug combination split; Evaluation: MAE; Unit: dimensionless.}

\subsection{\code{multi\_pred.PeptideMHC}: Peptide-MHC Binding Affinity Prediction} \label{subsec:peptidemhc}

\begin{mdframed}[hidealllines=true,backgroundcolor=tdc_color!20]
\xhdrn{Definition} In the human body, T cells monitor the existing peptides and trigger an immune response if the peptide is foreign. To decide whether or not if the peptide is not foreign, it must bound to a major histocompatibility complex (MHC) molecule. Therefore, predicting peptide-MHC binding affinity is pivotal for determining immunogenicity. There are two classes of MHC molecules: MHC Class I and MHC Class II. They are closely related in overall structure but differ in their subunit composition. This task is to predict the binding affinity between the peptide and the pseudo sequence in contact with the peptide representing MHC molecules.

\xhdrn{Impact}
Identifying the peptide that can bind to MHC can allow us to engineer peptides-based therapeutics such vaccines and cancer-specific peptides. 

\xhdrn{Generalization}
The models are expected to be generalized to unseen peptide-MHC pairs.

\xhdrn{Product}
Immunotherapy.

\xhdrn{Pipeline}
Activity - peptide design.
\end{mdframed}

\subsubsection{Datasets for \code{multi\_pred.PeptideMHC}}

\dataset{MHC1\_IEDB-IMGT\_Nielsen}: This MHC Class I data set has been used in training NetMHCpan-3.0~\citep{nielsen}. The label unit is log-transformed via 1-log(IC50)/log(50,000), where IC50 is in nM units. This data set was collected from the IEDB~\citep{iedb} and consists of 185,985 pairs, covering 43,018 peptides and 150 MHC classes. \\ \textit{Suggested data split: random split; Evaluation: MAE; Unit: log-ratio.}

\dataset{MHC2\_IEDB\_Jensen}: This MHC Class II data set was used to train the NetMHCIIpan~\citep{jensen}. The label unit is log-transformed via 1-log(IC50)/log(50,000), where IC50 is in nM units. This data ste was collected from the IEDB~\citep{iedb} and consists of 134,281 pairs, covering 17,003 peptides and 75 MHC classes.  \\ \textit{Suggested data split: random split; Evaluation: MAE; Unit: log-ratio.}

\subsection{\code{multi\_pred.AntibodyAff}: Antibody-Antigen Binding Affinity Prediction} \label{subsec:antibodyaff}

\begin{mdframed}[hidealllines=true,backgroundcolor=tdc_color!20]
\xhdrn{Definition} Antibodies recognize pathogen antigens and destroy them. The activity is measured by their binding affinities. This task is to predict the affinity from the amino acid sequences of both antigen and antibodies. 

\xhdrn{Impact} Compared to small-molecule drugs, antibodies have numerous ideal properties such as minimal adverse effect and also can bind to many "undruggable" targets due to different biochemical mechanisms. Besides, a reliable affinity predictor can help accelerate the antibody development processes by reducing the amount of wet-lab experiments.

\xhdrn{Generalization}
The models are expected to extrapolate to unseen classes of antigen and antibody pairs.

\xhdrn{Product}
Antibody, immunotherapy.

\xhdrn{Pipeline}
Activity.
\end{mdframed}

\subsubsection{Datasets for \code{multi\_pred.AntibodyAff}}

\dataset{Protein\_SAbDab}: This data set is processed from the SAbDab dataset~\citep{sabdab}, consisting of 493 pairs of antibody-antigen pairs with their affinities. \\ \textit{Suggested data split: random split; Evaluation: MAE; Unit: $K_D$(M).}

\subsection{\code{multi\_pred.MTI}: miRNA-Target Interaction Prediction} \label{subsec:mti}

\begin{mdframed}[hidealllines=true,backgroundcolor=tdc_color!20]
\xhdrn{Definition} MicroRNA (miRNA) is small noncoding RNA that plays an important role in regulating biological processes such as cell proliferation, cell differentiation and so on~\citep{chen2006role}. They usually function to downregulate gene targets. This task is to predict the interaction activity between miRNA and the gene target.

\xhdrn{Impact} Accurately predicting the unknown interaction between miRNA and target can lead to a more complete knowledge about disease mechanism and also could result in potential disease target biomarkers. They can also help identify miRNA hits for miRNA therapeutics candidates~\citep{hanna2019potential}.

\xhdrn{Generalization} The model needs to learn the biochemicals of miRNA and target proteins so that it can extrapolate to new set of novel miRNAs and targets in various disease groups and tissues.

\xhdrn{Product} Small-molecule, miRNA therapeutic.

\xhdrn{Pipeline} Basic biomedical research, target discovery, activity.
\end{mdframed}

\subsubsection{Datasets for \code{multi\_pred.MTI}}

\dataset{miRTarBase}: miRTarBase is a large public database that contains MTIs that are validated experimentally after manually surveying literature related to functional studies of miRNAs~\citep{mirtarbase}. It contains 400,082 MTI pairs with 3,465 miRNAs and 21,242 targets. We use miRBase~\citep{mirbase} to obtain miRNA mature sequence as the feature representation for miRNAs.  \\ \textit{Suggested data split: random split; Evaluation: AUROC.}

\subsection{\code{multi\_pred.Catalyst}: Reaction Catalyst Prediction} \label{subsec:catalyst}

\begin{mdframed}[hidealllines=true,backgroundcolor=tdc_color!20]
\xhdrn{Definition} 
During chemical reaction, catalyst is able to increase the rate of the reaction. Catalysts are not consumed in the catalyzed reaction but can act repeatedly. This learning task aims to predict the catalyst for a reaction given both reactant molecules and product molecules~\citep{zahrt2019prediction}.  

\xhdrn{Impact} 
Conventionally, chemists design and synthesize catalysts by trial and error with chemical intuition, which is usually time-consuming and costly. Machine learning model and automate and accelerate the process, understand the catalytic mechanism, and providing an insight into novel catalytic design~\citep{zahrt2019prediction,coley2019robotic}. 

\xhdrn{Generalization} 
In real-world discovery, as discussed, the molecule structures in reaction of interest evolve over time~\citep{sheridan2013time}. 
We expect model to generalize to the unseen molecules and reaction. 

\xhdrn{Product} Small-molecule.

\xhdrn{Pipeline} Manufacturing - synthesis planning.
\end{mdframed}

\subsubsection{Datasets for \code{multi\_pred.Catalyst}}

\dataset{USPTO\_Catalyst}: USPTO dataset is derived from the United States Patent and Trademark Office patent database~\citep{lowe2017chemical} using a refined extraction pipeline from NextMove software. TDC selects the most common catalysts that have occurences higher than 100 times. It contains 721,799 reactions with 10 reaction types, 712,757 reactants and 702,940 products with 888 common catalyst types.  \\ \textit{Suggested data split: random split; Evaluation: Micro-F1, Macro-F1.}

\section{Generative Learning Tasks in TDC} \label{sec:data-generation}

In this section, we describe generative learning tasks and the associated datasets in TDC.

\subsection{\code{generation.MolGen}: Molecule Generation} \label{subsec:molgen}

\begin{mdframed}[hidealllines=true,backgroundcolor=tdc_color!20]
\xhdrn{Definition} Molecule Generation is to generate diverse, novel molecules that has desirable chemical properties~\citep{molvae,kusner2017grammar,polykovskiy2018molecular,brown2019guacamol}. These properties are measured by oracle functions. A machine learning task first learns the molecular characteristics from a large set of molecules where each is evaluated through the oracles. Then, from the learned distribution, we can obtain novel candidates.

\xhdrn{Impact}
As the entire chemical space is far too large to screen for each target, high through screening can only be restricted to a set of existing molecule library. Many novel drug candidates are thus usually omitted. A machine learning that can generate novel molecule obeying some pre-defined optimal properties can circumvent this problem and obtain novel class of candidates.

\xhdrn{Generalization} The generated molecules have to obtain superior properties given a range of structurally diverse drugs. Besides, the generated molecules have to suffice other basic properties, such as synthesizablility and low off-target effects.

\xhdrn{Product} Small-molecule.

\xhdrn{Pipeline} Efficacy and safety - lead development and optimization, activity - hit identification.
\end{mdframed}

\subsubsection{Datasets for \code{generation.MolGen}}

\dataset{MOSES}: Molecular Sets (MOSES) is a benchmark platform for distribution learning based molecule generation~\citep{polykovskiy2018molecular}. Within this benchmark, MOSES provides a cleaned dataset of molecules that are ideal of optimization. It is processed from the ZINC Clean Leads dataset~\citep{zinc15}.
It contains 1,936,962 molecules.

\dataset{ZINC}: ZINC is a free database of commercially-available compounds for virtual screening. TDC uses a version from the original Mol-VAE paper~\citep{molvae}, which extracted randomly a set of 249,455 molecules from the 2012 version of ZINC~\citep{irwin2012zinc}.

\dataset{ChEMBL}: ChEMBL is a manually curated database of bioactive molecules with drug-like properties~\citep{chembl,chembl_web}. It brings together chemical, bioactivity and genomic data to aid the translation of genomic information into effective new drugs. It contains 1,961,462 molecules.

\subsection{\code{generation.RetroSyn}: Retrosynthesis Prediction} \label{subsec:retrosyn}

\begin{mdframed}[hidealllines=true,backgroundcolor=tdc_color!20]
\xhdrn{Definition} Retrosynthesis is the process of finding a set of reactants that can synthesize a target molecule, i.e., product, which is a fundamental task in drug manufacturing~\citep{liu2017retrosynthetic,zheng2019predicting}. The target is recursively transformed into simpler precursor molecules until commercially available ``starting'' molecules are identified. In a data sample, there is only one product molecule, reactants can be one or multiple molecules. 
Retrosynthesis prediction can be seen as reverse process of Reaction outcome prediction.

\xhdrn{Impact} 
Retrosynthesis planning is useful for chemists to design synthetic routes to target molecules. Computational retrosynthetic analysis tools can potentially greatly assist chemists in designing synthetic routes to novel molecules. Machine learning based methods will significantly save the time and cost. 

\xhdrn{Generalization} The model is expected to accurately generate reactant sets for novel drug candidates with distinct structures from the training set across reaction types with varying reaction conditions.

\xhdrn{Product} Small-molecule.

\xhdrn{Pipeline} Manufacturing - Synthesis planning.
\end{mdframed}

\subsubsection{Datasets for \code{generation.RetroSyn}}

\dataset{USPTO-50K}: USPTO (United States Patent and Trademark Office) 50K consists of 50K extracted atom-mapped reactions with 10 reaction types~\citep{schneider2015development}. It contains 50,036 reactions. \\ \textit{Suggested data split: random split; Evaluation: Top-K accuracy.}

\dataset{USPTO}: USPTO dataset is derived from the United States Patent and Trademark Office patent database~\citep{lowe2017chemical} using a refined extraction pipeline from NextMove software. It contains 1,939,253 reactions.  \\ \textit{Suggested data split: random split; Evaluation: Top-K accuracy.}

\subsection{\code{generation.Reaction}: Reaction Outcome Prediction} \label{subsec:reaction}

\begin{mdframed}[hidealllines=true,backgroundcolor=tdc_color!20]
\xhdrn{Definition} Reaction outcome prediction is to predict the reaction products given a set of reactants~\citep{jin2017predicting}. Reaction outcome prediction can be seen as reverse process of retrosynthesis prediction, as described above. 

\xhdrn{Impact} Predicting the products as a result of a chemical reaction is a fundamental problem in organic chemistry. It is quite challenging for many complex organic reactions. 
Conventional empirical methods that relies on experimentation requires intensive manual label of an experienced chemist, and are always time-consuming and expensive. Reaction Outcome Prediction aims at automating the process. 

\xhdrn{Generalization} The model is expected to accurately generate product for novel set of reactants across reaction types with varying reaction conditions.

\xhdrn{Product} Small-molecule.

\xhdrn{Pipeline} Manufacturing - Synthesis planning.
\end{mdframed}

\subsubsection{Datasets for \code{generation.Reaction}}

\dataset{USPTO}: USPTO dataset is derived from the United States Patent and Trademark Office patent database~\citep{lowe2017chemical} using a refined extraction pipeline from NextMove software. It contains 1,939,253 reactions.  \\ \textit{Suggested data split: random split; Evaluation: Top-K accuracy.}

%% file: 060datafunctions.tex
\section{\mname Data Functions} \label{sec:functions}

TDC implements a comprehensive suite of auxiliary functions frequently used in therapeutics ML. This functionality is wrapped in an easy-to-use interface. Broadly, we provide functions for a) evaluating model performance, b) generating realistic dataset splits, c) constructing oracle generators for molecules, and d) processing, formatting, and mapping of datasets. Next, we describe these functions; note that detailed documentation and examples of usage can be found at \href{https://tdcommons.ai}{\color{tdc_color}https://tdcommons.ai}.

\subsection{Machine Learning Model Evaluation}

To evaluate predictive prowess of ML models built on the TDC datasets, we provide model evaluators. The evaluators implement established performance measures and additional metrics used in biology and chemistry. 
\begin{itemize}[leftmargin=*,topsep=0pt]
 \setlength\itemsep{0em}
 \item \textbf{Regression:} TDC includes common regression metrics, including the mean squared error (MSE), mean absolute error (MAE), coefficient of determination ($R^2$), Pearson's correlation (PCC), and Spearman's correlation (Spearman's $\rho$). 
 \item \textbf{Binary Classification:} TDC includes common metrics, including the area under the receiver operating characteristic curve (AUROC), area under the precision-recall curve (AUPRC), accuracy, precision, recall,  precision at recall of K (PR@K), and recall at precision of K (RP@K).  
 \item \textbf{Multi-Class and Multi-label Classification:} TDC includes Micro-F1, Macro-F1, and Cohen’s Kappa.  
 \item \textbf{Token-Level Classification} conducts binary classification for each token in a sequence. TDC provides Avg-AUROC, which calculates the AUROC score between the sequence of 1/0 true labels and the sequence of predicted labels for every instance. Then, it averages AUROC scores across all instances.
 \item \textbf{Molecule Generation Metrics} evaluate distributional properties of generated molecules. TDC supports the following metrics: 
 \begin{itemize}
  \item \textbf{Diversity} of a set of molecules is defined as average pairwise Tanimoto distance between Morgan fingerprints of the molecules~\citep{benhenda2017chemgan}. 
  \item \textbf{KL divergence} (Kullback-Leibler Divergence) between probability distribution of a particular physicochemical descriptor on the training set and probability distribution of the same descriptor on the set of generated molecules~\citep{brown2019guacamol}. Models that capture distribution of molecules in the training set achieve a small KL divergence score. To increase the diversity of generated molecules, we want high KL divergence scores. 
  \item \textbf{FCD Score} (Fr\'echet ChemNet Distance) first takes the means and covariances of activations of the penultimate layer of ChemNet as calculated for the reference set and for the set of generated molecules~\citep{brown2019guacamol,preuer2018frechet}. The FCD score is then calculated as pairwise Fr\'echet distance between the reference set and the set of generated molecules. Similar molecular distributions are characterized by low FCD values.
  \item \textbf{Novelty} is the fraction of generated molecules that are not present in the training set~\citep{polykovskiy2018molecular}.
  \item \textbf{Validity} is calculated using the RDKit’s molecular structure parser that checks atoms’ valency and consistency of bonds in aromatic rings~\citep{polykovskiy2018molecular}.
  \item \textbf{Uniqueness} measures how often a model generates duplicate molecules~\citep{polykovskiy2018molecular}. When that happens often, the uniqueness score is low.
 \end{itemize}
\end{itemize}

\subsection{Realistic Dataset Splits}

A data split specifies a partitioning of the dataset into training, validation and test sets to train, tune and evaluate ML models. To date, TDC provides the following types of data splits:
\begin{itemize}[leftmargin=*,topsep=0pt]  
  \setlength\itemsep{0em}
  \item \textbf{Random Splits} represent the simplest strategy that can be used with any dataset. The random split selects data instances at random and partitions them into train, validation, and test sets. 
  \item \textbf{Scaffold Splits} partitions molecules into bins based on their Murcko scaffolds~\citep{molnet,yang2019analyzing}. These bins are then assigned to construct structurally diverse train, validation, and test sets. The scaffold split is more challenging than the random split and is also more realistic. 
  \item \textbf{Cold-Start Splits} are implemented for multi-instance prediction problems ({\em e.g.,} DTI, GDA, DrugRes, and MTI tasks that involve predicting properties of heterogeneous tuples consisting of object of different types, such as proteins and drugs). The cold-start split first splits the dataset into train, validation and test set on one entity type ({\em e.g.,} drugs) and then it moves all pairs associated with a given entity in each set to produce the final split. 
  \item \textbf{Combinatorial Splits} are used for combinatorial and polytherapy tasks. This split produces disjoint sets of drug combinations in train, validation, and test sets so that the generalizability of model predictions to unseen drug combinations can be tested.
 \end{itemize}

\subsection{Molecule Generation Oracles}
\label{sec:oracle}

Molecule generation aims to produce novel molecule with desired properties. The extent to which the generated molecules have properties of interest is quantified by a variety of scoring functions, referred to as oracles. To date, TDC provides a wrapper to easily access and process 17 oracles.

Specifically, we include popular oracles from the GuacaMol Benchmark~\citep{brown2019guacamol}, including rediscovery, similarity, median, isomers, scaffold hops, and others. We also include heuristics oracles, including synthetic accessibility (SA) score~\citep{ertl2009estimation}, quantitative estimate of drug-likeness (QED)~\citep{bickerton2012quantifying}, and penalized LogP~\citep{landrum2013rdkit}. A major limitation of {\em de novo} molecule generation oracles is that they focus on overly simplistic oracles mentioned above. As such, the oracles are either too easy to optimize or can produce unrealistic molecules. This issue was pointed out by \cite{coley2020autonomous} who found that current evaluations for generative models do not reflect the complexity of real discovery problems. Because of that, TDC collects novel oracles that are more appropriate for realistic {\em de novo} molecule generation. Next, we describe the details.
\begin{itemize}[leftmargin=*,topsep=0pt]
\item \textbf{Docking Score:} Docking is a theoretical evaluation of affinity ({\em i.e.,} free energy change of the binding process) between a small molecule and a target~\citep{kitchen2004docking}. A docking evaluation usually includes the conformational sampling of the ligand and the calculation of change of free energy. A molecule with higher affinity usually has a higher potential to pose higher bioactivity. Recently, \cite{cieplinski2020we} showed the importance of docking in molecule generation. For this reason, TDC includes a meta oracle for molecular docking where we adopted a Python wrapper from pyscreener~\citep{graff2020accelerating} to allow easy access to various docking software, including AutoDock Vina~\citep{trott2010autodock}, smina~\citep{koes2013lessons}, Quick Vina 2~\citep{alhossary2015fast}, PSOVina~\citep{ng2015psovina}, and DOCK6~\citep{allen2015dock}. 

\item \textbf{ASKCOS:} \cite{gao2020synthesizability} found that surrogate scoring models cannot sufficiently determine the level of difficulty to synthesize a compound. Following this observation, we provide a score derived from the analysis of full retrosynthetic pathway. To this end, TDC leverages ASKCOS~\citep{coley2019robotic}, an open-source framework that integrates efforts to generalize known chemistry to new substrates by applying retrosynthetic transformations, identifying suitable reaction conditions, and evaluating what reactions are likely to be successful. The data-driven models are trained with USPTO and Reaxys databases.

\item \textbf{Molecule.one:} Molecule.one API estimates synthetic accessibility~\citep{liu2020retrognn} of a molecule based on a number of factors, including the number of steps in the predicted synthetic pathway~\citep{sacha2020molecule} and the cost of the starting materials. Currently, the API token can be requested from the Molecule.one website and is provided on a one-to-one basis for research use. We are working with Molecule.one to provide a more open access from within TDC in the near future.

\item \textbf{IBM RXN:} IBM RXN Chemistry is an AI platform that integrates forward reaction prediction and retrosynthetic analysis. The backend of IBM RXN retrosynthetic analysis is a molecular transformer model~\citep{schwaller2019molecular}. The model was trained using USPTO and Pistachio databases. Because of the licensing of the retrosynthetic analysis software, TDC requires the API token as input to the oracle function, along with the input drug SMILES strings.

 \item \textbf{GSK3$\beta$:} Glycogen synthase kinase 3 beta (GSK3$\beta$) is an enzyme in humans that is encoded by GSK3$\beta$ gene. Abnormal regulation and expression of GSK3$\beta$ is associated with an increased susceptibility towards bipolar disorder. The oracle is a random forest classifer using ECFP6 fingerprints using the ExCAPE-DB dataset~\citep{sun2017excape,jin2020multi}. 
 
 \item \textbf{JNK3:} c-Jun N-terminal Kinases-3 (JNK3) belong to the mitogen-activated protein kinase family. The kinases are responsive to stress stimuli, such as cytokines, ultraviolet irradiation, heat shock, and osmotic shock. The oracle is a random forest classifer using ECFP6 fingerprints using the ExCAPE-DB dataset~\citep{sun2017excape,jin2020multi}. 
 
 \item \textbf{DRD2:} DRD2 is a dopamine type 2 receptor. The oracle is constructed by \cite{olivecrona2017molecular} using a support vector machine classifier with a Gaussian kernel and ECFP6 fingerprints on the ExCAPE-DB dataset~\citep{sun2017excape}. 
\end{itemize}

\subsection{Data Processing} 

Finally, TDC supports several utility functions for data processing, such as visualization of label distribution, data binarization, conversion of label units, summary of data statistics, data balancing, graph transformations, negative sampling, and database queries.

\subsubsection{Data Processing Example: Data Formatting} 

Biochemical entities can be represented in various machine learning formats. One of the challenges that hinders machine learning researchers with limited biomedical training is to transform across various formats. TDC provides a \codebox{MolConvert} class that enables format transformation in a few lines of code. Specifically, for 2D molecules, it takes in SMILES, SELFIES~\citep{krenn2020self}, and transform them to molecular graph objects in Deep Graph Library\footnote{\url{https://docs.dgl.ai}}, Pytorch Geometric Library\footnote{\url{https://pytorch-geometric.readthedocs.io}}, and various molecular features such as ECFP2-6, MACCS, Daylight, RDKit2D, Morgan and PubChem. For 3D molecules, it takes in XYZ file, SDF file and transform them to 3D molecular graphs objects, Coulomb matrix and any 2D formats. New formats for more entities will also be included in the future.

%% file: 080library.tex
\section{\mname's Tools, Libraries, and  Resources} \label{sec:library}

TDC has a flexible ecosystem of tools, libraries, and community resources to let researchers push the state-of-the-art in ML and go from model building and training to deployment much more easily.

To boost the accessibility of the project, TDC can be installed through Python Package Index (PyPI) via:

\vspace{-3mm}
\begin{minted}[frame=lines,
framesep=2mm,
baselinestretch=1,
fontsize=\footnotesize
]{bash}
pip install PyTDC
\end{minted}
\vspace{-3mm}

TDC provides a collection of workflows with intuitive, high-level APIs for both beginners and experts to create machine learning models in Python. Building off the modularized ``Problem--Learning Task--Data Set'' structure (see Section~\ref{sec:design}) in TDC, we provide a three-layer API to access any learning task and dataset. This hierarchical API design allows us to easily incorporate new tasks and datasets. 

Suppose you want to retrieve dataset ``DILI'' to study learning task ``Tox'' that belongs to a class of problems ``single\_pred''. To obtain the dataset and its associated data split, use the following:
\vspace{-3mm}
\begin{minted}[frame=lines,
framesep=2mm,
baselinestretch=1,
fontsize=\footnotesize
]{python}
from tdc.single_pred import Tox
data = Tox(name = 'DILI')
df = data.get_data()
\end{minted}
\vspace{-3mm}
The user only needs to specify these three variables and TDC automatically retrieve the processed machine learning-ready dataset from TDC server and generate a \texttt{data} object, which contains numerous utility functions that can be directly applied on the dataset. For example, to get the various training, validation, and test splits, type the following: 
\vspace{-3mm}
\begin{minted}[frame=lines,
framesep=2mm,
baselinestretch=1,
fontsize=\footnotesize
]{python}
from tdc.single_pred import Tox
data = Tox(name = 'DILI')
split = data.get_split(method = 'random', seed = 42, frac = [0.7, 0.1, 0.2])
\end{minted}
\vspace{-3mm}
For other data functions, TDC provides one-liners. For example, to access the ``MSE'' evaluator:
\vspace{-3mm}
\begin{minted}[frame=lines,
framesep=2mm,
baselinestretch=1,
fontsize=\footnotesize
]{python}
from tdc import Evaluator
evaluator = Evaluator(name = 'MSE')
score = evaluator(y_true, y_pred)
\end{minted}
\vspace{-3mm}
To access any of the 17 oracles currently implemented in TDC, specify the oracle name to obtain the oracle function and provide SMILES fingerprints as inputs:
\vspace{-3mm}
\begin{minted}[frame=lines,
framesep=2mm,
baselinestretch=1,
fontsize=\footnotesize
]{python}
from tdc import Oracle
oracle = Oracle(name = 'JNK3')
oracle(['C[C@@H]1CCN(C(=O)CCCc2ccccc2)C[C@@H]1O'])
\end{minted}
\vspace{-3mm}
Further, TDC allows user to access each dataset in a benchmark group (see Section~\ref{sec:overview}). For example, we want to access the ``ADMET\_Group'':
\vspace{-3mm}
\begin{minted}[frame=lines,
framesep=2mm,
baselinestretch=1,
fontsize=\footnotesize
]{python}
from tdc import BenchmarkGroup
group = BenchmarkGroup(name = 'ADMET_Group')
predictions = {}

for benchmark in group:
    name = benchmark['name']
    train_val, test = benchmark['train_val'], benchmark['test']
    ## --- train your model --- ##
    predictions[name] = y_pred

group.evaluate(predictions)
\end{minted}

\xhdr{Documentation, Examples, and Tutorials} 
Comprehensive documentation and examples are provided on the project website\footnote{\url{https://tdcommons.ai}}, along with a set of tutorial Jupyter notebooks\footnote{\url{https://github.com/mims-harvard/TDC/tree/master/tutorials}}.

\xhdr{Project Host, Accessibility, and Collaboration}
To foster development and community collaboration, TDC is publicly host on GitHub\footnote{\url{https://github.com/mims-harvard/TDC}}, where developers leverage source control to track the history of the project and collaborate on bug fix and new functionality development. 

\xhdr{Library Dependency and Compatible Environments} 
TDC is designed for Python 3.5+, and mainly relies on major scientific computing and machine learning libraries including \texttt{numpy}, \texttt{pandas}, and \texttt{scikit-learn}, where additional libraries, such as \texttt{networkx} and \texttt{PyTorch} may be required for specific functionalities. It is tested and designed to work under various operating systems, including \texttt{MacOS}, \texttt{Linux}, and \texttt{Windows}.

\xhdr{Project Sustainability} Many open-source design techniques are leveraged to ensure the robustness and sustainability of TDC. Continuous integration (CI) tools, including \textit{Travis-CI}\footnote{\url{https://travis-ci.org/github/mims-harvard/TDC}} and \textit{CircleCI}\footnote{\url{https://app.circleci.com/pipelines/github/mims-harvard/TDC}}, are enabled for conducting daily test execution. All branches are actively monitored by the CI tools, and all commits and pull requests are covered by unit test. For quality assurance, TDC follows \texttt{PEP8} standard, and we follow the Python programming guidelines for maintainbility.

%% file: 070leaderboard.tex
\section{TDC Leaderboards and Experiments on Selected Datasets} \label{sec:leaderboard}

TDC benchmarks and leaderboards enable systematic model development and evaluation. We illustrate them through three examples. All datasets, code, and evaluation procedures to reproduce these experiments are accessible from \url{https://github.com/mims-harvard/TDC/tree/master/examples}.

\subsection{Twenty-Two Datasets in the ADMET Benchmark Group}

\xhdr{Motivation} A small-molecule drug needs to travel from the site of administration (e.g., oral) to the site of action (e.g., a tissue) and then decomposes, exits the body. Therefore, the chemical is required to have numerous ideal absorption, distribution, metabolism, excretion, and toxicity (ADMET) properties~\citep{van2003admet}. 
Thus, an early and accurate ADMET profiling during the discovery stage is an essential condition for the successful development of a small-molecule candidate. 
An accurate ML model that can predict various ADMET endpoints are thus highly sought-after.

\xhdr{Experimental setup} We leverage 22 ADMET datasets included in TDC -- the largest public ADMET benchmark. The included endpoints are widely used in the pharmaceutical companies, such as metabolism with various CYP enzymes, half-life, clearance, and off-target effects. In real-world discovery, the drug structures of interest evolve. Thus, ADMET prediction requires a model to generalize to a set of unseen drugs that are structurally distant to the known drug set. We adopt scaffold split to simulate this distant effect. Data are split into 7:1:2 train:validation:test where train and validation set are shuffled five times to create five random runs. For binary classification, AUROC is used for balanced data and AUPRC when the number of positives are smaller than negatives and for regression task, MAE is used and Spearman correlation for benchmarks where a trend is more important than the absolute error. 

\xhdr{Baselines} 
The focus in this task is representation learning. We include (1) multi-layer perceptron (MLP) with expert-curated fingerprint (Morgan fingerprint~\citep{morgan} with 1,024 bits) or descriptor (RDKit2D~\citep{landrum2013rdkit}, 200-dim); (2) convolutional neural network (CNN) on SMILES strings, which applies 1D convolution over a string representation of the molecule~\citep{deeppurpose}; (3) state-of-the-art (SOTA) ML models use graph neural network based models on molecular 2D graphs, including neural fingerprint (NeuralFP)~\citep{neuralfp}, graph convolutional network (GCN)~\citep{kipf2016semi}, and attentive fingerprint (AttentiveFP)~\citep{attentivefp}, three powerful Graph neural network (GNN) models. In addition, recently, \citep{pretrain_gnn} has adapted a pretraining strategy to molecule graph, where we include two strategies attribute masking (AttMasking) and context prediction (ContextPred). Methods follow the default hyperparameters described in the original papers.

\xhdr{Results} 
Results are shown in Table~\ref{tab:admet-results}. Overall, we find that pretraining GIN (Graph Isomorphism Network)~\citep{xu2018powerful} with context prediction has the best performances in 8 endpoints, attribute masking has the best ones in 5 endpoints, with 13 combined for pretraining strategies, especially in CYP enzyme predictions. Expert-curated descriptor RDKit2D also has five endpoints that achieve the best results, while SMILES-based CNN has one best-performing one. Our systematic evaluation yield three key findings. First, the ML SOTA models do not work well consistently for these novel realistic endpoints. In some cases, methods based on learned features are worse than the efficient domain features. This gap highlights the necessity for realistic benchmarking Second, performances vary across feature types given different endpoints. For example, in TDC.CYP3A4-S, the SMILES-based CNN is 8.7\%-14.9\% better than the graph-based methods. We suspect this is due to that different feature types contain different signals (e.g. GNN focuses on a local aggregation of substructures whereas descriptors are global biochemical features). Thus, future integration of these signals could potentially improve the performance. Third, the best performing methods use pretraining strategies, highlighting an exciting avenue in recent advances in self-supervised learning to the biomedical setting. 

\begin{table}[h]
    \centering
    \caption{\textbf{Leaderboard on the TDC ADMET Benchmark Group.} Average and standard deviation across five runs are reported. Arrows ($\uparrow$, $\downarrow$) indicate the direction of better performance. The best method is bolded and the second best is underlined. 
    }
    \adjustbox{max width=\textwidth}{

    \begin{tabular}{l|l|cc|c|ccccc}
    \midrule
    \multicolumn{2}{c|}{Raw Feature Type} & \multicolumn{2}{c|}{\cellcolor{dc_color} Expert-Curated Methods} & \multicolumn{1}{c|}{\cellcolor{ds_color} SMILES} & \multicolumn{5}{c}{\cellcolor{ms_color} Molecular Graph-Based Methods (state-of-the-Art in ML)} \\ 
    \midrule
    \multirow{2}{*}{Dataset} & Metric & Morgan & RDKit2D & CNN & NeuralFP & GCN & AttentiveFP & AttrMasking & ContextPred \\ \cmidrule{2-10}
    & \# Params. & 1477K & 633K & 227K & 480K & 192K & 301K & 2067K & 2067K \\ \midrule

    \dataset{Caco2} ($\downarrow$) & MAE & 0.908\std{0.060} & \bf 0.393\std{0.024} & 0.446\std{0.036} & 0.530\std{0.102} & 0.599\std{0.104} & \underline{0.401\std{0.032}} & 0.546\std{0.052} & 0.502\std{0.036} \\ 
    
    \dataset{HIA} ($\uparrow$) & AUROC & 0.807\std{0.072} & 0.972\std{0.008} & 0.869\std{0.026} & 0.943\std{0.014} & 0.936\std{0.024} & 0.974\std{0.007} &  \bf 0.978\std{0.006} & \underline{0.975\std{0.004}} \\
    
    \dataset{Pgp} ($\uparrow$) & AUROC & 0.880\std{0.006} & 0.918\std{0.007} & 0.908\std{0.012} & 0.902\std{0.020} & 0.895\std{0.021} & 0.892\std{0.012} & \bf 0.929\std{0.006} & \underline{0.923\std{0.005}} \\ 
    
    \dataset{Bioav} ($\uparrow$) & AUROC & 0.581\std{0.086} & \bf 0.672\std{0.021} & 0.613\std{0.013} & 0.632\std{0.036} & 0.566\std{0.115} & 0.632\std{0.039} & 0.577\std{0.087} & \underline{0.671\std{0.026}}\\
    
    \dataset{Lipo} ($\downarrow$)  & MAE & 0.701\std{0.009} & 0.574\std{0.017} & 0.743\std{0.020} & 0.563\std{0.023} & \underline{ 0.541\std{0.011}} &  0.572\std{0.007} &0.547\std{0.024} & \bf 0.535\std{0.012}\\ 
    
    \dataset{AqSol} ($\downarrow$) & MAE & 1.203\std{0.019}  & \underline{0.827\std{0.047}} & 1.023\std{0.023} & 0.947\std{0.016} &  0.907\std{0.020} & \bf 0.776\std{0.008} &1.026\std{0.020} & 1.040\std{0.045} \\ \midrule
    
    \dataset{BBB} ($\uparrow$) & AUROC & 0.823\std{0.015} & 0.889\std{0.016} & 0.781\std{0.030} & 0.836\std{0.009} & 0.842\std{0.016} & 0.855\std{0.011} & \underline{0.892\std{0.012}} &  \bf 0.897\std{0.004} \\
    
    \dataset{PPBR} ($\downarrow$) & MAE & 12.848\std{0.362} & 9.994\std{0.319} & 11.106\std{0.358} & \bf 9.292\std{0.384} & 10.194\std{0.373} & \underline{9.373\std{0.335}} & 10.075\std{0.202} & 9.445\std{0.224} \\
    
    \dataset{VD} ($\uparrow$) & Spearman & 0.493\std{0.011}& \bf 0.561\std{0.025} & 0.226\std{0.114} & 0.258\std{0.162} & 0.457\std{0.050} & 0.241\std{0.145} & \underline{0.559\std{0.019}} & 0.485\std{0.092} \\ \midrule
    
    \dataset{CYP2D6-I} ($\uparrow$) & AUPRC & 0.587\std{0.011} & 0.616\std{0.007} & 0.544\std{0.053} & 0.627\std{0.009} & 0.616\std{0.020} & 0.646\std{0.014} & \underline{0.721\std{0.009}} & \bf 0.739\std{0.005}\\
    
    \dataset{CYP3A4-I} ($\uparrow$) & AUPRC & 0.827\std{0.009} & 0.829\std{0.007} & 0.821\std{0.003} & 0.849\std{0.004} & 0.840\std{0.010} & 0.851\std{0.006} & \underline{0.902\std{0.002}} & \bf 0.904\std{0.002}\\
    
    \dataset{CYP2C9-I} ($\uparrow$) & AUPRC & 0.715\std{0.004} & 0.742\std{0.006}  & 0.713\std{0.006} & 0.739\std{0.010} &0.735\std{0.004} & 0.749\std{0.004} & \underline{0.829\std{0.003}} & \bf 0.839\std{0.003}\\
    
    \dataset{CYP2D6-S} ($\uparrow$) & AUPRC & 0.671\std{0.066} & 0.677\std{0.047}  & 0.485\std{0.037} & 0.572\std{0.062} & 0.617\std{0.039} & 0.574\std{0.030} & \underline{0.704\std{0.028}} & \bf 0.736\std{0.024}\\
    
    \dataset{CYP3A4-S} ($\uparrow$) & AUROC &  0.633\std{0.013} & \underline{0.639\std{0.012}} & \bf 0.662\std{0.031} & 0.578\std{0.020} & 0.590\std{0.023} & 0.576\std{0.025} & 0.582\std{0.021} & 0.609\std{0.025}\\
    
    \dataset{CYP2C9-S} ($\uparrow$) & AUPRC & 0.380\std{0.015} & 0.360\std{0.040} & 0.367\std{0.059} & 0.359\std{0.059} & 0.344\std{0.051} & 0.375\std{0.032} & \underline{0.381\std{0.045}} & \bf 0.392\std{0.026}\\ \midrule
    
    \dataset{Half\_Life} ($\uparrow$) & Spearman & \bf 0.329\std{0.083} & 0.184\std{0.111} & 0.038\std{0.138} & 0.177\std{0.165} & \underline{0.239\std{0.100}} & 0.085\std{0.068} & 0.151\std{0.068} & 0.129\std{0.114}\\
    
    \dataset{CL-Micro} ($\uparrow$) & Spearman & 0.492\std{0.020} & \bf 0.586\std{0.014} & 0.252\std{0.116} & 0.529\std{0.015} & 0.532\std{0.033}& 0.365\std{0.055} & \underline{0.585\std{0.034}} & 0.578\std{0.007}\\
    
    \dataset{CL-Hepa} ($\uparrow$) & Spearman& 0.272\std{0.068} & 0.382\std{0.007} & 0.235\std{0.021} & 0.401\std{0.037} & 0.366\std{0.063} & 0.289\std{0.022} & \underline{0.413\std{0.028}} & \bf 0.439\std{0.026} \\ \midrule
    
    \dataset{hERG} ($\uparrow$) & AUROC & 0.736\std{0.023} & \bf 0.841\std{0.020} & 0.754\std{0.037} & 0.722\std{0.034} & 0.738\std{0.038} & \underline{0.825\std{0.007}} & 0.778\std{0.046} & 0.756\std{0.023}\\
    
    \dataset{AMES} ($\uparrow$)  & AUROC & 0.794\std{0.008} & 0.823\std{0.011} & 0.776\std{0.015} & 0.823\std{0.006} & 0.818\std{0.010} & 0.814\std{0.008} & \bf 0.842\std{0.008} & \underline{0.837\std{0.009}}\\
    
    \dataset{DILI} ($\uparrow$) & AUROC & 0.832\std{0.021} & 0.875\std{0.019} & 0.792\std{0.016} & 0.851\std{0.026} & 0.859\std{0.033} & \underline{0.886\std{0.015}} & \bf 0.919\std{0.008} & 0.861\std{0.018}\\ 
    
    \dataset{LD50} ($\downarrow$) & MAE & 0.649\std{0.019} & \underline{0.678\std{0.003}} & 0.675\std{0.011} & 0.667\std{0.020} & 0.649\std{0.026} & 0.678\std{0.012} & \bf 0.685\std{0.025} & 0.669\std{0.030}\\ \bottomrule
    \end{tabular}
    }
    \label{tab:admet-results}
\end{table}

\subsection{Domain Generalization in the Drug-target Interaction Benchmark}

\xhdr{Motivation}
Drug-target interactions (DTI) characterize the binding of compounds to disease targets. Identifying high-affinity compounds is the first crucial step for drug discovery. Recent ML models have shown strong performances in DTI prediction~\citep{deeppurpose}, but they adopt a random dataset splitting where testing sets contain unseen pair of compound-target, but both of the compounds and targets are seen. However, pharmaceutical companies develop compound screening campaigns for novel targets or screen a novel class of compounds for known targets. These novel compounds and targets shift over the years. Thus, it requires a DTI ML model to achieve consistent performances to the subtle domain shifts along the temporal dimension.
Recently, numerous domain generalization methods have been developed in the context of images and languages~\citep{koh2020wilds} but merely in biomedical space. 

\xhdr{Experimental setup} In this benchmark, we use DTIs in \dataset{BindingDB} that have patent information. Specifically, we formulate each domain consisting of DTIs that are patented in a specific year. We test various domain generalization methods to predict out-of-distribution DTIs in 2019-2021 after training on 2013-2018 DTIs, simulating the realistic scenario. Note that time information for specific targets and compounds are usually private data. Thus, we solicit the patent year of the DTI as a reasonable proxy to simulate this realistic challenge. We use the popular deep learning based DTI model DeepDTA~\citep{ozturk2018deepdta} as the backbone of any domain generalization algorithms. The evaluation metric is pearson correlation coefficient (PCC). Validation set selection is crucial for a fair domain generalization methods comparison. Following the strategy of "Training-domain validation set" in ~\cite{gulrajani2020search}, from the 2013-2018 DTIs, we randomly set 20\% of them as the validation set and use them as the in-distribution performance calculations as they follow the same as the training set and 2018-2021 are only used during testing, which we called \textit{out-of-distribution}. 

\xhdr{Baselines} 
%
ERM (Empirical Risk Minimization)~\citep{erm} is the standard training strategy where errors across all domains and data are minimized. We then include various types of SOTA domain generalization algorithms: MMD (Maximum Mean Discrepancy)~\citep{mmd} optimizes the similarities of maximum mean discrepancy across domains, CORAL (Correlation Alignment)~\citep{coral} matches the mean and covariance of features across domains; IRM (Invariant Risk Minimization)~\citep{irm} obtains features where a linear classifier is optimal across domains; GroupDRO (distributionally robust neural networks for group shifts)~\citep{dro} optimizes ERM and adjusts the weights of domains with larger errors; MTL (Marginal Transfer Learning)~\citep{mtl} concatenates the original features with an augmented vector using the marginal distribution of feature vectors, which practically is the mean of the feature embedding; ANDMask~\citep{andmask} masks gradients that have inconsistent signs in the corresponding weights across domains. Note that majority of the methods are developed for classification tasks, we modify the objective functions to regression and keep the rest the same. Methods follow the default hyperparameters described in the paper.

\begin{figure}
\begin{minipage}{\textwidth}
\begin{minipage}[b]{0.62\textwidth}

\centering
    \includegraphics[width=\textwidth]{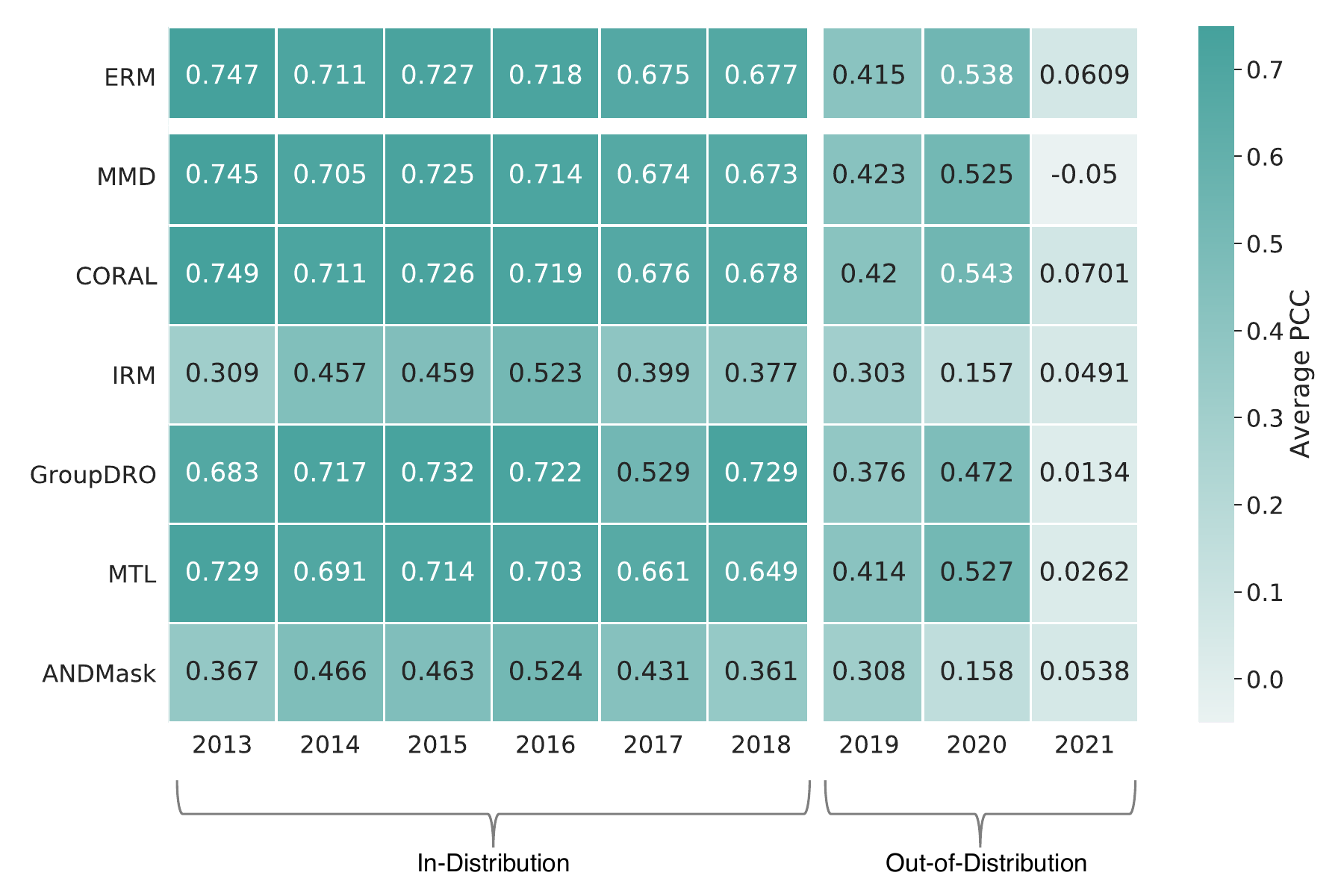}
    \captionof{figure}{\textbf{Heatmap visualization of domain generalization methods performance across each domain in the TDC DTI-DG benchmark using \dataset{BindingDB}.} We observe a significant gap between the in-distribution and out-of-distribution performance and highlight the demand for algorithmic innovation. } \label{fig:dg_heatmap}
    
\end{minipage}
\hfill
\begin{minipage}[b]{0.32\textwidth}
    
    \centering
    \captionof{table}{\textbf{Leaderboard on TDC DTI-DG benchmark using \dataset{BindingDB}}. \textit{In-Dist.} aggregates the in-split validation set and follows the same data distribution as the training set (2013-2018). \textit{Out-Dist.} aggregates the testing domains (2019-2021). The goal is to maximize the test domain performance. Reported results include the average and standard deviation of Pearson Correlation Coefficient across five random runs. The best method is bolded and the second best is underlined. }
    
    \begin{adjustbox}{width = \textwidth}
    \begin{tabular}[b]{l|c|c}
    \toprule
    Method & In-Dist. & Out-Dist. \\ \midrule
    ERM & \underline{0.703\std{0.005}} & 0.427\std{0.012} \\ \midrule
    MMD & 0.700\std{0.002} & \bf 0.433\std{0.010} \\ 
    CORAL & \bf 0.704\std{0.003} & \underline{0.432\std{0.010}} \\ 
    IRM & 0.420\std{0.008} & 0.284\std{0.021} \\ 
    GroupDRO & 0.681\std{0.010} & 0.384\std{0.006} \\ 
    MTL & 0.685\std{0.009} & 0.425\std{0.010} \\ 
    ANDMask & 0.436\std{0.014} & 0.288\std{0.019} \\ 
    \bottomrule
    \end{tabular}
    
    \label{tab:dg_agg}
    \end{adjustbox}
    
  \end{minipage}
  \end{minipage}
\end{figure}

\xhdr{Results}
Results are shown in Table~\ref{tab:dg_agg} and Figure~\ref{fig:dg_heatmap}. We observe that in-distribution reaches ~0.7 PCC and are very stable across the years, suggesting the high predictive power of ML models in the unrealistic but widely adopted ML settings. However, out-of-distribution performance significantly degrades from 33.9\% to 43.6\% across methods, suggesting that domain shift exists and realistic constraint breaks usual training strategies. Second, although the best performed methods are MMD and CORAL, the standard training strategy has similar performances as current ML SOTA domain generalization algorithms, which confirms with the systematic study conducted by~\cite{gulrajani2020search}, highlighting a demand for robust domain generalization methods that are specialized in biomedical problems.

\subsection{Molecule Generation in the Docking Generation Benchmark}

\xhdr{Motivation}
AI-assisted drug design aims to generate novel molecular structures with desired pharmaceutical properties. Recent progress in generative modeling has shown great promising results in this area. However, the current experiments focus on optimizing simple heuristic oracles, such as QED (quantitative estimate of drug-likeness) and LogP (Octanol-water partition coefficient)~\citep{jin2018learning,You2018-xh,zhou2019optimization}, while an experimental evaluation, such as a bioassay, or a high-fidelity simulation, is much more costly in terms of resources that require a more data-efficient strategy. Further, as generative models can explore chemical space beyond a predefined one, the structure of the generated molecular might be valid but not synthesizable~\citep{gao2020synthesizability}. Therefore, we leverage docking simulation~\citep{cieplinski2020we,steinmann2021using} as an oracle and build up benchmark groups. Docking evaluates the affinity between a ligand (a small molecular drug) and a target (a protein involved in the disease), and is widely used in drug discovery in practice~\citep{lyu2019ultra}. In addition to the objective function value, we add a quality filter and a synthetic accessibility score to evaluate the generation quality within a limited number of oracle calls.

\xhdr{Experimental setup} We leverage \dataset{ZINC} dataset as the molecule library and \dataset{Docking} oracle function as the molecule docking score evaluator against the target DRD3, which is a crucial disease target for neurology diseases such as tremor and schizophrenia. To imitate a low-data scenario, we limit the number of oracle callings available to four levels: 100, 500, 1000, 5000. In addition to typical oracle scores, we investigate additional metrics to evaluate the quality of generated molecules, including (1) Top100/Top10/Top1: Average docking score of top-100/10/1 generated molecules for a given target; (2) Diversity: Average pairwise Tanimoto distance of Morgan fingerprints for Top 100 generated molecules; (3) Novelty: Fraction of generated molecules that are not present in the training set; (4) m1: Synthesizability score of molecules obtained via molecule.one retrosynthesis model~\citep{sacha2020molecule}; (5) \%pass: Fraction of generated molecules that successfully pass through apriori defined filters; (6) Top1 \%pass: The lowest docking score for molecules that pass the filter. Each model is run three times with different random seeds.

\xhdr{Baselines} 
We compare domain SOTA methods including Screening~\citep{lyu2019ultra} (simulated as random sampling), Graph-GA (graph-based genetic algorithm)~\citep{jensen2019graph}, and ML SOTA methods including string-based LSTM~\citep{segler2018generating}, GCPN (Graph Convolutional Policy Network)~\citep{You2018-xh}, MolDQN (Deep Q-Network)~\citep{zhou2019optimization} and MARS (Markov molecular Sampling)~\citep{xie2021mars}. We also include \textit{best-in-data}, which choose 100 molecules with the highest docking score from ZINC 250K database as reference. Methods follow the default hyperparameters described in the paper.

\xhdr{Results} 
Results are shown in Table~\ref{tab:docking-res}. 
Overall, we observe that almost all models cannot perform well under a limited oracle setting. The majority of the methods cannot surpass the best-in-data docking scores under 100, 500, 1,000 allowable oracle callings. In the 5,000 oracle callings setting, Graph-GA (-14.811) and LSTM (-13.017) finally surpass the best-in-data result. Graph-GA dominates the leaderboard with 0 learnable parameters in terms of optimization ability, while a simple SMILES LSTM ranked behind. The SOAT ML models that reported excellent performances in unlimited trivial oracles cannot beat virtual screening when allowing less than 5,000 oracle calls. This result questions the utility of the current ML SOTA methods and calls for a shift of focus on the current ML molecular generation communities to consider realistic constraints during evaluation.

As for the synthesizability, as the number of allowable oracles calls increases, the more significant fraction generates undesired molecular structures despite the increasing affinity. We observe a monotonous increment in the m1 score of the best performing Graph GA method when we allow more oracle calls. In the 5,000 calls category, only 2.3\% - 52.7\% of the generated molecules pass the molecule filters, and within the passed molecules, the best docking score drops significantly compared to before the filter. By contrast, LSTM keeps a relatively good generated quality in all categories, showing ML generative models have an advantage in learning the distribution of training sets and producing ``normal'' molecules. Also, the recent synthesizable constrained generation~\citep{korovina2020chembo,gottipati2020learning,bradshaw2020barking} is a promising approach to tackle this problem. We expect to see more ML models explicitly considering synthesizability.

\begin{table}[h!]
    \centering
    \caption{\textbf{Leaderboard on TDC DRD3 docking benchmark using \dataset{ZINC} and \dataset{Docking}.} Mean and standard deviation across three runs are reported. Arrows ($\uparrow$, $\downarrow$) indicate the direction of better performance. The best method is bolded and the second best is underlined. }
    \adjustbox{max width=\textwidth}{

    \begin{tabular}{l|c|c|cc|cccc}
    \midrule
    \multicolumn{3}{c|}{Method Category} & 
    \multicolumn{2}{c|}{\cellcolor{ds_color}Domain-Specific Methods} & \multicolumn{4}{c}{\cellcolor{ms_color}State-of-the-Art Methods in ML} \\ 
    \midrule
    Metric & Best-in-data & \# Calls & Screening & Graph-GA & LSTM & GCPN & MolDQN & MARS  \\ \midrule
    
    \# Params. & - & - & 0 & 0 & 3149K & 18K & 2694K & 153K \\ \midrule
    
 Top100 ($\downarrow$) & -12.080 & \multirow{8}{*}{100}& \underline{-7.554\std{0.065}}& -7.222\std{0.013}& \bf -7.594\std{0.182}& 3.860\std{0.102}& -5.178\std{0.341}& -5.928\std{0.298}\\
Top10 ($\downarrow$) & -12.590 & & -9.727\std{0.276}& \bf -10.177\std{0.158}& \underline{-10.033\std{0.186}}& -5.617\std{0.413}& -6.438\std{0.176}& -8.133\std{0.328}\\
Top1 ($\downarrow$) & -12.800 & & -10.367\std{0.464}& \bf -11.767\std{1.087}& -11.133\std{0.634}& \underline{-11.633\std{2.217}}& -7.020\std{0.194}& -9.100\std{0.712}\\
Diversity ($\uparrow$) & 0.864 & & 0.881\std{0.002}& 0.885\std{0.001}& 0.884\std{0.002}& \bf 0.909\std{0.001}& \underline{0.907\std{0.001}}& 0.873\std{0.010}\\
Novelty ($\uparrow$) & - & & -& 1.000\std{0.000}& 1.000\std{0.000}& 1.000\std{0.000}& 1.000\std{0.000}& 1.000\std{0.000}\\
\%Pass ($\uparrow$) & 0.780 & & 0.717\std{0.005}& 0.693\std{0.037}& \underline{0.763\std{0.019}}& 0.093\std{0.009}& 0.017\std{0.012}& \bf 0.807\std{0.033}\\
Top1 Pass ($\downarrow$) & -11.700 & & -2.467\std{2.229}& 0.000\std{0.000}& -1.100\std{1.417}& 7.667\std{0.262}& \underline{-3.630\std{2.588}}& \bf -3.633\std{0.946}\\
m1 ($\downarrow$) & 5.100 & & \underline{4.845\std{0.235}}& 5.223\std{0.256}& 5.219\std{0.247}& 10.000\std{0.000}& 10.000\std{0.000}& \bf 4.470\std{1.047}\\
\midrule
Top100 ($\downarrow$) & -12.080 & \multirow{8}{*}{500}& -9.341\std{0.039}& \bf -10.036\std{0.221}& \underline{-9.419\std{0.173}}& -8.119\std{0.104}& -6.357\std{0.084}& -7.278\std{0.198}\\
Top10 ($\downarrow$) & -12.590 & & -10.517\std{0.135}& \bf -11.527\std{0.533}& \underline{-10.687\std{0.335}}& -10.230\std{0.354}& -7.173\std{0.166}& -9.067\std{0.377}\\
Top1 ($\downarrow$) & -12.800 & & -11.167\std{0.309}& \bf -12.500\std{0.748}& -11.367\std{0.579}& \underline{-11.967\std{0.680}}& -7.620\std{0.185}& -9.833\std{0.309}\\
Diversity ($\uparrow$) & 0.864 & & 0.870\std{0.003}& 0.857\std{0.005}& 0.875\std{0.005}& \bf 0.914\std{0.001}& \underline{0.903\std{0.002}}& 0.866\std{0.005}\\
Novelty ($\uparrow$) & - & & -& 1.000\std{0.000}& 1.000\std{0.000}& 1.000\std{0.000}& 1.000\std{0.000}& 1.000\std{0.000}\\
\%Pass ($\uparrow$) & 0.780 & & \bf 0.770\std{0.029}& 0.710\std{0.080}& \underline{0.727\std{0.012}}& 0.127\std{0.005}& 0.030\std{0.016}& 0.660\std{0.050}\\
Top1 Pass ($\downarrow$) & -11.700 & & \underline{-8.767\std{0.047}}& \bf -9.300\std{0.163}& -8.767\std{0.170}& -7.200\std{0.141}& -6.030\std{0.073}& -6.100\std{0.141}\\
m1 ($\downarrow$) & 5.100 & & \bf 5.672\std{1.211}& 6.493\std{0.341}& \underline{5.787\std{0.934}}& 10.000\std{0.000}& 10.000\std{0.000}& 5.827\std{0.937}\\
\midrule
Top100 ($\downarrow$) & -12.080 & \multirow{8}{*}{1000}& -9.693\std{0.019}& \bf -11.224\std{0.484}& \underline{-9.971\std{0.115}}& -9.053\std{0.080}& -6.738\std{0.042}& -8.224\std{0.196}\\
Top10 ($\downarrow$) & -12.590 & & -10.777\std{0.189}& \bf -12.400\std{0.782}& \underline{-11.163\std{0.141}}& -11.027\std{0.273}& -7.506\std{0.085}& -9.843\std{0.068}\\
Top1 ($\downarrow$) & -12.800 & & -11.500\std{0.432}& \bf -13.233\std{0.713}& -11.967\std{0.205}& \underline{-12.033\std{0.618}}& -7.800\std{0.042}& -11.100\std{0.141}\\
Diversity ($\uparrow$) & 0.864 & & 0.873\std{0.003}& 0.815\std{0.046}& 0.871\std{0.004}& \bf 0.913\std{0.001}& \underline{0.904\std{0.001}}& 0.871\std{0.004}\\
Novelty ($\uparrow$) & - & & -& 1.000\std{0.000}& 1.000\std{0.000}& 1.000\std{0.000}& 1.000\std{0.000}& 1.000\std{0.000}\\
\%Pass ($\uparrow$) & 0.780 & & 0.757\std{0.026}& \bf 0.777\std{0.096}& \underline{0.777\std{0.026}}& 0.170\std{0.022}& 0.033\std{0.005}& 0.563\std{0.052}\\
Top1 Pass ($\downarrow$) & -11.700 & & -9.167\std{0.047}& \bf -10.600\std{0.374}& \underline{-9.367\std{0.094}}& -8.167\std{0.047}& -6.450\std{0.085}& -7.367\std{0.205}\\
m1 ($\downarrow$) & 5.100 & & \underline{5.527\std{0.780}}& 7.695\std{0.909}& \bf 4.818\std{0.541}& 10.000\std{0.000}& 10.000\std{0.000}& 6.037\std{0.137}\\
\midrule
Top100 ($\downarrow$) & -12.080 & \multirow{8}{*}{5000}& -10.542\std{0.035}& \bf -14.811\std{0.413}& \underline{-13.017\std{0.385}}& -10.045\std{0.226}& -8.236\std{0.089}& -9.509\std{0.035}\\
Top10 ($\downarrow$) & -12.590 & & -11.483\std{0.056}& \bf -15.930\std{0.336}& \underline{-14.030\std{0.421}}& -11.483\std{0.581}& -9.348\std{0.188}& -10.693\std{0.172}\\
Top1 ($\downarrow$) & -12.800 & & -12.100\std{0.356}& \bf -16.533\std{0.309}& \underline{-14.533\std{0.525}}& -12.300\std{0.993}& -9.990\std{0.194}& -11.433\std{0.450}\\
Diversity ($\uparrow$) & 0.864 & & 0.872\std{0.003}& 0.626\std{0.092}& 0.740\std{0.056}& \bf 0.922\std{0.002}& \underline{0.893\std{0.005}}& 0.873\std{0.002}\\
Novelty ($\uparrow$) & - & & -& 1.000\std{0.000}& 1.000\std{0.000}& 1.000\std{0.000}& 1.000\std{0.000}& 1.000\std{0.000}\\
\%Pass ($\uparrow$) & 0.780 & & \bf 0.683\std{0.073}& 0.393\std{0.308}& 0.257\std{0.103}& 0.167\std{0.045}& 0.023\std{0.012}& \underline{0.527\std{0.087}}\\
Top1 Pass ($\downarrow$) & -11.700 & & -10.100\std{0.000}& \bf -14.267\std{0.450}& \underline{-12.533\std{0.403}}& -9.367\std{0.170}& -7.980\std{0.112}& -9.000\std{0.082}\\
m1 ($\downarrow$) & 5.100 & & \bf 5.610\std{0.805}& 9.669\std{0.468}& \underline{5.826\std{1.908}}& 10.000\std{0.000}& 10.000\std{0.000}& 7.073\std{0.798}\\

     \bottomrule
    \end{tabular}
    }
    \label{tab:docking-res}
\end{table}


%% file: 090conclusion.tex
\section{Conclusion and Future Directions} \label{sec:discussion}

Therapeutics machine learning is an emerging field with many hard algorithmic challenges and applications with immense opportunities for expansion, innovation, and impact. 

To this end, our Therapeutics Data Commons (TDC) is a platform of AI-ready datasets and learning tasks for drug discovery and development. Curated datasets, strategies for systematic model development and evaluation, and an ecosystem of tools, leaderboards and community resources in TDC serve as a meeting point for domain and machine learning scientists. We envision that TDC can considerably accelerate machine learning model development, validation and transition into production and clinical implementation.

To facilitate algorithmic and scientific innovation in therapeutics, we will support the continued development of TDC to provide AI-ready datasets and enhance outreach to build an inclusive research community:
\begin{itemize}[leftmargin=*,topsep=0pt]
    \setlength\itemsep{0em}
    \item \textbf{New Learning Tasks and Datasets:} We are actively working to include new learning tasks and datasets and keep abreast with the state-of-the-art. We now work on tasks related to emerging therapeutic products, including antibody-drug conjugates (ADCs) and proteolysis targeting chimera (PROTACs), and new pipelines, including clinical trial design, drug delivery, and postmarketing safety.
    \item \textbf{New ML Tools:} We plan to implement additional data functions and provide additional tools, libraries, and community resources. 
    \item \textbf{New Leaderboards and Competitions:} We plan to design new leaderboards for tasks that are of interest to the therapeutics community and have great potential to benefit from advanced machine learning. 
\end{itemize}
Lastly, TDC is an open science initiative. We welcome contributions from the research community.

%% file: ref.bib
@article{attentivefp,
  title={Pushing the boundaries of molecular representation for drug discovery with the graph attention mechanism},
  author={Xiong, Zhaoping and Wang, Dingyan and Liu, Xiaohong and Zhong, Feisheng and Wan, Xiaozhe and Li, Xutong and Li, Zhaojun and Luo, Xiaomin and Chen, Kaixian and Jiang, Hualiang and others},
  journal={Journal of Medicinal Chemistry},
  volume={63},
  number={16},
  pages={8749--8760},
  year={2019},
  publisher={ACS Publications}
}

@article{neuralfp,
  title={Convolutional networks on graphs for learning molecular fingerprints},
  author={Duvenaud, David and Maclaurin, Dougal and Aguilera-Iparraguirre, Jorge and G{\'o}mez-Bombarelli, Rafael and Hirzel, Timothy and Aspuru-Guzik, Al{\'a}n and Adams, Ryan P},
  journal={NeurIPS},
  year={2015}
}

@article{pretrain_gnn,
  title={Strategies for pre-training graph neural networks},
  author={Hu, Weihua and Liu, Bowen and Gomes, Joseph and Zitnik, Marinka and Liang, Percy and Pande, Vijay and Leskovec, Jure},
  journal={ICLR},
  year={2020}
}

@article{ozturk2018deepdta,
  title={DeepDTA: deep drug--target binding affinity prediction},
  author={{\"O}zt{\"u}rk, Hakime and {\"O}zg{\"u}r, Arzucan and Ozkirimli, Elif},
  journal={Bioinformatics},
  volume={34},
  number={17},
  pages={i821--i829},
  year={2018},
  publisher={Oxford University Press}
}

@article{lyu2019ultra,
  title={Ultra-large library docking for discovering new chemotypes},
  author={Lyu, Jiankun and Wang, Sheng and Balius, Trent E and Singh, Isha and Levit, Anat and Moroz, Yurii S and O’Meara, Matthew J and Che, Tao and Algaa, Enkhjargal and Tolmachova, Kateryna and others},
  journal={Nature},
  volume={566},
  number={7743},
  pages={224--229},
  year={2019},
  publisher={Nature Publishing Group}
}

@article{koh2020wilds,
  title={Wilds: A benchmark of in-the-wild distribution shifts},
  author={Koh, Pang Wei and Sagawa, Shiori and Marklund, Henrik and Xie, Sang Michael and Zhang, Marvin and Balsubramani, Akshay and Hu, Weihua and Yasunaga, Michihiro and Phillips, Richard Lanas and Gao, Irena and others},
  journal={ICML},
  year={2021}
}

@article{steinmann2021using,
  title={Using a genetic algorithm to find molecules with good docking scores},
  author={Steinmann, Casper and Jensen, Jan H},
  journal={PeerJ Physical Chemistry},
  volume={3},
  pages={e18},
  year={2021},
  publisher={PeerJ Inc.}
}

@article{xu2018powerful,
  title={How powerful are graph neural networks?},
  author={Xu, Keyulu and Hu, Weihua and Leskovec, Jure and Jegelka, Stefanie},
  journal={ICLR},
  year={2018}
}

@article{morgan,
  title={Extended-connectivity fingerprints},
  author={Rogers, David and Hahn, Mathew},
  journal={Journal of Chemical Information and Modeling},
  volume={50},
  number={5},
  pages={742--754},
  year={2010},
  publisher={ACS Publications}
}

@article{kipf2016semi,
  title={Semi-supervised classification with graph convolutional networks},
  author={Kipf, Thomas N and Welling, Max},
  journal={ICLR},
  year={2017}
}

@article{abul2019personalized,
  title={Personalized medicine and the power of electronic health records},
  author={Abul-Husn, Noura S and Kenny, Eimear E},
  journal={Cell},
  volume={177},
  number={1},
  pages={58--69},
  year={2019}
}

@article{karczewski2018integrative,
  title={Integrative omics for health and disease},
  author={Karczewski, Konrad J and Snyder, Michael P},
  journal={Nature Reviews Genetics},
  volume={19},
  number={5},
  pages={299},
  year={2018},
  publisher={Nature Publishing Group}
}

@article{nosengo2016new,
  title={New tricks for old drugs},
  author={Nosengo, Nicola},
  journal={Nature},
  volume={534},
  number={7607},
  pages={314--317},
  year={2016}
}

@article{zitnik2018biosnap,
  title={{BioSNAP Datasets}: Stanford biomedical network dataset collection},
  author={Zitnik, Marinka and Sosic, Rok and Leskovec, Jure},
  journal={http://snap. stanford. edu/biodata},
  volume={5},
  number={1},
  year={2018}
}

@article{pushpakom2019drug,
  title={Drug repurposing: progress, challenges and recommendations},
  author={Pushpakom, Sudeep and Iorio, Francesco and Eyers, Patrick A and Escott, K Jane and Hopper, Shirley and Wells, Andrew and Doig, Andrew and Guilliams, Tim and Latimer, Joanna and McNamee, Christine and others},
  journal={Nature Reviews Drug discovery},
  volume={18},
  number={1},
  pages={41--58},
  year={2019}
}

@article{gulrajani2020search,
  title={In search of lost domain generalization},
  author={Gulrajani, Ishaan and Lopez-Paz, David},
  journal={ICLR},
  year={2021}
}

@article{erm,
  title={An overview of statistical learning theory},
  author={Vapnik, Vladimir N},
  journal={IEEE Transactions on Neural Networks},
  volume={10},
  number={5},
  pages={988--999},
  year={1999},
  publisher={IEEE}
}

@article{dro,
  title={Distributionally robust neural networks for group shifts: On the importance of regularization for worst-case generalization},
  author={Sagawa, Shiori and Koh, Pang Wei and Hashimoto, Tatsunori B and Liang, Percy},
  journal={ICLR},
  year={2020}
}

@inproceedings{mmd,
  title={Domain generalization with adversarial feature learning},
  author={Li, Haoliang and Pan, Sinno Jialin and Wang, Shiqi and Kot, Alex C},
  booktitle={CVPR},
  pages={5400--5409},
  year={2018}
}

@inproceedings{coral,
  title={Deep coral: Correlation alignment for deep domain adaptation},
  author={Sun, Baochen and Saenko, Kate},
  booktitle={ECCV},
  pages={443--450},
  year={2016},
  organization={Springer}
}

@inproceedings{irm,
  title={Invariant risk minimization games},
  author={Ahuja, Kartik and Shanmugam, Karthikeyan and Varshney, Kush and Dhurandhar, Amit},
  booktitle={ICML},
  pages={145--155},
  year={2020}
}

@article{mtl,
  title={Domain Generalization by Marginal Transfer Learning.},
  author={Blanchard, Gilles and Deshmukh, Aniket Anand and Dogan, {\"U}r{\"u}n and Lee, Gyemin and Scott, Clayton},
  journal={JMLR},
  volume={22},
  pages={2--1},
  year={2021}
}

@article{andmask,
  title={Learning explanations that are hard to vary},
  author={Parascandolo, Giambattista and Neitz, Alexander and Orvieto, Antonio and Gresele, Luigi and Sch{\"o}lkopf, Bernhard},
  journal={ICLR},
  year={2021}
}

@INPROCEEDINGS{You2018-xh,
  title     = "Graph Convolutional Policy Network for Goal-directed Molecular
               Graph Generation",
  booktitle = "NIPS",
  author    = "You, Jiaxuan and Liu, Bowen and Ying, Rex and Pande, Vijay and Leskovec, Jure",
  year      =  2018
}

@article{zhou2019optimization,
  title={Optimization of molecules via deep reinforcement learning},
  author={Zhou, Zhenpeng and Kearnes, Steven and Li, Li and Zare, Richard N and Riley, Patrick},
  journal={Scientific reports},
  volume={9},
  number={1},
  pages={1--10},
  year={2019},
  publisher={Nature Publishing Group}
}

@article{segler2018generating,
  title={Generating focused molecule libraries for drug discovery with recurrent neural networks},
  author={Segler, Marwin HS and Kogej, Thierry and Tyrchan, Christian and Waller, Mark P},
  journal={ACS Central Science},
  volume={4},
  number={1},
  pages={120--131},
  year={2018},
  publisher={ACS Publications}
}

@article{jensen2019graph,
  title={A graph-based genetic algorithm and generative model/Monte Carlo tree search for the exploration of chemical space},
  author={Jensen, Jan H},
  journal={Chemical Science},
  volume={10},
  number={12},
  pages={3567--3572},
  year={2019},
  publisher={Royal Society of Chemistry}
}

@inproceedings{xie2021mars,
  title={{MARS}: Markov Molecular Sampling for Multi-objective Drug Discovery },
  author={Xie, Yutong and Shi, Chence and Zhou, Hao and Yang, Yuwei and Zhang, Weinan and Yu, Yong and Li, Lei},
  booktitle={ICLR},
  year={2021}
}

@inproceedings{korovina2020chembo,
  title={Chembo: Bayesian optimization of small organic molecules with synthesizable recommendations},
  author={Korovina, Ksenia and Xu, Sailun and Kandasamy, Kirthevasan and Neiswanger, Willie and Poczos, Barnabas and Schneider, Jeff and Xing, Eric},
  booktitle={AISTATS},
  pages={3393--3403},
  year={2020},
  organization={PMLR}
}

@inproceedings{gottipati2020learning,
  title={Learning to navigate the synthetically accessible chemical space using reinforcement learning},
  author={Gottipati, Sai Krishna and Sattarov, Boris and Niu, Sufeng and Pathak, Yashaswi and Wei, Haoran and Liu, Shengchao and Blackburn, Simon and Thomas, Karam and Coley, Connor and Tang, Jian and others},
  booktitle={ICML},
  pages={3668--3679},
  year={2020}
}

@article{bradshaw2020barking,
  title={Barking up the right tree: an approach to search over molecule synthesis DAGs},
  author={Bradshaw, John and Paige, Brooks and Kusner, Matt J and Segler, Marwin HS and Hern{\'a}ndez-Lobato, Jos{\'e} Miguel},
  journal={NeurIPS},
  year={2020}
}

@article{yang2019analyzing,
  title={Analyzing learned molecular representations for property prediction},
  author={Yang, Kevin and Swanson, Kyle and Jin, Wengong and Coley, Connor and Eiden, Philipp and Gao, Hua and Guzman-Perez, Angel and Hopper, Timothy and Kelley, Brian and Mathea, Miriam and others},
  journal={Journal of Chemical Information and Modeling},
  volume={59},
  number={8},
  pages={3370--3388},
  year={2019},
  publisher={ACS Publications}
}

@article{preuer2018frechet,
  title={Fr{\'e}chet ChemNet distance: a metric for generative models for molecules in drug discovery},
  author={Preuer, Kristina and Renz, Philipp and Unterthiner, Thomas and Hochreiter, Sepp and Klambauer, Günter},
  journal={Journal of Chemical Information and Modeling},
  volume={58},
  number={9},
  pages={1736--1741},
  year={2018},
  publisher={ACS Publications}
}

@article{schneider2015development,
  title={Development of a novel fingerprint for chemical reactions and its application to large-scale reaction classification and similarity},
  author={Schneider, Nadine and Lowe, Daniel M and Sayle, Roger A and Landrum, Gregory A},
  journal={Journal of Chemical Information and Modeling},
  volume={55},
  number={1},
  pages={39--53},
  year={2015},
  publisher={ACS Publications}
}

@article{irwin2012zinc,
  title={ZINC: a free tool to discover chemistry for biology},
  author={Irwin, John J and Sterling, Teague and Mysinger, Michael M and Bolstad, Erin S and Coleman, Ryan G},
  journal={Journal of Chemical Information and Modeling},
  volume={52},
  number={7},
  pages={1757--1768},
  year={2012},
  publisher={ACS Publications}
}

@article{lauer2012developability,
  title={Developability index: a rapid in silico tool for the screening of antibody aggregation propensity},
  author={Lauer, Timothy M and Agrawal, Neeraj J and Chennamsetty, Naresh and Egodage, Kamal and Helk, Bernhard and Trout, Bernhardt L},
  journal={Journal of Pharmaceutical Sciences},
  volume={101},
  number={1},
  pages={102--115},
  year={2012},
  publisher={Elsevier}
}

@article{haghighatlari2020chemml,
  title={ChemML: A machine learning and informatics program package for the analysis, mining, and modeling of chemical and materials data},
  author={Haghighatlari, Mojtaba and Vishwakarma, Gaurav and Altarawy, Doaa and Subramanian, Ramachandran and Kota, Bhargava U and Sonpal, Aditya and Setlur, Srirangaraj and Hachmann, Johannes},
  journal={Wiley Interdisciplinary Reviews: Computational Molecular Science},
  volume={10},
  number={4},
  pages={e1458},
  year={2020},
  publisher={Wiley Online Library}
}

@article{hochreiter2006new,
  title={A new summarization method for Affymetrix probe level data},
  author={Hochreiter, Sepp and Clevert, Djork-Arne and Obermayer, Klaus},
  journal={Bioinformatics},
  volume={22},
  number={8},
  pages={943--949},
  year={2006},
  publisher={Oxford University Press}
}

@inproceedings{rao2019evaluating,
  title={Evaluating protein transfer learning with tape},
  author={Rao, Roshan and Bhattacharya, Nicholas and Thomas, Neil and Duan, Yan and Chen, Peter and Canny, John and Abbeel, Pieter and Song, Yun},
  booktitle={NeurIPS},
  pages={9689--9701},
  year={2019}
}

@article{van2003admet,
  title={{ADMET} in silico modelling: towards prediction paradise?},
  author={Van De Waterbeemd, Han and Gifford, Eric},
  journal={Nature Reviews Drug discovery},
  volume={2},
  number={3},
  pages={192--204},
  year={2003},
  publisher={Nature publishing group}
}

@article{kennedy1997managing,
  title={Managing the drug discovery/development interface},
  author={Kennedy, Tony},
  journal={Drug Discovery Today},
  volume={2},
  number={10},
  pages={436--444},
  year={1997},
  publisher={Elsevier}
}

@article{gao2020deep,
  title={Deep learning in protein structural modeling and design},
  author={Gao, Wenhao and Mahajan, Sai Pooja and Sulam, Jeremias and Gray, Jeffrey J},
  journal={Patterns},
  pages={100142},
  year={2020},
  publisher={Elsevier}
}

@article{ahneman2018predicting,
  title={Predicting reaction performance in C--N cross-coupling using machine learning},
  author={Ahneman, Derek T and Estrada, Jes{\'u}s G and Lin, Shishi and Dreher, Spencer D and Doyle, Abigail G},
  journal={Science},
  volume={360},
  number={6385},
  pages={186--190},
  year={2018},
  publisher={American Association for the Advancement of Science}
}

@misc{lowe2017chemical,
  title={Chemical reactions from US patents (1976-Sep2016). Figshare},
  author={Lowe, Daniel Mark},
  year={2017}
}

@article{caco2,
  title={ADME properties evaluation in drug discovery: prediction of Caco-2 cell permeability using a combination of NSGA-II and boosting},
  author={Wang, Ning-Ning and Dong, Jie and Deng, Yin-Hua and Zhu, Min-Feng and Wen, Ming and Yao, Zhi-Jiang and Lu, Ai-Ping and Wang, Jian-Bing and Cao, Dong-Sheng},
  journal={Journal of Chemical Information and Modeling},
  volume={56},
  number={4},
  pages={763--773},
  year={2016},
  publisher={ACS Publications}
}

@article{hia,
  title={ADME evaluation in drug discovery. 7. Prediction of oral absorption by correlation and classification},
  author={Hou, Tingjun and Wang, Junmei and Zhang, Wei and Xu, Xiaojie},
  journal={Journal of Chemical Information and Modeling},
  volume={47},
  number={1},
  pages={208--218},
  year={2007},
  publisher={ACS Publications}
}

@article{pgp,
  title={A novel approach for predicting P-glycoprotein (ABCB1) inhibition using molecular interaction fields},
  author={Broccatelli, Fabio and Carosati, Emanuele and Neri, Annalisa and Frosini, Maria and Goracci, Laura and Oprea, Tudor I and Cruciani, Gabriele},
  journal={Journal of Medicinal Chemistry},
  volume={54},
  number={6},
  pages={1740--1751},
  year={2011},
  publisher={ACS Publications}
}

@article{bioavail,
  title={Prediction models of human plasma protein binding rate and oral bioavailability derived by using GA--CG--SVM method},
  author={Ma, Chang-Ying and Yang, Sheng-Yong and Zhang, Hui and Xiang, Ming-Li and Huang, Qi and Wei, Yu-Quan},
  journal={Journal of Pharmaceutical and Biomedical Analysis},
  volume={47},
  number={4-5},
  pages={677--682},
  year={2008},
  publisher={Elsevier}
}

@article{astrazeneca,
  title={Experimental in vitro Dmpk and physicochemical data on a set of publicly disclosed compounds},
  author={AstraZeneca},
  journal={ChEMBL},
  year={2016}
}

@article{molnet,
  title={MoleculeNet: a benchmark for molecular machine learning},
  author={Wu, Zhenqin and Ramsundar, Bharath and Feinberg, Evan N and Gomes, Joseph and Geniesse, Caleb and Pappu, Aneesh S and Leswing, Karl and Pande, Vijay},
  journal={Chemical Science},
  volume={9},
  number={2},
  pages={513--530},
  year={2018},
  publisher={Royal Society of Chemistry}
}

@article{aqsoldb,
  title={AqSolDB, a curated reference set of aqueous solubility and 2D descriptors for a diverse set of compounds},
  author={Sorkun, Murat Cihan and Khetan, Abhishek and Er, S{\"u}leyman},
  journal={Scientific Data},
  volume={6},
  number={1},
  pages={1--8},
  year={2019},
  publisher={Nature Publishing Group}
}

@article{bbb,
  title={A Bayesian approach to in silico blood-brain barrier penetration modeling},
  author={Martins, Ines Filipa and Teixeira, Ana L and Pinheiro, Luis and Falcao, Andre O},
  journal={Journal of Chemical Information and Modeling},
  volume={52},
  number={6},
  pages={1686--1697},
  year={2012},
  publisher={ACS Publications}
}

@article{VDss,
  title={In silico prediction of volume of distribution in humans. Extensive data set and the exploration of linear and nonlinear methods coupled with molecular interaction fields descriptors},
  author={Lombardo, Franco and Jing, Yankang},
  journal={Journal of Chemical Information and Modeling},
  volume={56},
  number={10},
  pages={2042--2052},
  year={2016},
  publisher={ACS Publications}
}

@article{cypp450,
  title={Comprehensive characterization of cytochrome P450 isozyme selectivity across chemical libraries},
  author={Veith, Henrike and Southall, Noel and Huang, Ruili and James, Tim and Fayne, Darren and Artemenko, Natalia and Shen, Min and Inglese, James and Austin, Christopher P and Lloyd, David G and others},
  journal={Nature Biotechnology},
  volume={27},
  number={11},
  pages={1050--1055},
  year={2009},
  publisher={Nature Publishing Group}
}

@article{cyp_substrate,
  title={Selecting relevant descriptors for classification by bayesian estimates: a comparison with decision trees and support vector machines approaches for disparate data sets},
  author={Carbon-Mangels, Miriam and Hutter, Michael C},
  journal={Molecular Informatics},
  volume={30},
  number={10},
  pages={885--895},
  year={2011},
  publisher={Wiley Online Library}
}

@article{hu2020open,
  title={{Open Graph Benchmark}: Datasets for machine learning on graphs},
  author={Hu, Weihua and Fey, Matthias and Zitnik, Marinka and Dong, Yuxiao and Ren, Hongyu and Liu, Bowen and Catasta, Michele and Leskovec, Jure},
  journal={NeurIPS},
  year={2020}
}

@article{halflife,
  title={Trend analysis of a database of intravenous pharmacokinetic parameters in humans for 670 drug compounds},
  author={Obach, R Scott and Lombardo, Franco and Waters, Nigel J},
  journal={Drug Metabolism and Disposition},
  volume={36},
  number={7},
  pages={1385--1405},
  year={2008}
}

@article{ld50,
  title={Quantitative structure- activity relationship modeling of rat acute toxicity by oral exposure},
  author={Zhu, Hao and Martin, Todd M and Ye, Lin and Sedykh, Alexander and Young, Douglas M and Tropsha, Alexander},
  journal={Chemical Research in Toxicology},
  volume={22},
  number={12},
  pages={1913--1921},
  year={2009},
  publisher={ACS Publications}
}

@article{herg,
  title={ADMET evaluation in drug discovery. 16. Predicting hERG blockers by combining multiple pharmacophores and machine learning approaches},
  author={Wang, Shuangquan and Sun, Huiyong and Liu, Hui and Li, Dan and Li, Youyong and Hou, Tingjun},
  journal={Molecular Pharmaceutics},
  volume={13},
  number={8},
  pages={2855--2866},
  year={2016},
  publisher={ACS Publications}
}

@article{ames,
  title={In silico prediction of chemical Ames mutagenicity},
  author={Xu, Congying and Cheng, Feixiong and Chen, Lei and Du, Zheng and Li, Weihua and Liu, Guixia and Lee, Philip W and Tang, Yun},
  journal={Journal of Chemical Information and Modeling},
  volume={52},
  number={11},
  pages={2840--2847},
  year={2012},
  publisher={ACS Publications}
}

@article{dili,
  title={Deep learning for drug-induced liver injury},
  author={Xu, Youjun and Dai, Ziwei and Chen, Fangjin and Gao, Shuaishi and Pei, Jianfeng and Lai, Luhua},
  journal={Journal of Chemical Information and Modeling},
  volume={55},
  number={10},
  pages={2085--2093},
  year={2015},
  publisher={ACS Publications}
}

@article{skin_reaction,
  title={Predicting chemically-induced skin reactions. Part {I}: {QSAR} models of skin sensitization and their application to identify potentially hazardous compounds},
  author={Alves, Vinicius M and Muratov, Eugene and Fourches, Denis and Strickland, Judy and Kleinstreuer, Nicole and Andrade, Carolina H and Tropsha, Alexander},
  journal={Toxicology and Applied Pharmacology},
  volume={284},
  number={2},
  pages={262--272},
  year={2015},
  publisher={Elsevier}
}

@article{carcinogen,
  title={Computer-aided prediction of rodent carcinogenicity by {PASS} and {CISOC-PSCT}},
  author={Lagunin, Alexey and Filimonov, Dmitrii and Zakharov, Alexey and Xie, Wei and Huang, Ying and Zhu, Fucheng and Shen, Tianxiang and Yao, Jianhua and Poroikov, Vladimir},
  journal={QSAR \& Combinatorial Science},
  volume={28},
  number={8},
  pages={806--810},
  year={2009},
  publisher={Wiley Online Library}
}

@article{tox21,
  title={{DeepTox}: toxicity prediction using deep learning},
  author={Mayr, Andreas and Klambauer, G{\"u}nter and Unterthiner, Thomas and Hochreiter, Sepp},
  journal={Frontiers in Environmental Science},
  volume={3},
  pages={80},
  year={2016},
  publisher={Frontiers}
}

@article{clintox,
  title={A data-driven approach to predicting successes and failures of clinical trials},
  author={Gayvert, Kaitlyn M and Madhukar, Neel S and Elemento, Olivier},
  journal={Cell Chemical Biology},
  volume={23},
  number={10},
  pages={1294--1301},
  year={2016},
  publisher={Elsevier}
}

@article{sarscov2vitro,
  title={In vitro screening of a {FDA} approved chemical library reveals potential inhibitors of {SARS-CoV-2} replication},
  author={Touret, Franck and Gilles, Magali and Barral, Karine and Nougair{\`e}de, Antoine and van Helden, Jacques and Decroly, Etienne and de Lamballerie, Xavier and Coutard, Bruno},
  journal={Scientific Reports},
  volume={10},
  number={1},
  pages={1--8},
  year={2020},
  publisher={Nature Publishing Group}
}

@online{diamondprotease,
  author = {{Diamond Light Source}},
  title = {{Main protease structure and XChem fragment screen}},
  year = 2020,
  url = {https://www.diamond.ac.uk/covid-19/for-scientists/Main-protease-structure-and-XChem.html},
  urldate = {2021-01-16}
}

@online{mitaicures,
  author = {{MIT}},
  title = {{MIT AI Cures}},
  year = 2020,
  url = {https://www.aicures.mit.edu/},
  urldate = {2021-01-16}
}

@online{aids,
  author = {{NIH}},
  title = {{AIDS Antiviral Screen Data}},
  year = 2015,
  url = {https://wiki.nci.nih.gov/display/NCIDTPdata/AIDS+Antiviral+Screen+Data},
  urldate = {2021-01-16}
}

@misc{hiv,
  title={{AIDS Antiviral Screen Data}},
  author={National Cancer Institute},
  url={https://wiki.nci.nih.gov/display/NCIDTPdata/AIDS+Antiviral+Screen+Data},
  year={2015}
}

@article{qm7_1,
  title={970 million druglike small molecules for virtual screening in the chemical universe database {GDB-13}},
  author={Blum, Lorenz C and Reymond, Jean-Louis},
  journal={Journal of the American Chemical Society},
  volume={131},
  number={25},
  pages={8732--8733},
  year={2009},
  publisher={ACS Publications}
}

@article{qm7_2,
  title={Machine learning of molecular electronic properties in chemical compound space},
  author={Montavon, Gr{\'e}goire and Rupp, Matthias and Gobre, Vivekanand and Vazquez-Mayagoitia, Alvaro and Hansen, Katja and Tkatchenko, Alexandre and M{\"u}ller, Klaus-Robert and Von Lilienfeld, O Anatole},
  journal={New Journal of Physics},
  volume={15},
  number={9},
  pages={095003},
  year={2013},
  publisher={IOP Publishing}
}

@article{qm89_1,
  title={Enumeration of 166 billion organic small molecules in the chemical universe database {GDB-17}},
  author={Ruddigkeit, Lars and Van Deursen, Ruud and Blum, Lorenz C and Reymond, Jean-Louis},
  journal={Journal of Chemical Information and Modeling},
  volume={52},
  number={11},
  pages={2864--2875},
  year={2012},
  publisher={ACS Publications}
}

@article{qm89_2,
  title={Electronic spectra from {TDDFT} and machine learning in chemical space},
  author={Ramakrishnan, Raghunathan and Hartmann, Mia and Tapavicza, Enrico and Von Lilienfeld, O Anatole},
  journal={The Journal of Chemical Physics},
  volume={143},
  number={8},
  pages={084111},
  year={2015},
  publisher={AIP Publishing LLC}
}

@article{ramakrishnan2014quantum,
  title={Quantum chemistry structures and properties of 134 kilo molecules},
  author={Ramakrishnan, Raghunathan and Dral, Pavlo O and Rupp, Matthias and Von Lilienfeld, O Anatole},
  journal={Scientific Data},
  volume={1},
  number={1},
  pages={1--7},
  year={2014},
  publisher={Nature Publishing Group}
}

@book{deepchem,
  title={Deep learning for the life sciences: applying deep learning to genomics, microscopy, drug discovery, and more},
  author={Ramsundar, Bharath and Eastman, Peter and Walters, Patrick and Pande, Vijay},
  year={2019},
  publisher={{O'Reilly Media, Inc.}}
}

@phdthesis{uspto,
  title={Extraction of chemical structures and reactions from the literature},
  author={Lowe, Daniel Mark},
  year={2012},
  school={University of Cambridge}
}

@article{liberis,
  title={Parapred: antibody paratope prediction using convolutional and recurrent neural networks},
  author={Liberis, Edgar and Veli{\v{c}}kovi{\'c}, Petar and Sormanni, Pietro and Vendruscolo, Michele and Li{\`o}, Pietro},
  journal={Bioinformatics},
  volume={34},
  number={17},
  pages={2944--2950},
  year={2018},
  publisher={Oxford University Press}
}

@article{sabdab,
  title={{SAbDab}: the structural antibody database},
  author={Dunbar, James and Krawczyk, Konrad and Leem, Jinwoo and Baker, Terry and Fuchs, Angelika and Georges, Guy and Shi, Jiye and Deane, Charlotte M},
  journal={Nucleic Acids Research},
  volume={42},
  number={D1},
  pages={D1140--D1146},
  year={2014},
  publisher={Oxford University Press}
}

@article{iedb,
  title={The immune epitope database ({IEDB}): 2018 update},
  author={Vita, Randi and Mahajan, Swapnil and Overton, James A and Dhanda, Sandeep Kumar and Martini, Sheridan and Cantrell, Jason R and Wheeler, Daniel K and Sette, Alessandro and Peters, Bjoern},
  journal={Nucleic Acids Research},
  volume={47},
  number={D1},
  pages={D339--D343},
  year={2019},
  publisher={Oxford University Press}
}

@article{jespersen,
  title={{BepiPred-2.0}: improving sequence-based {B-cell} epitope prediction using conformational epitopes},
  author={Jespersen, Martin Closter and Peters, Bjoern and Nielsen, Morten and Marcatili, Paolo},
  journal={Nucleic Acids Research},
  volume={45},
  number={W1},
  pages={W24--W29},
  year={2017},
  publisher={Oxford University Press}
}

@article{pdb,
  title={The protein data bank},
  author={Berman, Helen M and Westbrook, John and Feng, Zukang and Gilliland, Gary and Bhat, Talapady N and Weissig, Helge and Shindyalov, Ilya N and Bourne, Philip E},
  journal={Nucleic Acids Research},
  volume={28},
  number={1},
  pages={235--242},
  year={2000},
  publisher={Oxford University Press}
}

@article{tap,
  title={Five computational developability guidelines for therapeutic antibody profiling},
  author={Raybould, Matthew IJ and Marks, Claire and Krawczyk, Konrad and Taddese, Bruck and Nowak, Jaroslaw and Lewis, Alan P and Bujotzek, Alexander and Shi, Jiye and Deane, Charlotte M},
  journal={Proceedings of the National Academy of Sciences},
  volume={116},
  number={10},
  pages={4025--4030},
  year={2019},
  publisher={National Acad Sciences}
}

@article{sabdab_chen,
  title={Predicting antibody developability from sequence using machine learning},
  author={Chen, Xingyao and Dougherty, Thomas and Hong, Chan and Schibler, Rachel and Zhao, Yi Cong and Sadeghi, Reza and Matasci, Naim and Wu, Yi-Chieh and Kerman, Ian},
  journal={bioRxiv},
  year={2020},
  publisher={Cold Spring Harbor Laboratory}
}

@article{biovia,
  title={{BIOVIA} pipeline pilot},
  author={Biovia, Dassault Syst{\`e}mes},
  journal={Dassault Syst{\`e}mes: San Diego, BW, Release},
  year={2017}
}

@article{leenay,
  title={Large dataset enables prediction of repair after {CRISPR--Cas9} editing in primary {T} cells},
  author={Leenay, Ryan T and Aghazadeh, Amirali and Hiatt, Joseph and Tse, David and Roth, Theodore L and Apathy, Ryan and Shifrut, Eric and Hultquist, Judd F and Krogan, Nevan and Wu, Zhenqin and others},
  journal={Nature Biotechnology},
  volume={37},
  number={9},
  pages={1034--1037},
  year={2019},
  publisher={Nature Publishing Group}
}

@article{lim2017principles,
  title={The principles of engineering immune cells to treat cancer},
  author={Lim, Wendell A and June, Carl H},
  journal={Cell},
  volume={168},
  number={4},
  pages={724--740},
  year={2017},
  publisher={Elsevier}
}

@article{van2016dna,
  title={{DNA} repair profiling reveals nonrandom outcomes at {Cas9}-mediated breaks},
  author={van Overbeek, Megan and Capurso, Daniel and Carter, Matthew M and Thompson, Matthew S and Frias, Elizabeth and Russ, Carsten and Reece-Hoyes, John S and Nye, Christopher and Gradia, Scott and Vidal, Bastien and others},
  journal={Molecular Cell},
  volume={63},
  number={4},
  pages={633--646},
  year={2016},
  publisher={Elsevier}
}

@article{bindingdb,
  title={{BindingDB}: a web-accessible database of experimentally determined protein--ligand binding affinities},
  author={Liu, Tiqing and Lin, Yuhmei and Wen, Xin and Jorissen, Robert N and Gilson, Michael K},
  journal={Nucleic Acids Research},
  volume={35},
  pages={D198--D201},
  year={2007},
  publisher={Oxford University Press}
}

@article{deeppurpose,
  title={{DeepPurpose}: A Deep Learning Library for Drug-Target Interaction Prediction},
  author={Huang, Kexin and Fu, Tianfan and Glass, Lucas M and Zitnik, Marinka and Xiao, Cao and Sun, Jimeng},
  journal={Bioinformatics},
  year={2020}
}

@article{davis,
  title={Comprehensive analysis of kinase inhibitor selectivity},
  author={Davis, Mindy I and Hunt, Jeremy P and Herrgard, Sanna and Ciceri, Pietro and Wodicka, Lisa M and Pallares, Gabriel and Hocker, Michael and Treiber, Daniel K and Zarrinkar, Patrick P},
  journal={Nature Biotechnology},
  volume={29},
  number={11},
  pages={1046--1051},
  year={2011},
  publisher={Nature Publishing Group}
}

@article{kiba,
  title={Making sense of large-scale kinase inhibitor bioactivity data sets: a comparative and integrative analysis},
  author={Tang, Jing and Szwajda, Agnieszka and Shakyawar, Sushil and Xu, Tao and Hintsanen, Petteri and Wennerberg, Krister and Aittokallio, Tero},
  journal={Journal of Chemical Information and Modeling},
  volume={54},
  number={3},
  pages={735--743},
  year={2014},
  publisher={ACS Publications}
}

@article{thomsen2007systematic,
  title={Systematic review of the incidence and characteristics of preventable adverse drug events in ambulatory care},
  author={Thomsen, Linda Aagaard and Winterstein, Almut G and S{\o} ndergaard, Birthe and Haugb{\o} lle, Lotte Stig and Melander, Arne},
  journal={Annals of Pharmacotherapy},
  volume={41},
  number={9},
  pages={1411--1426},
  year={2007},
  publisher={SAGE Publications Sage CA: Los Angeles, CA}
}

@article{lazarou1998incidence,
  title={Incidence of adverse drug reactions in hospitalized patients: a meta-analysis of prospective studies},
  author={Lazarou, Jason and Pomeranz, Bruce H and Corey, Paul N},
  journal={JAMA},
  volume={279},
  number={15},
  pages={1200--1205},
  year={1998},
  publisher={American Medical Association}
}

@article{drugbank,
  title={{DrugBank 5.0}: a major update to the {DrugBank} database for 2018},
  author={Wishart, David S and Feunang, Yannick D and Guo, An C and Lo, Elvis J and Marcu, Ana and Grant, Jason R and Sajed, Tanvir and Johnson, Daniel and Li, Carin and Sayeeda, Zinat and others},
  journal={Nucleic Acids Research},
  volume={46},
  number={D1},
  pages={D1074--D1082},
  year={2018},
  publisher={Oxford University Press}
}

@article{twosides,
  title={Data-driven prediction of drug effects and interactions},
  author={Tatonetti, Nicholas P and Patrick, P Ye and Daneshjou, Roxana and Altman, Russ B},
  journal={Science Translational Medicine},
  volume={4},
  number={125},
  pages={125ra31--125ra31},
  year={2012},
  publisher={American Association for the Advancement of Science}
}

@article{huri,
  title={A reference map of the human binary protein interactome},
  author={Luck, Katja and Kim, Dae-Kyum and Lambourne, Luke and Spirohn, Kerstin and Begg, Bridget E and Bian, Wenting and Brignall, Ruth and Cafarelli, Tiziana and Campos-Laborie, Francisco J and Charloteaux, Benoit and others},
  journal={Nature},
  volume={580},
  number={7803},
  pages={402--408},
  year={2020},
  publisher={Nature Publishing Group}
}

@article{disgenet,
  title={The {DisGeNET} knowledge platform for disease genomics: 2019 update},
  author={Pi{\~n}ero, Janet and Ram{\'\i}rez-Anguita, Juan Manuel and Sa{\"u}ch-Pitarch, Josep and Ronzano, Francesco and Centeno, Emilio and Sanz, Ferran and Furlong, Laura I},
  journal={Nucleic Acids Research},
  volume={48},
  number={D1},
  pages={D845--D855},
  year={2020},
  publisher={Oxford University Press}
}

@article{gdsc,
  title={Genomics of Drug Sensitivity in Cancer {(GDSC)}: a resource for therapeutic biomarker discovery in cancer cells},
  author={Yang, Wanjuan and Soares, Jorge and Greninger, Patricia and Edelman, Elena J and Lightfoot, Howard and Forbes, Simon and Bindal, Nidhi and Beare, Dave and Smith, James A and Thompson, I Richard and others},
  journal={Nucleic Acids Research},
  volume={41},
  number={D1},
  pages={D955--D961},
  year={2012},
  publisher={Oxford University Press}
}

@article{OncoPolyPharmacology,
  title={An unbiased oncology compound screen to identify novel combination strategies},
  author={O'Neil, Jennifer and Benita, Yair and Feldman, Igor and Chenard, Melissa and Roberts, Brian and Liu, Yaping and Li, Jing and Kral, Astrid and Lejnine, Serguei and Loboda, Andrey and others},
  journal={Molecular Cancer Therapeutics},
  volume={15},
  number={6},
  pages={1155--1162},
  year={2016},
  publisher={AACR}
}

@article{drugcomb,
  title={{DrugComb}: an integrative cancer drug combination data portal},
  author={Zagidullin, Bulat and Aldahdooh, Jehad and Zheng, Shuyu and Wang, Wenyu and Wang, Yinyin and Saad, Joseph and Malyutina, Alina and Jafari, Mohieddin and Tanoli, Ziaurrehman and Pessia, Alberto and others},
  journal={Nucleic Acids Research},
  volume={47},
  number={W1},
  pages={W43--W51},
  year={2019},
  publisher={Oxford University Press}
}

@article{cellminer,
  title={{CellMiner}: a web-based suite of genomic and pharmacologic tools to explore transcript and drug patterns in the NCI-60 cell line set},
  author={Reinhold, William C and Sunshine, Margot and Liu, Hongfang and Varma, Sudhir and Kohn, Kurt W and Morris, Joel and Doroshow, James and Pommier, Yves},
  journal={Cancer Research},
  volume={72},
  number={14},
  pages={3499--3511},
  year={2012},
  publisher={AACR}
}

@article{preuer2018deepsynergy,
  title={{DeepSynergy}: predicting anti-cancer drug synergy with Deep Learning},
  author={Preuer, Kristina and Lewis, Richard PI and Hochreiter, Sepp and Bender, Andreas and Bulusu, Krishna C and Klambauer, G{\"u}nter},
  journal={Bioinformatics},
  volume={34},
  number={9},
  pages={1538--1546},
  year={2018},
  publisher={Oxford University Press}
}

@article{nielsen,
  title={{NetMHCpan-3.0}: improved prediction of binding to {MHC class I} molecules integrating information from multiple receptor and peptide length datasets},
  author={Nielsen, Morten and Andreatta, Massimo},
  journal={Genome Medicine},
  volume={8},
  number={1},
  pages={1--9},
  year={2016},
  publisher={BioMed Central}
}

@article{jensen,
  title={Improved methods for predicting peptide binding affinity to {MHC class II} molecules},
  author={Jensen, Kamilla Kjaergaard and Andreatta, Massimo and Marcatili, Paolo and Buus, S{\o}ren and Greenbaum, Jason A and Yan, Zhen and Sette, Alessandro and Peters, Bjoern and Nielsen, Morten},
  journal={Immunology},
  volume={154},
  number={3},
  pages={394--406},
  year={2018},
  publisher={Wiley Online Library}
}

@article{mirtarbase,
  title={{miRTarBase} update 2018: a resource for experimentally validated {microRNA}-target interactions},
  author={Chou, Chih-Hung and Shrestha, Sirjana and Yang, Chi-Dung and Chang, Nai-Wen and Lin, Yu-Ling and Liao, Kuang-Wen and Huang, Wei-Chi and Sun, Ting-Hsuan and Tu, Siang-Jyun and Lee, Wei-Hsiang and others},
  journal={Nucleic Acids Research},
  volume={46},
  number={D1},
  pages={D296--D302},
  year={2018},
  publisher={Oxford University Press}
}

@article{mirbase,
  title={{miRBase}: from {microRNA} sequences to function},
  author={Kozomara, Ana and Birgaoanu, Maria and Griffiths-Jones, Sam},
  journal={Nucleic Acids Research},
  volume={47},
  number={D1},
  pages={D155--D162},
  year={2019},
  publisher={Oxford University Press}
}

@article{polykovskiy2018molecular,
  title={Molecular sets {(MOSES)}: a benchmarking platform for molecular generation models},
  author={Polykovskiy, Daniil and Zhebrak, Alexander and Sanchez-Lengeling, Benjamin and Golovanov, Sergey and Tatanov, Oktai and Belyaev, Stanislav and Kurbanov, Rauf and Artamonov, Aleksey and Aladinskiy, Vladimir and Veselov, Mark and others},
  journal={Frontiers in Pharmacology},
  year={2018}
}

@article{benhenda2017chemgan,
  title={{ChemGAN} challenge for drug discovery: can {AI} reproduce natural chemical diversity?},
  author={Benhenda, Mostapha},
  journal={arXiv:1708.08227},
  year={2017}
}

@article{brown2019guacamol,
  title={{GuacaMol}: benchmarking models for de novo molecular design},
  author={Brown, Nathan and Fiscato, Marco and Segler, Marwin HS and Vaucher, Alain C},
  journal={Journal of Chemical Information and Modeling},
  volume={59},
  number={3},
  pages={1096--1108},
  year={2019},
  publisher={ACS Publications}
}

@article{zinc15,
  title={ZINC 15--ligand discovery for everyone},
  author={Sterling, Teague and Irwin, John J},
  journal={Journal of Chemical Information and Modeling},
  volume={55},
  number={11},
  pages={2324--2337},
  year={2015},
  publisher={ACS Publications}
}

@article{molvae,
  title={Automatic chemical design using a data-driven continuous representation of molecules},
  author={G{\'o}mez-Bombarelli, Rafael and Wei, Jennifer N and Duvenaud, David and Hern{\'a}ndez-Lobato, Jos{\'e} Miguel and S{\'a}nchez-Lengeling, Benjam{\'\i}n and Sheberla, Dennis and Aguilera-Iparraguirre, Jorge and Hirzel, Timothy D and Adams, Ryan P and Aspuru-Guzik, Al{\'a}n},
  journal={ACS Central Science},
  volume={4},
  number={2},
  pages={268--276},
  year={2018},
  publisher={ACS Publications}
}

@article{chembl,
  title={{ChEMBL}: towards direct deposition of bioassay data},
  author={Mendez, David and Gaulton, Anna and Bento, A Patr{\'\i}cia and Chambers, Jon and De Veij, Marleen and F{\'e}lix, Eloy and Magari{\~n}os, Mar{\'\i}a Paula and Mosquera, Juan F and Mutowo, Prudence and Nowotka, Micha{\l} and others},
  journal={Nucleic Acids Research},
  volume={47},
  number={D1},
  pages={D930--D940},
  year={2019},
  publisher={Oxford University Press}
}

@article{chembl_web,
  title={{ChEMBL} web services: streamlining access to drug discovery data and utilities},
  author={Davies, Mark and Nowotka, Micha{\l} and Papadatos, George and Dedman, Nathan and Gaulton, Anna and Atkinson, Francis and Bellis, Louisa and Overington, John P},
  journal={Nucleic Acids Research},
  volume={43},
  number={W1},
  pages={W612--W620},
  year={2015},
  publisher={Oxford University Press}
}

@article{jin2018learning,
  title={Learning multimodal graph-to-graph translation for molecular optimization},
  author={Jin, Wengong and Yang, Kevin and Barzilay, Regina and Jaakkola, Tommi},
  journal={ICLR},
  year={2019}
}

@article{olivecrona2017molecular,
  title={Molecular de-novo design through deep reinforcement learning},
  author={Olivecrona, Marcus and Blaschke, Thomas and Engkvist, Ola and Chen, Hongming},
  journal={Journal of Cheminformatics},
  volume={9},
  number={1},
  pages={48},
  year={2017},
  publisher={Springer}
}

@article{bickerton2012quantifying,
  title={Quantifying the chemical beauty of drugs},
  author={Bickerton, G Richard and Paolini, Gaia V and Besnard, J{\'e}r{\'e}my and Muresan, Sorel and Hopkins, Andrew L},
  journal={Nature Chemistry},
  volume={4},
  number={2},
  pages={90--98},
  year={2012},
  publisher={Nature Publishing Group}
}

@article{kusner2017grammar,
  title={Grammar variational autoencoder},
  author={Kusner, Matt J and Paige, Brooks and Hern{\'a}ndez-Lobato, Jos{\'e} Miguel},
  journal={ICML},
  year={2017}
}

@article{liu2017retrosynthetic,
  title={Retrosynthetic reaction prediction using neural sequence-to-sequence models},
  author={Liu, Bowen and Ramsundar, Bharath and Kawthekar, Prasad and Shi, Jade and Gomes, Joseph and Luu Nguyen, Quang and Ho, Stephen and Sloane, Jack and Wender, Paul and Pande, Vijay},
  journal={ACS Central Science},
  volume={3},
  number={10},
  pages={1103--1113},
  year={2017},
  publisher={ACS Publications}
}

@article{zheng2019predicting,
  title={Predicting retrosynthetic reactions using self-corrected transformer neural networks},
  author={Zheng, Shuangjia and Rao, Jiahua and Zhang, Zhongyue and Xu, Jun and Yang, Yuedong},
  journal={Journal of Chemical Information and Modeling},
  volume={60},
  number={1},
  pages={47--55},
  year={2019},
  publisher={ACS Publications}
}

@article{krenn2020self,
  title={Self-Referencing Embedded Strings {(SELFIES)}: A 100\% robust molecular string representation},
  author={Krenn, Mario and H{\"a}se, Florian and Nigam, AkshatKumar and Friederich, Pascal and Aspuru-Guzik, Alan},
  journal={Machine Learning: Science and Technology},
  volume={1},
  number={4},
  pages={045024},
  year={2020},
  publisher={IOP Publishing}
}

@inproceedings{jin2017predicting,
  title={Predicting organic reaction outcomes with weisfeiler-lehman network},
  author={Jin, Wengong and Coley, Connor and Barzilay, Regina and Jaakkola, Tommi},
  booktitle={NeurIPS},
  pages={2607--2616},
  year={2017}
}

@article{sheridan2013time,
  title={Time-split cross-validation as a method for estimating the goodness of prospective prediction.},
  author={Sheridan, Robert P},
  journal={Journal of Chemical Information and Modeling},
  volume={53},
  number={4},
  pages={783--790},
  year={2013},
  publisher={ACS Publications}
}

@article{bemis1996properties,
  title={The properties of known drugs.},
  author={Bemis, Guy W and Murcko, Mark A},
  journal={Journal of Medicinal Chemistry},
  volume={39},
  number={15},
  pages={2887--2893},
  year={1996},
  publisher={ACS Publications}
}

@article{sambuy2005caco,
  title={{The Caco-2 cell line as a model of the intestinal barrier: influence of cell and culture-related factors on Caco-2 cell functional characteristics}},
  author={Sambuy, Y and De Angelis, I and Ranaldi, G and Scarino, ML and Stammati, A and Zucco, F},
  journal={Cell Biology and Toxicology},
  volume={21},
  number={1},
  pages={1--26},
  year={2005},
  publisher={Springer}
}

@article{wessel1998prediction,
  title={Prediction of human intestinal absorption of drug compounds from molecular structure},
  author={Wessel, Matthew D and Jurs, Peter C and Tolan, John W and Muskal, Steven M},
  journal={Journal of Chemical Information and Computer Sciences},
  volume={38},
  number={4},
  pages={726--735},
  year={1998},
  publisher={ACS Publications}
}

@article{amin2013p,
  title={P-glycoprotein inhibition for optimal drug delivery},
  author={Amin, Md Lutful},
  journal={Drug Target Insights},
  volume={7},
  pages={DTI--S12519},
  year={2013},
  publisher={SAGE Publications Sage UK: London, England}
}

@article{shen2013inhibition,
  title={{Inhibition of Wnt/$\beta$-catenin signaling downregulates P-glycoprotein and reverses multi-drug resistance of cholangiocarcinoma}},
  author={Shen, Dong-Yan and Zhang, Wei and Zeng, Xin and Liu, Chang-Qin},
  journal={Cancer Science},
  volume={104},
  number={10},
  pages={1303--1308},
  year={2013},
  publisher={Wiley Online Library}
}

@article{toutain2004bioavailability,
  title={Bioavailability and its assessment},
  author={Toutain, Pierre-Louis and BOUSQUET-M{\'E}LOU, Alain},
  journal={Journal of Veterinary Pharmacology and Therapeutics},
  volume={27},
  number={6},
  pages={455--466},
  year={2004},
  publisher={Wiley Online Library}
}

@article{waring2010lipophilicity,
  title={Lipophilicity in drug discovery},
  author={Waring, Michael J},
  journal={Expert Opinion on Drug Discovery},
  volume={5},
  number={3},
  pages={235--248},
  year={2010},
  publisher={Taylor \& Francis}
}

@article{savjani2012drug,
  title={Drug solubility: importance and enhancement techniques},
  author={Savjani, Ketan T and Gajjar, Anuradha K and Savjani, Jignasa K},
  journal={ISRN Pharmaceutics},
  volume={2012},
  year={2012},
  publisher={Hindawi}
}

@article{abbott2010structure,
  title={Structure and function of the blood--brain barrier},
  author={Abbott, N Joan and Patabendige, Adjanie AK and Dolman, Diana EM and Yusof, Siti R and Begley, David J},
  journal={Neurobiology of Disease},
  volume={37},
  number={1},
  pages={13--25},
  year={2010},
  publisher={Elsevier}
}

@article{lindup1981clinical,
  title={Clinical pharmacology: plasma protein binding of drugs.},
  author={Lindup, WE and Orme, MC},
  journal={British Medical Journal},
  volume={282},
  number={6259},
  pages={212},
  year={1981},
  publisher={BMJ Publishing Group}
}

@article{sjostrand1953volume,
  title={Volume and distribution of blood and their significance in regulating the circulation},
  author={Sj{\"o}strand, Torgny},
  journal={Physiological Reviews},
  volume={33},
  number={2},
  pages={202--228},
  year={1953}
}

@article{mcdonnell2013basic,
  title={Basic review of the cytochrome p450 system},
  author={McDonnell, Anne M and Dang, Cathyyen H},
  journal={Journal of the Advanced Practitioner in Oncology},
  volume={4},
  number={4},
  pages={263},
  year={2013},
  publisher={Harborside Press}
}

@article{teh2011pharmacogenomics,
  title={Pharmacogenomics of {CYP2D6}: molecular genetics, interethnic differences and clinical importance},
  author={Teh, Lay Kek and Bertilsson, Leif},
  journal={Drug Metabolism and Pharmacokinetics},
  pages={1112190300--1112190300},
  year={2011},
  publisher={The Japanese Society for the Study of Xenobiotics}
}

@article{zanger2013cytochrome,
  title={Cytochrome {P450} enzymes in drug metabolism: regulation of gene expression, enzyme activities, and impact of genetic variation},
  author={Zanger, Ulrich M and Schwab, Matthias},
  journal={Pharmacology \& Therapeutics},
  volume={138},
  number={1},
  pages={103--141},
  year={2013},
  publisher={Elsevier}
}

@article{benet1995basic,
  title={Basic principles of pharmacokinetics},
  author={Benet, Leslie Z and Zia-Amirhosseini, Parnian},
  journal={Toxicologic Pathology},
  volume={23},
  number={2},
  pages={115--123},
  year={1995},
  publisher={Sage Publications Sage CA: Thousand Oaks, CA}
}

@article{toutain2004plasma,
  title={Plasma Clearance},
  author={Toutain, Pierre-Louis and Bousquet-M{\'e}lou, Alain},
  journal={Journal of Veterinary Pharmacology and Therapeutics},
  volume={27},
  number={6},
  pages={415--425},
  year={2004},
  publisher={Wiley Online Library}
}

@article{di2012mechanistic,
  title={Mechanistic insights from comparing intrinsic clearance values between human liver microsomes and hepatocytes to guide drug design},
  author={Di, Li and Keefer, Christopher and Scott, Dennis O and Strelevitz, Timothy J and Chang, George and Bi, Yi-An and Lai, Yurong and Duckworth, Jonathon and Fenner, Katherine and Troutman, Matthew D and others},
  journal={European Journal of Medicinal Chemistry},
  volume={57},
  pages={441--448},
  year={2012},
  publisher={Elsevier}
}

@article{kramer2007application,
  title={The application of discovery toxicology and pathology towards the design of safer pharmaceutical lead candidates},
  author={Kramer, Jeffrey A and Sagartz, John E and Morris, Dale L},
  journal={Nature Reviews Drug Discovery},
  volume={6},
  number={8},
  pages={636--649},
  year={2007},
  publisher={Nature Publishing Group}
}

@article{assis2009human,
  title={Human drug hepatotoxicity: a contemporary clinical perspective},
  author={Assis, David N and Navarro, Victor J},
  journal={Expert Opinion on Drug Metabolism \& Toxicology},
  volume={5},
  number={5},
  pages={463--473},
  year={2009},
  publisher={Taylor \& Francis}
}

@misc{landrum2013rdkit,
  title={{RDKit}: A software suite for cheminformatics, computational chemistry, and predictive modeling},
  author={Landrum, Greg},
  year={2013},
  publisher={Academic Press}
}

@article{kitchen2004docking,
  title={Docking and scoring in virtual screening for drug discovery: methods and applications},
  author={Kitchen, Douglas B and Decornez, H{\'e}l{\`e}ne and Furr, John R and Bajorath, J{\"u}rgen},
  journal={Nature Reviews Drug discovery},
  volume={3},
  number={11},
  pages={935--949},
  year={2004},
  publisher={Nature Publishing Group}
}

@article{coley2020autonomous,
  title={Autonomous discovery in the chemical sciences part {II}: Outlook},
  author={Coley, Connor W and Eyke, Natalie S and Jensen, Klavs F},
  journal={Angewandte Chemie},
  volume={59},
  number={52},
  pages={23414--23436},
  year={2020},
  publisher={Wiley Online Library}
}

@article{cieplinski2020we,
  title={We should at least be able to Design Molecules that Dock Well},
  author={Cieplinski, Tobiasz and Danel, Tomasz and Podlewska, Sabina and Jastrzebski, Stanislaw},
  journal={arXiv:2006.16955},
  year={2020}
}

@article{graff2020accelerating,
  title={Accelerating high-throughput virtual screening through molecular pool-based active learning},
  author={Graff, David E and Shakhnovich, Eugene I and Coley, Connor W},
  journal={arXiv:2012.07127},
  year={2020}
}

@article{trott2010autodock,
  title={{AutoDock Vina}: improving the speed and accuracy of docking with a new scoring function, efficient optimization, and multithreading},
  author={Trott, Oleg and Olson, Arthur J},
  journal={Journal of Computational Chemistry},
  volume={31},
  number={2},
  pages={455--461},
  year={2010},
  publisher={Wiley Online Library}
}

@article{koes2013lessons,
  title={Lessons learned in empirical scoring with smina from the {CSAR} 2011 benchmarking exercise},
  author={Koes, David Ryan and Baumgartner, Matthew P and Camacho, Carlos J},
  journal={Journal of Chemical Information and Modeling},
  volume={53},
  number={8},
  pages={1893--1904},
  year={2013},
  publisher={ACS Publications}
}

@article{alhossary2015fast,
  title={Fast, accurate, and reliable molecular docking with {QuickVina 2}},
  author={Alhossary, Amr and Handoko, Stephanus Daniel and Mu, Yuguang and Kwoh, Chee-Keong},
  journal={Bioinformatics},
  volume={31},
  number={13},
  pages={2214--2216},
  year={2015},
  publisher={Oxford University Press}
}

@article{ng2015psovina,
  title={{PSOVina}: The hybrid particle swarm optimization algorithm for protein--ligand docking},
  author={Ng, Marcus CK and Fong, Simon and Siu, Shirley WI},
  journal={Journal of Bioinformatics and Computational Biology},
  volume={13},
  number={03},
  pages={1541007},
  year={2015},
  publisher={World Scientific}
}

@article{allen2015dock,
  title={{DOCK 6}: Impact of new features and current docking performance},
  author={Allen, William J and Balius, Trent E and Mukherjee, Sudipto and Brozell, Scott R and Moustakas, Demetri T and Lang, P Therese and Case, David A and Kuntz, Irwin D and Rizzo, Robert C},
  journal={Journal of Computational Chemistry},
  volume={36},
  number={15},
  pages={1132--1156},
  year={2015},
  publisher={Wiley Online Library}
}

@article{liu2020retrognn,
  title={{RetroGNN}: Approximating Retrosynthesis by Graph Neural Networks for De Novo Drug Design},
  author={Liu, Cheng-Hao and Korablyov, Maksym and Jastrz{\k{e}}bski, Stanis{\l}aw and W{\l}odarczyk-Pruszy{\'n}ski, Pawe{\l} and Bengio, Yoshua and Segler, Marwin HS},
  journal={arXiv:2011.13042},
  year={2020}
}

@article{sacha2020molecule,
  title={Molecule Edit Graph Attention Network: Modeling Chemical Reactions as Sequences of Graph Edits},
  author={Sacha, Miko{\l}aj and B{\l}a{\.z}, Miko{\l}aj and Byrski, Piotr and W{\l}odarczyk-Pruszy{\'n}ski, Pawe{\l} and Jastrz{\k{e}}bski, Stanis{\l}aw},
  journal={arXiv:2006.15426},
  year={2020}
}

@article{schwaller2019molecular,
  title={Molecular transformer: A model for uncertainty-calibrated chemical reaction prediction},
  author={Schwaller, Philippe and Laino, Teodoro and Gaudin, Th{\'e}ophile and Bolgar, Peter and Hunter, Christopher A and Bekas, Costas and Lee, Alpha A},
  journal={ACS Central Science},
  volume={5},
  number={9},
  pages={1572--1583},
  year={2019},
  publisher={ACS Publications}
}

@article{sun2017excape,
  title={{ExCAPE-DB}: an integrated large scale dataset facilitating Big Data analysis in chemogenomics},
  author={Sun, Jiangming and Jeliazkova, Nina and Chupakhin, Vladimir and Golib-Dzib, Jose-Felipe and Engkvist, Ola and Carlsson, Lars and Wegner, J{\"o}rg and Ceulemans, Hugo and Georgiev, Ivan and Jeliazkov, Vedrin and others},
  journal={Journal of Cheminformatics},
  volume={9},
  number={1},
  pages={17},
  year={2017},
  publisher={Springer}
}

@inproceedings{jin2020multi,
  title={Multi-objective molecule generation using interpretable substructures},
  author={Jin, Wengong and Barzilay, Regina and Jaakkola, Tommi},
  booktitle={ICML},
  pages={4849--4859},
  year={2020},
}

@article{gao2020synthesizability,
  title={The synthesizability of molecules proposed by generative models},
  author={Gao, Wenhao and Coley, Connor W},
  journal={Journal of Chemical Information and Modeling},
  year={2020},
  publisher={ACS Publications}
}

@article{ertl2009estimation,
  title={Estimation of synthetic accessibility score of drug-like molecules based on molecular complexity and fragment contributions},
  author={Ertl, Peter and Schuffenhauer, Ansgar},
  journal={Journal of Cheminformatics},
  volume={1},
  number={1},
  pages={8},
  year={2009},
  publisher={Springer}
}

@article{schwaller2020prediction,
  title={Prediction of chemical reaction yields using deep learning},
  author={Schwaller, Philippe and Vaucher, Alain C and Laino, Teodoro and Reymond, Jean-Louis},
  journal={ChemRxiv},
  volume={10},
  year={2020}
}

@article{coley2019robotic,
  title={A robotic platform for flow synthesis of organic compounds informed by AI planning},
  author={Coley, Connor W and Thomas, Dale A and Lummiss, Justin AM and Jaworski, Jonathan N and Breen, Christopher P and Schultz, Victor and Hart, Travis and Fishman, Joshua S and Rogers, Luke and Gao, Hanyu and others},
  journal={Science},
  volume={365},
  number={6453},
  pages={eaax1566},
  year={2019},
  publisher={American Association for the Advancement of Science}
}

@article{zahrt2019prediction,
  title={Prediction of higher-selectivity catalysts by computer-driven workflow and machine learning},
  author={Zahrt, Andrew F and Henle, Jeremy J and Rose, Brennan T and Wang, Yang and Darrow, William T and Denmark, Scott E},
  journal={Science},
  volume={363},
  number={6424},
  year={2019},
  publisher={American Association for the Advancement of Science}
}

@article{hughes2011principles,
  title={Principles of early drug discovery},
  author={Hughes, James P and Rees, Stephen and Kalindjian, S Barrett and Philpott, Karen L},
  journal={British Journal of Pharmacology},
  volume={162},
  number={6},
  pages={1239--1249},
  year={2011},
  publisher={Wiley Online Library}
}

@article{szklarczyk2015string,
  title={{STRING v10}: protein--protein interaction networks, integrated over the tree of life},
  author={Szklarczyk, Damian and Franceschini, Andrea and Wyder, Stefan and Forslund, Kristoffer and Heller, Davide and Huerta-Cepas, Jaime and Simonovic, Milan and Roth, Alexander and Santos, Alberto and Tsafou, Kalliopi P and others},
  journal={Nucleic Acids Research},
  volume={43},
  number={D1},
  pages={D447--D452},
  year={2015},
  publisher={Oxford University Press}
}

@article{baptista2020deep,
  title={Deep learning for drug response prediction in cancer},
  author={Baptista, Delora and Ferreira, Pedro G and Rocha, Miguel},
  journal={Briefings in Bioinformatics},
  year={2020}
}

@article{chen2006role,
  title={{The role of microRNA-1 and microRNA-133 in skeletal muscle proliferation and differentiation}},
  author={Chen, Jian-Fu and Mandel, Elizabeth M and Thomson, J Michael and Wu, Qiulian and Callis, Thomas E and Hammond, Scott M and Conlon, Frank L and Wang, Da-Zhi},
  journal={Nature Genetics},
  volume={38},
  number={2},
  pages={228--233},
  year={2006},
  publisher={Nature Publishing Group}
}

@article{hanna2019potential,
  title={The potential for {microRNA} therapeutics and clinical research},
  author={Hanna, Johora and Hossain, Gazi S and Kocerha, Jannet},
  journal={Frontiers in Genetics},
  volume={10},
  pages={478},
  year={2019},
  publisher={Frontiers}
}

@inproceedings{agrawal2018large,
  title={Large-scale analysis of disease pathways in the human interactome},
  author={Agrawal, Monica and Zitnik, Marinka and Leskovec, Jure and others},
  booktitle={Pacific Symposium on Biocomputing},
  pages={111--122},
  year={2018}
}

@article{zitnik2015gene,
  title={Gene prioritization by compressive data fusion and chaining},
  author={Zitnik, Marinka and Nam, Edward A and Dinh, Christopher and Kuspa, Adam and Shaulsky, Gad and Zupan, Blaz},
  journal={PLoS Computational Biology},
  volume={11},
  number={10},
  pages={e1004552},
  year={2015}
}

@article{huang2020skipgnn,
  title={{SkipGNN}: predicting molecular interactions with skip-graph networks},
  author={Huang, Kexin and Xiao, Cao and Glass, Lucas M and Zitnik, Marinka and Sun, Jimeng},
  journal={Scientific Reports},
  volume={10},
  number={1},
  pages={1--16},
  year={2020}
}

@article{gainza2020deciphering,
  title={Deciphering interaction fingerprints from protein molecular surfaces using geometric deep learning},
  author={Gainza, Pablo and Sverrisson, Freyr and Monti, Frederico and Rodola, Emanuele and Boscaini, D and Bronstein, MM and Correia, BE},
  journal={Nature Methods},
  volume={17},
  number={2},
  pages={184--192},
  year={2020}
}

@article{gysi2020network,
  title={Network medicine framework for identifying drug repurposing opportunities for {COVID-19}},
  author={Gysi, Deisy Morselli and Do Valle, {\'I}talo and Zitnik, Marinka and Ameli, Asher and Gan, Xiao and Varol, Onur and Sanchez, Helia and Baron, Rebecca Marlene and Ghiassian, Dina and Loscalzo, Joseph and others},
  journal={ArXiv:2004.07229},
  year={2020}
}

@article{zitnik2018modeling,
  title={Modeling polypharmacy side effects with graph convolutional networks},
  author={Zitnik, Marinka and Agrawal, Monica and Leskovec, Jure},
  journal={Bioinformatics},
  volume={34},
  number={13},
  pages={i457--i466},
  year={2018},
  publisher={Oxford University Press}
}

@article{stokes2020deep,
  title={A deep learning approach to antibiotic discovery},
  author={Stokes, Jonathan M and Yang, Kevin and Swanson, Kyle and Jin, Wengong and Cubillos-Ruiz, Andres and Donghia, Nina M and MacNair, Craig R and French, Shawn and Carfrae, Lindsey A and Bloom-Ackerman, Zohar and others},
  journal={Cell},
  volume={180},
  number={4},
  pages={688--702},
  year={2020},
  publisher={Elsevier}
}

@article{jumper2020high,
  title={High accuracy protein structure prediction using deep learning},
  author={Jumper, John and Evans, R and Pritzel, A and Green, T and Figurnov, M and Tunyasuvunakool, K and Ronneberger, O and Bates, R and Zidek, A and Bridgland, A and others},
  journal={Fourteenth Critical Assessment of Techniques for Protein Structure Prediction},
  volume={22},
  pages={24},
  year={2020}
}

@inproceedings{wang2019superglue,
  title={{SuperGLUE}: A stickier benchmark for general-purpose language understanding systems},
  author={Wang, Alex and Pruksachatkun, Yada and Nangia, Nikita and Singh, Amanpreet and Michael, Julian and Hill, Felix and Levy, Omer and Bowman, Samuel},
  booktitle={NeurIPS},
  pages={3266--3280},
  year={2019}
}

@inproceedings{deng2009imagenet,
  title={{ImageNet}: A large-scale hierarchical image database},
  author={Deng, Jia and Dong, Wei and Socher, Richard and Li, Li-Jia and Li, Kai and Fei-Fei, Li},
  booktitle={CVPR},
  pages={248--255},
  year={2009}
}

@article{wang2020therapeutic,
  title={Therapeutic target database 2020: enriched resource for facilitating research and early development of targeted therapeutics},
  author={Wang, Yunxia and Zhang, Song and Li, Fengcheng and Zhou, Ying and Zhang, Ying and Wang, Zhengwen and Zhang, Runyuan and Zhu, Jiang and Ren, Yuxiang and Tan, Ying and others},
  journal={Nucleic Acids Research},
  volume={48},
  number={D1},
  pages={D1031--D1041},
  year={2020},
  publisher={Oxford University Press}
}

@article{usmani2017thpdb,
  title={{THPdb: database of FDA-approved peptide and protein therapeutics}},
  author={Usmani, Salman Sadullah and Bedi, Gursimran and Samuel, Jesse S and Singh, Sandeep and Kalra, Sourav and Kumar, Pawan and Ahuja, Anjuman Arora and Sharma, Meenu and Gautam, Ankur and Raghava, Gajendra PS},
  journal={PLOS ONE},
  volume={12},
  number={7},
  pages={e0181748},
  year={2017},
  publisher={Public Library of Science San Francisco, CA USA}
}

@article{korshunovaopenchem,
  title={{OpenChem}: A Deep Learning Toolkit for Computational Chemistry and Drug Design},
  author={Korshunova, Maria and Ginsburg, Boris and Tropsha, Alexander and Isayev, Olexandr},
  journal={Journal of Chemical Information and Modeling},
  publisher={ACS Publications},
  year={2021},
}
